\documentclass[10pt,twocolumn,letterpaper]{article}

\usepackage[pagenumbers]{cvpr} 

\usepackage{graphicx}
\usepackage{amsmath}
\usepackage{amssymb}
\usepackage{booktabs}
\usepackage{multirow}
\usepackage{diagbox}
\usepackage{float}
\usepackage{algorithm}
\usepackage[noend]{algpseudocode}
\usepackage{appendix}

\usepackage[pagebackref,breaklinks,colorlinks]{hyperref}

\usepackage[capitalize]{cleveref}
\crefname{section}{Sec.}{Secs.}
\Crefname{section}{Section}{Sections}
\Crefname{table}{Table}{Tables}
\crefname{table}{Tab.}{Tabs.}

\def\cvprPaperID{8913} 
\def\confName{CVPR}
\def\confYear{2023}

\begin{document}

\title{FedTune: A Deep Dive into Efficient Federated Fine-Tuning with Pre-trained Transformers}



\author{%
   Jinyu~Chen$^1$,
   Wenchao~Xu$^1$,
   Song~Guo$^1$,
   Junxiao~Wang$^1$,
   Jie~Zhang$^1$,
   and~Haozhao~Wang$^2$,\\
  \textsuperscript{1}Department of Computing, The Hong Kong Polytechnic University\\
  \textsuperscript{2}School of Computer Science and Technology, Huazhong University of Science and Technology
}

\maketitle

\begin{abstract}
   Federated Learning (FL) is an emerging paradigm that enables distributed users to collaboratively and iteratively train machine learning models without sharing their private data.
   Motivated by the effectiveness and robustness of self-attention-based architectures, researchers are turning to using pre-trained Transformers (i.e., foundation models) instead of traditional convolutional neural networks in FL to leverage their excellent transfer learning capabilities.  
   %
   %
   %
   Despite recent progress, how pre-trained Transformer models play a role in FL remains obscure, that is, how to efficiently fine-tune these pre-trained models in FL and how FL users could benefit from this new paradigm. 
   %
   In this paper, we explore this issue and demonstrate that the fine-tuned Transformers achieve extraordinary performance on FL, and that the lightweight fine-tuning method facilitates a fast convergence rate and low communication costs.
   Concretely, we conduct a rigorous empirical study of three tuning methods (i.e., modifying the input, adding extra modules, and adjusting the backbone) using two types of pre-trained models (i.e., vision-language models and vision models) for FL.
   Our experiments show that 1)
   Fine-tuning the bias term of the backbone performs best when relying on a strong pre-trained model; 
   2) The vision-language model (e.g., CLIP) outperforms the pure vision model (e.g., ViT) and is more robust to the few-shot settings;
   3) Compared to pure local training, FL with pre-trained models has a higher accuracy because it alleviates the problem of over-fitting. 
   We will release our code and encourage further exploration of pre-trained Transformers and FL.

\end{abstract}

\section{Introduction}
\label{sec:intro}

The past few years have witnessed the rapid development of Federated Learning (FL), a decentralized machine learning framework that preserves user privacy since no raw data are shared to a remote server~\cite{mcmahan2017communication}.
%
In FL, each participant performs the local training using private data and uploads the gradient to the server, where all user's updates are aggregated to refine the global model, which is then broadcast backward to all clients for the next iteration.
%
FL have benefited a wide range of applications, such as healthcare~\cite{antunes2022federated}, recommendation~\cite{ammad2019federated} and Internet of Things~\cite{khan2021federated}. 
However, existing FL methods primarily trained a convolutional model from scratch~\cite{zhang2020personalized,ma2022layer}. These approaches are often plagued by data scarcity, high computational complexity, high communication costs, and diverse client data distributions.

%

Recently, Transformers and their self-attention-based architectures have attracted considerable attention.
Qu \etal~\cite{qu2022rethinking} was the first to apply Transformers to FL and showed that Transformers are more effective and robust to the client heterogeneity compared to canonical architectures (i.e., CNNs).
Previous works exploring the reasons behind their effectiveness found that Transformers are highly robust to severe occlusions, perturbations, domain
shifts~\cite{bhojanapalli2021understanding,naseer2021intriguing}, as well as synthetic and natural adversarial examples~\cite{mahmood2021robustness,paul2022vision}. 
Evidence also suggests that Transformers typically require a large amount of training data to perform better than CNNs when trained from scratch~\cite{dosovitskiy2020image}.

Thanks to the emerging paradigms which are called large-scale pre-training and parameter-efficient fine-tuning~\cite{bommasani2021opportunities}, Transformers can be pre-trained on massive data and adapted (i.e., fine-tuned) to a wide range of downstream tasks~\cite{sun2022paradigm,yuan2021florence}.
Despite recent progress, a fundamental question remains open, i.e., \emph{how to efficiently fine-tune these pre-trained models in FL and how FL benefits from this new paradigm}.
In this paper, we explore this issue and investigate FL from a new and comprehensive perspective by \emph{collaboratively fine-tuning the large-scale pre-trained model (a.k.a. foundation model) in a parameter-efficient manner}.
%
This novel training paradigm will enable clients to learn on the basis of knowledgeable experts and thus reduce the resource cost of the system.
%
%
%
%
%

%
%

Concretely, our investigation is based on a generic class of Transformers that have been demonstrated primitive potential and have been deployed in the real world for visual recognition tasks.
The Transformers include vision-language models (i.e., CLIP~\cite{radford2021learning}) and pure vision models (i.e., ViT~\cite{dosovitskiy2020image}).
%
%
To apply the pre-trained models to FL's downstream tasks, we classify existing fine-tuning approaches into three categories, i.e., modifying from the input (e.g., prompt tuning~\cite{zhou2022learning}), adding extra modules (e.g., adapter tuning~\cite{gao2021clip}), and adjusting the original backbone (e.g., bias tuning~\cite{cai2020tinytl}).
%
%

We conduct a rigorous empirical study of three tuning methods using two types of pre-trained models under various FL settings.
%
%
Experimental results show that 1) FL using pre-trained models can achieve higher accuracy, and CLIP with bias tuning has the best performance.
%
2) CLIP outperforms ViT and is more robust to the few-shot data distribution.
3) Bias tuning is the most effective method when relying on a strong pre-trained model.
%
%
4) FL is necessary when training with the foundation model, while pure local training is a strong baseline in the extreme non-IID setting.
%
%
5) The trained parameters are lightweight, and the pre-trained model helps FL to converge faster, thus reducing resource costs, and has a large performance gain compared to traditional CNN models.
%
The contributions of this paper are summarized as follows:
\begin{itemize}
    \item We investigate FL using large-scale pre-trained Transformer models and their comprehensive fine-tuning methods, which significantly changes the landscape of current FL research.
    
    \item We take the first step to conduct an in-depth measurement on various parameter-efficient tuning methods in FL using different pre-trained Transformer models (i.e., foundation models), and the results reveal the extraordinary performance of pre-trained models in FL, especially using CLIP models with bias tuning.
    
    \item Our work can serve as a benchmark and guide for future exploration of foundation models and FL.
\end{itemize}

\section{Related Work}
\label{relatedwork}
\subsection{Large-scale Pre-trained Model}

In recent years, the paradigm of AI has shifted from feature engineering or architecture engineering to the pre-training and fine-tuning mode that can release the potential of big data and big model~\cite{bommasani2021opportunities}. Large-scale pre-trained model firstly shines in the NLP area~\cite{han2021pre} where GPT constructed a generative task to derive a pre-trained model for feature extraction~\cite{radford2018improving}. Researchers designed BERT with two pre-training tasks (i.e., next sentence prediction and mask prediction) and showed their powerful abilities for NLP tasks~\cite{devlin2018bert}. GPT-3 with 175 billion parameters was built upon GPT and was pre-trained to be an excellent few-shot learner~\cite{brown2020language}. Given that there exists countless vision-language data on the internet, pre-training with these multi-modal data helps construct strong encoders. CLIP demonstrated that a simple contrastive learning task with lots of multi-modal inputs can achieve desirable zero-shot learning performance~\cite{radford2021learning}. Currently, pre-training with Transformer is widely adopted in vision tasks. Vision Transformer (ViT) with a patch-level self-attention architecture was pre-trained on the ImageNet-21k dataset and outperformed previous CNN models~\cite{dosovitskiy2020image}. Swin Transformer applied the shifted window in ViT and improved the efficiency~\cite{liu2021swin}. MAE pre-trained the ViT model through mask and reconstruction tasks~\cite{he2022masked}. However, all of them were trained and inferred in a centralized manner without considering user privacy.

\subsection{Parameter-efficient Fine-tuning}

A critical step in transferring these pre-trained models to various downstream tasks is fine-tuning.  Houlsby \emph{et al.} designed a down-up projection module in each transformer layer for BERT~\cite{houlsby2019parameter}. Prompt engineering was proposed to construct an informative input for the foundation model~\cite{liu2021pre}. Prefix tuning prepended some tunable prefix prompts at each Transformer layer~\cite{li2021prefix}. Coop first introduced prompt learning into vision-language models and added the learnable vectors in the text input~\cite{zhou2022learning}. CoCoop learned a lightweight meta-network to generate a prompt based on the image feature~\cite{zhou2022conditional}. VPT proposed a prompt learning method for ViT and achieved desirable performance for vision tasks~\cite{jia2022vpt}. Another line of work fine-tuned the backbone of the pre-trained model. Bias tuning only adapted the bias term of the network~\cite{cai2020tinytl}, and LoRA added trainable low-rank matrices parallel to the multi-head attention~\cite{hu2021lora}. He \emph{et al.} unified these fine-tuning approaches into a theoretical framework~\cite{he2021towards}. However, all of these methods were trained centrally in the server with the assumption that the source data is accessible. 

\subsection{Federated Learning}

Federated learning attracts lots of attention in academia. Existing works focused on dealing with non-IID issues~\cite{luo2021no,gao2022feddc}, designing a personalized model~\cite{li2021model,ma2022layer}, improving the communication efficiency~\cite{wu2022communication,wang2022progfed}, and so on.
Recently, a few works tended to apply model pre-training to FL for better parameter initialization. 
Qu \emph{et al.} showed that Transformer can better deal with data heterogeneity and pre-training helps improve the accuracy~\cite{qu2022rethinking}. Chen \emph{et al.} investigated that a traditional CNN model pre-trained with ImageNet can alleviate the problem of non-IID in FL and proposed a method for pre-training with synthetic
data~\cite{chen2022pre}. Nguyen \emph{et al.} also found that model initialization with the pre-trained parameters can reduce the training time and improve the accuracy~\cite{nguyen2022begin}. 
FedPCL used pre-trained models for feature extraction and fused them with a projection network which was collaboratively trained through the prototype-wise contrastive learning~\cite{tan2022federated}. PromptFL applied prompt learning with CLIP in FL and achieved desirable results compared to full tuning~\cite{guo2022promptfl}. However, none of the work investigates the effect of parameter-efficient tuning methods with different foundation models in FL, which is of great importance for applying pre-trained models to FL.

\section{Fine-tuning Methods with Pre-trained Models in Federated Learning}

In this section, we present background on pre-trained models and their fine-tuning methods. Then, we modify the existing FL framework to efficient fine-tuning with pre-trained Transformers.

\subsection{Pre-trained Models}

In this paper, we focus on vision tasks and introduce two kinds of pre-trained models, i.e., vision-language model and vision model.

\begin{figure}[htp]
  \centering
   \includegraphics[width=0.8\linewidth]{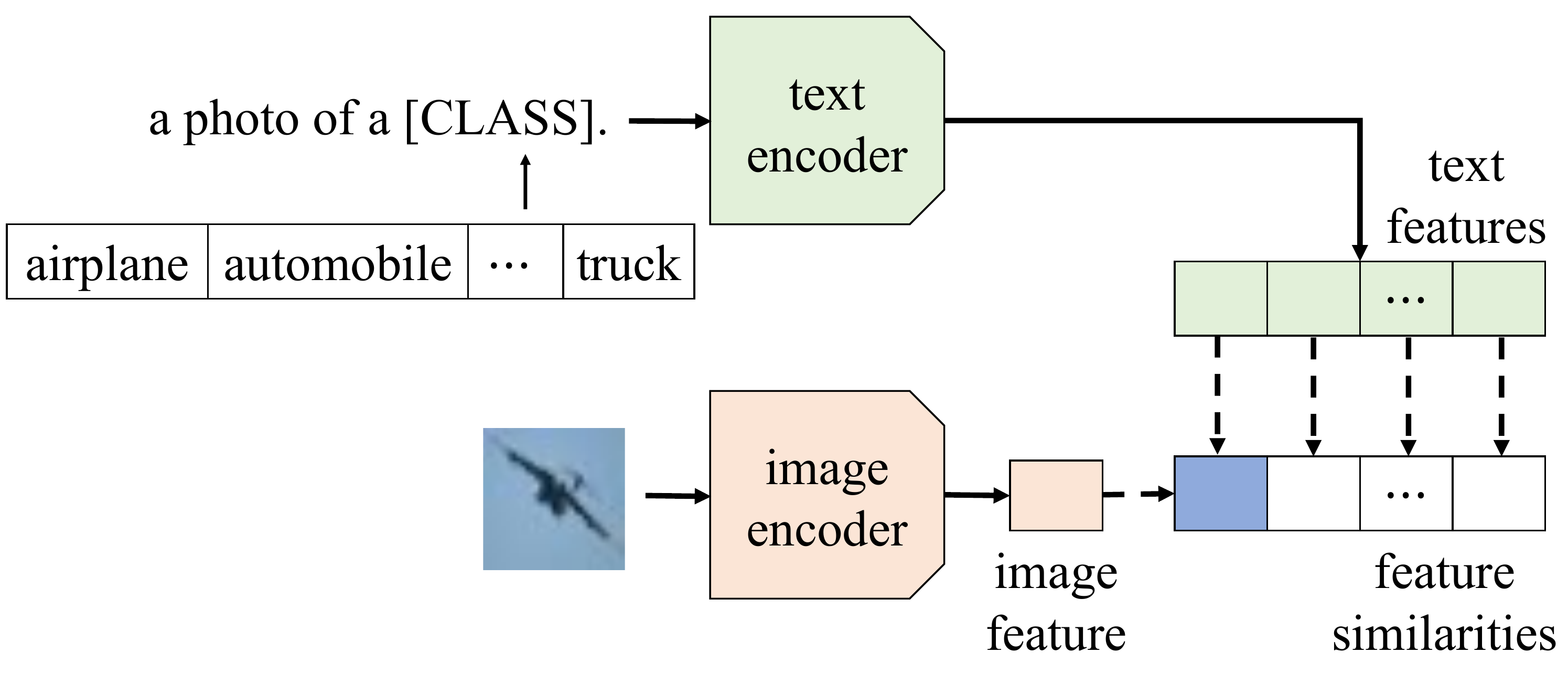}
   \caption{CLIP Model.}
   \label{fig:clip}
\end{figure}

\textbf{Vision-Language Model.}
The vision-language pre-trained model aims to use visual and textual information for mutual supervision during pre-training.
We introduce one of the most famous vision-language models--CLIP~\cite{radford2021learning} as an illustration. As shown in Fig.~\ref{fig:clip}, a CLIP model includes two separate encoders, i.e., a text encoder with a Transformer architecture~\cite{vaswani2017attention} and an image encoder with a CNN network~\cite{he2016deep} or a ViT model~\cite{dosovitskiy2020image}. CLIP is pre-trained on 400 million (image, text) pairs collected from the internet. During the training stage, a batch of $N$ (image, text) pairs are sampled from the dataset, and $N \times N$ image and text features are extracted by encoders. Then, pair-level similarities are calculated based on the cosine similarity between visual and textual features. Finally, a contrastive representation learning technique~\cite{oord2018representation} is applied to minimize the distances between positive pairs and maximize the distances between negative pairs. With plentiful dataset resources and 
large foundation models, the text encoder and image encoder are equipped with strong capabilities for textual and visual understanding.

During testing, the class names are concatenated with the prompt ``a photo of a" and then inputted into the text encoder. The image feature is extracted by the image encoder, and the prediction probability is calculated based on the feature similarities. Finally, the class with the largest similarity score is selected.

\begin{figure}[t]
    \centering
    \begin{minipage}[t]{0.24\textwidth}
    \centering
    \includegraphics[width=1\textwidth]{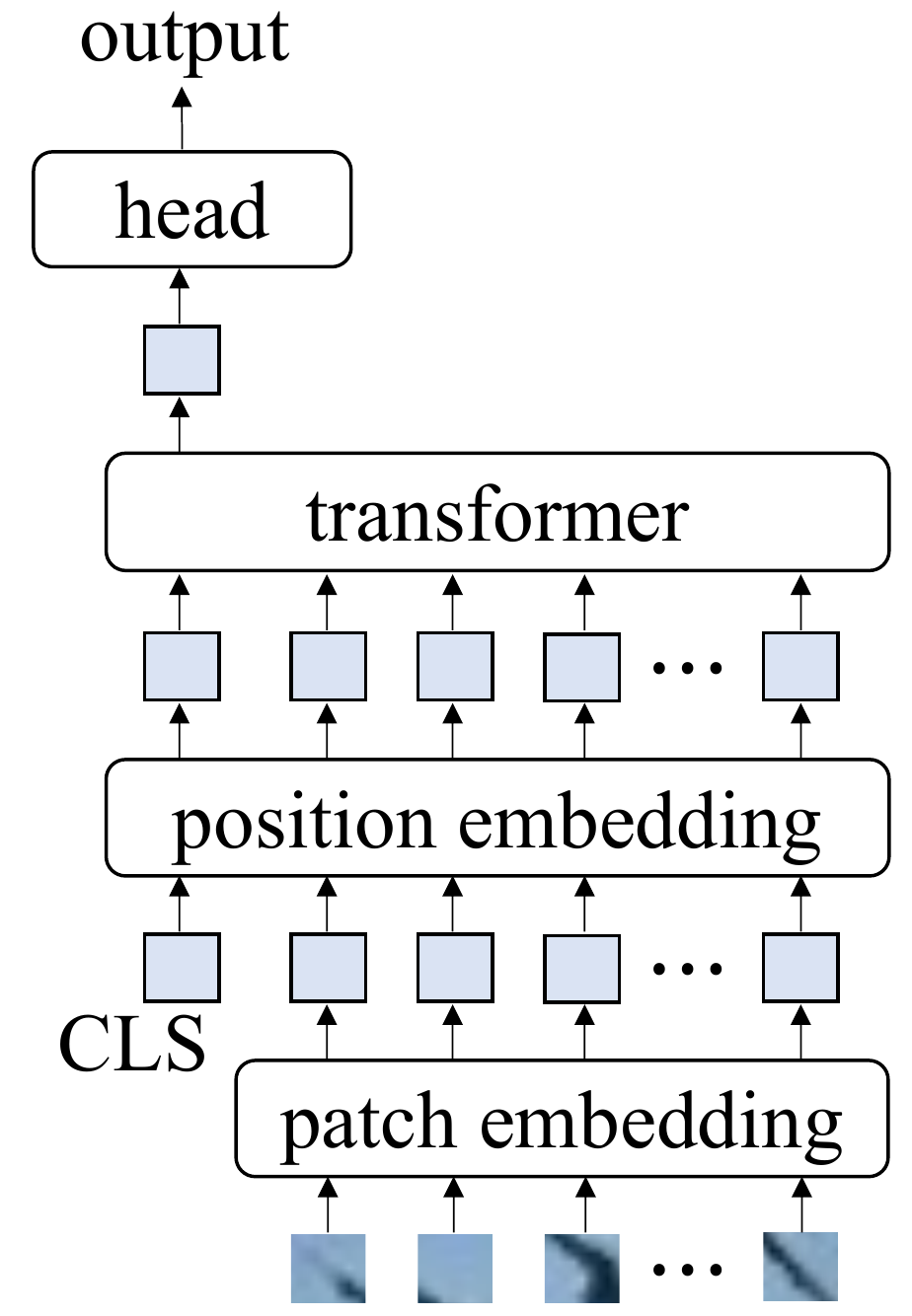}
    \caption{Vision Transformer.}
    \label{fig:vit}
    \end{minipage}
    \begin{minipage}[t]{0.23\textwidth}
    \centering
    \includegraphics[width=1\textwidth]{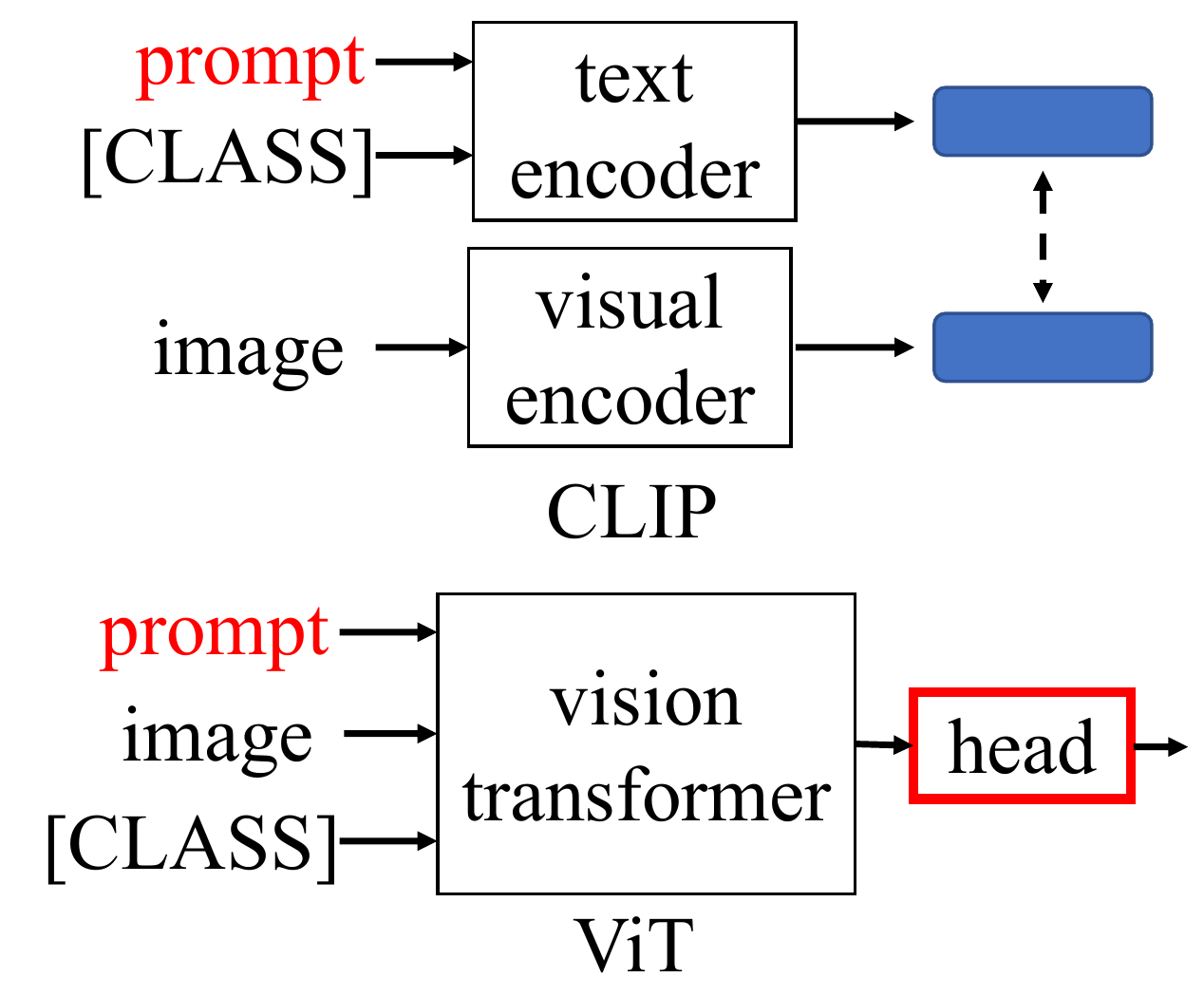}
    \caption{Prompt Tuning.}
    \label{prompt}
    \end{minipage}
\end{figure}

\textbf{Vision Model.} Vision Transformer~\cite{dosovitskiy2020image} is now prevalent in computer vision research and shows remarkable performance compared to traditional CNN models. The model architecture of ViT is shown in Fig.~\ref{fig:vit}. A 2D image is reshaped into a list of small patches that are processed by patch embedding. A class token is appended to the embedding vectors and added with position tokens. Then, the Transformer equipped with self-attention layers is used to process the token sequence. Finally, the output of the class token is projected with the ViT head to get the prediction result. 

\begin{figure}[htp]
    \centering
    \begin{minipage}[t]{0.25\textwidth}
    \centering
    \includegraphics[width=1\textwidth]{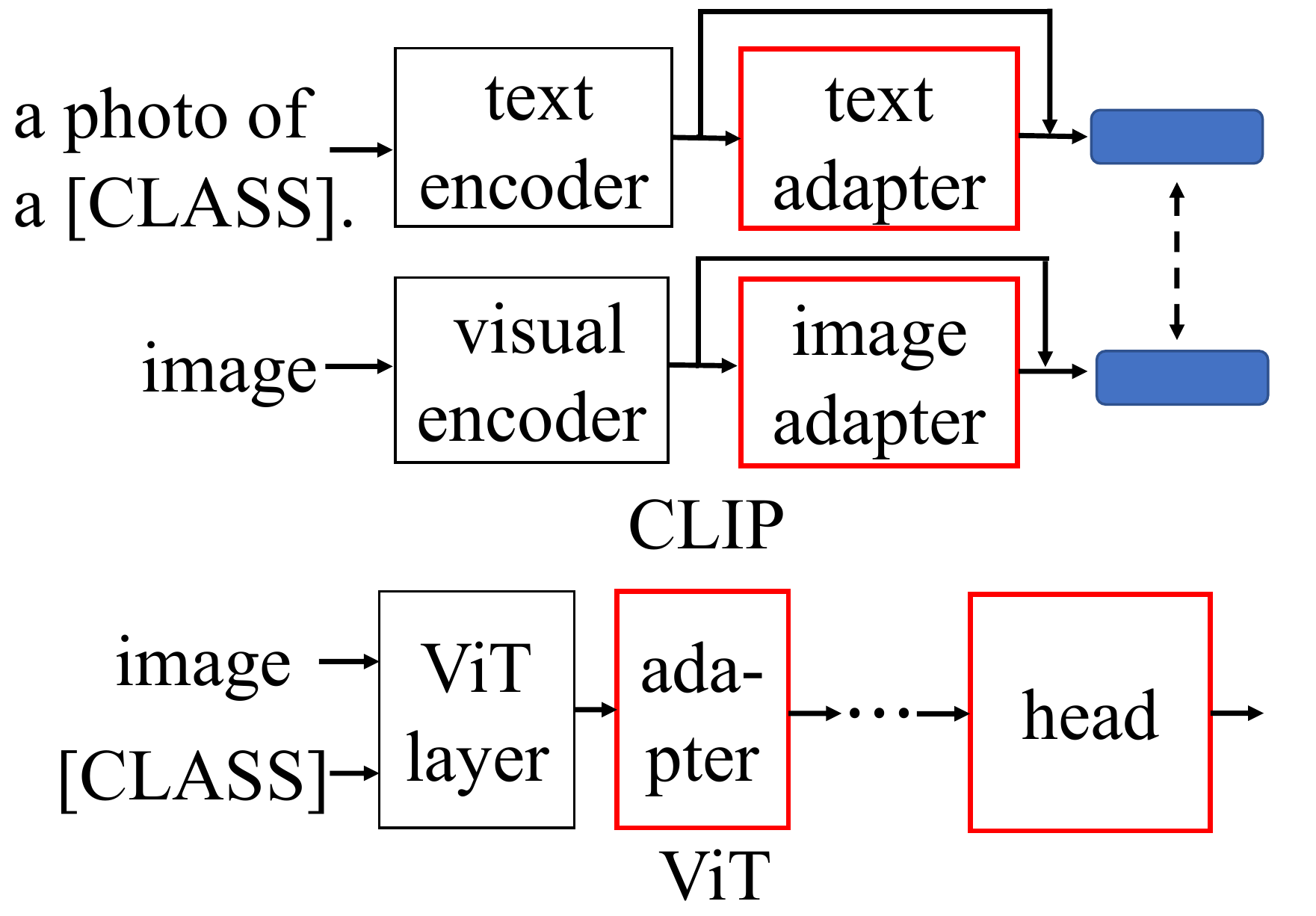}
    \caption{Adapter Tuning.}
    \label{fig:adapter}
    \end{minipage}
    \begin{minipage}[t]{0.22\textwidth}
    \centering
    \includegraphics[width=1\textwidth]{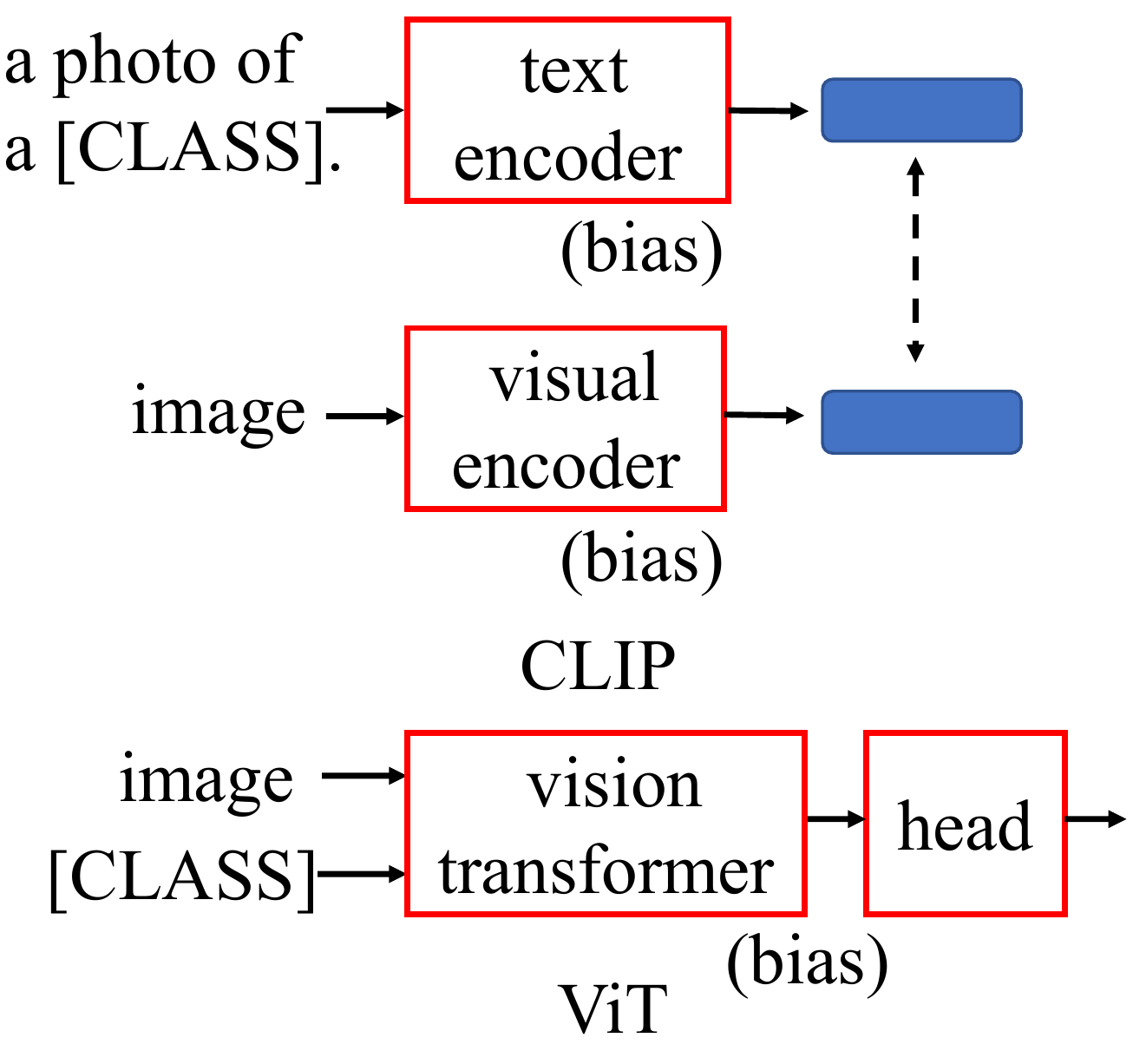}
    \caption{Bias Tuning.}
    \label{bias}
    \end{minipage}
\end{figure}

\subsection{Fine-tuning Methods}

Many parameter-efficient tuning methods are proposed to transfer the powerful pre-trained model to downstream tasks. We divide current fine-tuning methods into three categories as below.

\textbf{Modify from input.}
Prompt learning is an efficient and effective fine-tuning approach that prepends some instructive tokens to the input and doesn't need to revise the foundation model. Given that designing a good prompt manually is hard and laborious, existing works used soft learnable prompts that are trained with downstream data. Coop~\cite{zhou2022learning} first introduced prompt learning in CLIP and added some trainable prompts in replace of the ``a photo of a" prompt. VPT~\cite{jia2022vpt} was the first to use prompt learning in ViT. It introduced a small number of learnable prompts in front of the image tokens. The concatenated tokens are then inputted into the Transformer model. A vivid illustration of these methods is shown in Fig.~\ref{prompt}.

\textbf{Modify with additional modules.}
Another tuning method is inserting additional modules, called adapters, into the foundation model. The adapter is used to fit downstream datasets while the pre-trained model keeps frozen. As shown in Fig.~\ref{fig:adapter}, CLIP-Adapter added the residual adapter after the encoders to fine-tune the CLIP model~\cite{gao2021clip}. For vision tasks, a fine-grained adapter is designed for ViT, and some Multi-Layer Perception (MLP) modules are inserted into each Transformer layer with residual connection~\cite{pfeiffer2020adapterhub, jia2022vpt}.

\textbf{Modify from the backbone.}
Previous work demonstrated that the major obstacle to training the foundation model is the activation term~\cite{cai2020tinytl}. To save memory consumption and improve performance, TinyTL fine-tuned only the bias term of parameters for efficient on-device learning. We regard this kind of method as modifying from the backbone. As shown in Fig.~\ref{bias}, we transfer CLIP and ViT to downstream tasks by tuning the bias term of their backbones.

\subsection{Federated Learning}

\begin{figure}[htp]
  \centering
   \includegraphics[width=0.8\linewidth]{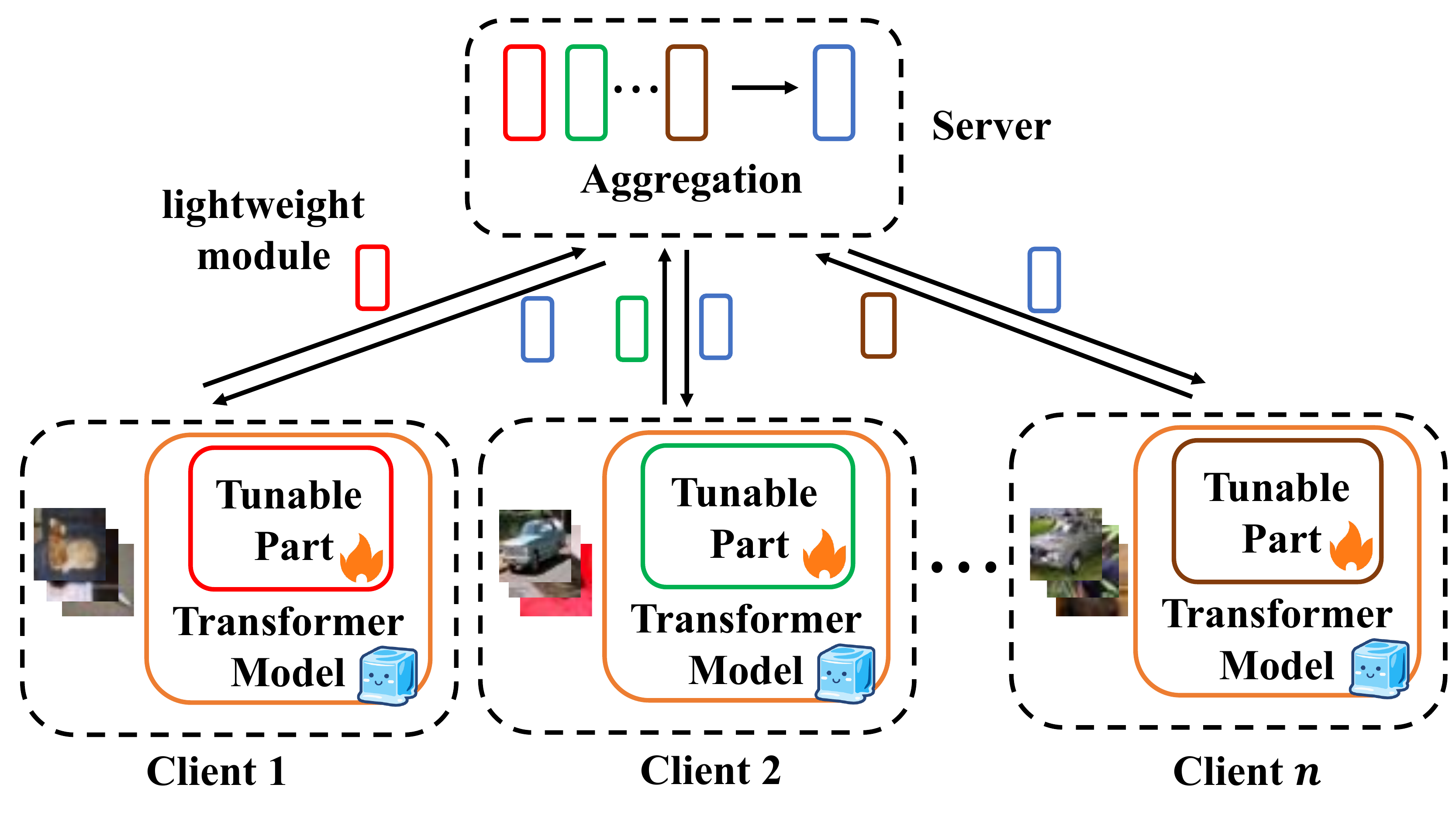}
   \caption{The framework of federated fine-tuning with pre-trained Transformers.}
   \label{fig:fed_framework}
\end{figure}

We design a novel FL framework with the foundation model and use one of the most famous algorithms (i.e., FedAVG~\cite{mcmahan2017communication}) for aggregation. The framework is shown in Fig.~\ref{fig:fed_framework}. At each global round, a fraction of users are randomly sampled as active workers, and the global model is distributed to downstream clients. At the local training stage, the selected clients freeze the backbone of the foundation model and fine-tune the lightweight modules with a few epochs. Then, the gradients of the fine-tuned parameters are extracted and sent back to the server. The server aggregates them into a new global model with weighted averages. This process repeats until reaching the convergence condition or pre-defined epochs.\footnote{Formal problem definition and the pseudo-code are shown in the appendix.}

\section{Measurements and Analyses on Efficient Federated Fine-tuning}

To have a clear understanding of parameter-efficient fine-tuning of foundation models in FL, we conduct an in-depth measurement study in real settings. Our experiments are designed to unveil the effect of pre-trained Transformers in FL and answer the following questions:

\begin{itemize}
    \item Are fine-tuning methods able to learn a better global model for FL? Are these methods robust to the heterogeneous client distributions?
    \item Considering that the pre-trained model achieves a desirable accuracy with few local data, is FL still necessary?
    \item How much performance can the fine-tuned Transformers improve compared to the CNN model?
    \item How is the approach affected by various settings?
    \item What is the resource consumption of these methods?
\end{itemize}

\subsection{Experimental Setup}

We evaluate the approaches on the CIFAR-10 dataset and CIFAR-100 dataset.\footnote{The results of CIFAR-100 is reported in the appendix.} According to previous parameter-efficient tuning methods~\cite{radford2021learning,zhou2022learning}, we consider the few-shot learning scenario. There are 10 clients participating in the FL, and all users are selected at each global round.
We simulate three kinds of local data distributions: (1) \textbf{IID}. Each client is randomly assigned 10\% of the overall dataset, and the training data is further sampled for 1, 2, 4, 8, and 16 shots. (2) \textbf{non-IID}. The data is sorted by label and divided into 20 shards. Each client randomly picks 2 shards. The training data of clients is sampled for 1, 2, 4, 8, and 16 shots. (3) Generating with a Dirichlet distribution $\textbf{Dir}(\alpha)$~\cite{zhu2021data}. We first randomly select 80 samples for each class in the training set and then distribute the training data to clients according to the Dirichlet distribution. Smaller $\alpha$ means higher data heterogeneity (i.e., higher non-IID). The values of $\alpha$ are set as 0.01, 0.1, 1, and 100. The testing data is allocated to all users according to the distribution of the training data. We use the weighted average accuracy across clients for evaluation.

We use CLIP with ViT-Base/16 as a vision-language model and ViT Base with 224 $\times$ 224 image size and 16 patches as a pure vision model.
The CLIP model is pre-trained on 400 million (image, text) pairs, and the ViT model is pre-trained on the ImageNet-21K dataset. Users fine-tune the model with 5 local epochs, and the global communication round is 50 epochs. We use the cross-entropy loss and the SGD optimizer with 0.002 learning rate for local training.

Implementation. All algorithms are implemented in PyTorch and run on an RTX 3090Ti GPU.

\subsection{Comparison Results}

\begin{figure}[htp]
  \centering
  \begin{subfigure}{0.48\linewidth}
    \includegraphics[width=1\textwidth]{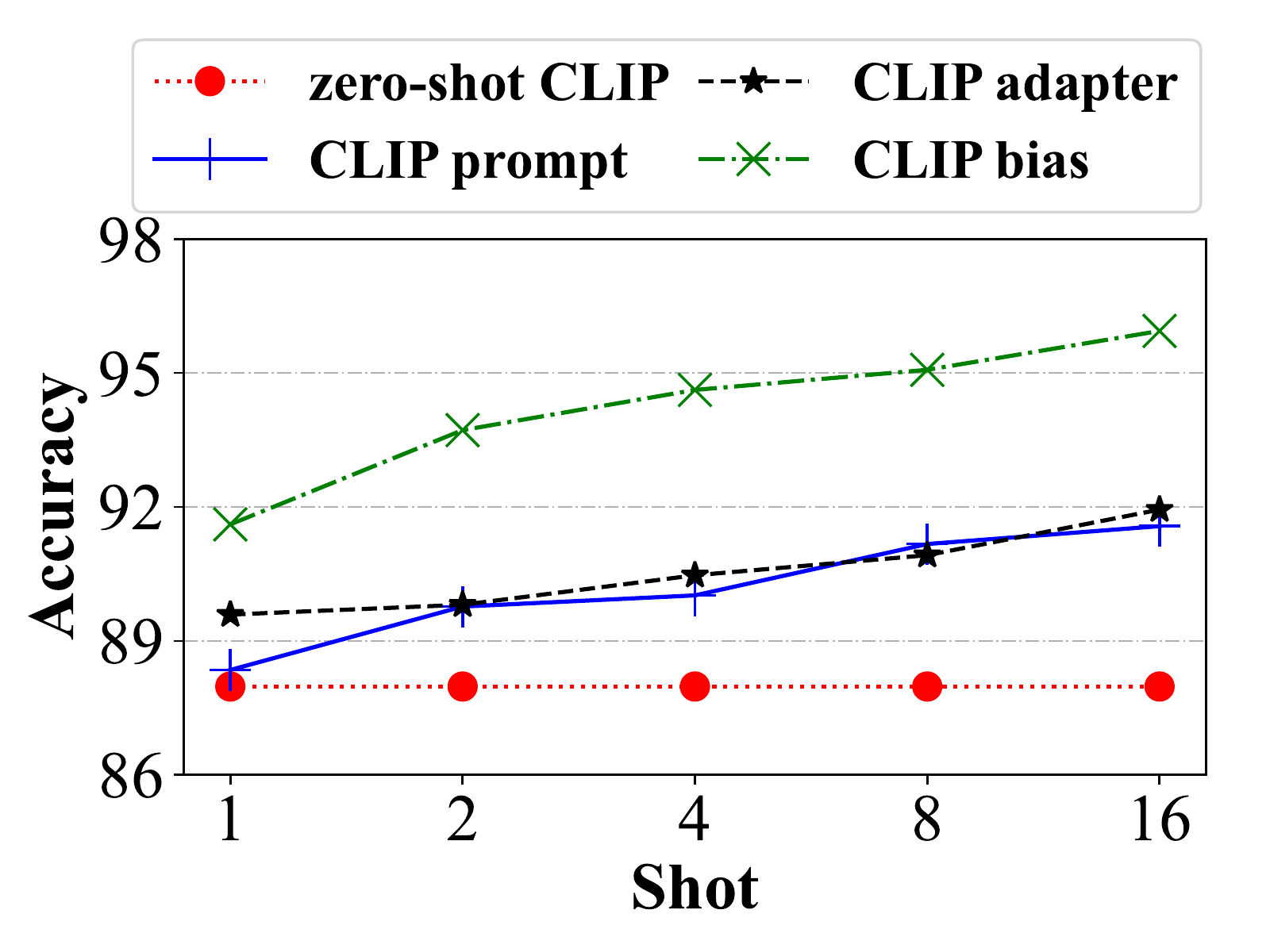}
    \caption{Accuracy comparison of CLIP models under IID setting.}
    \label{fig:acc-a}
  \end{subfigure}
  \begin{subfigure}{0.48\linewidth}
    \includegraphics[width=1\textwidth]{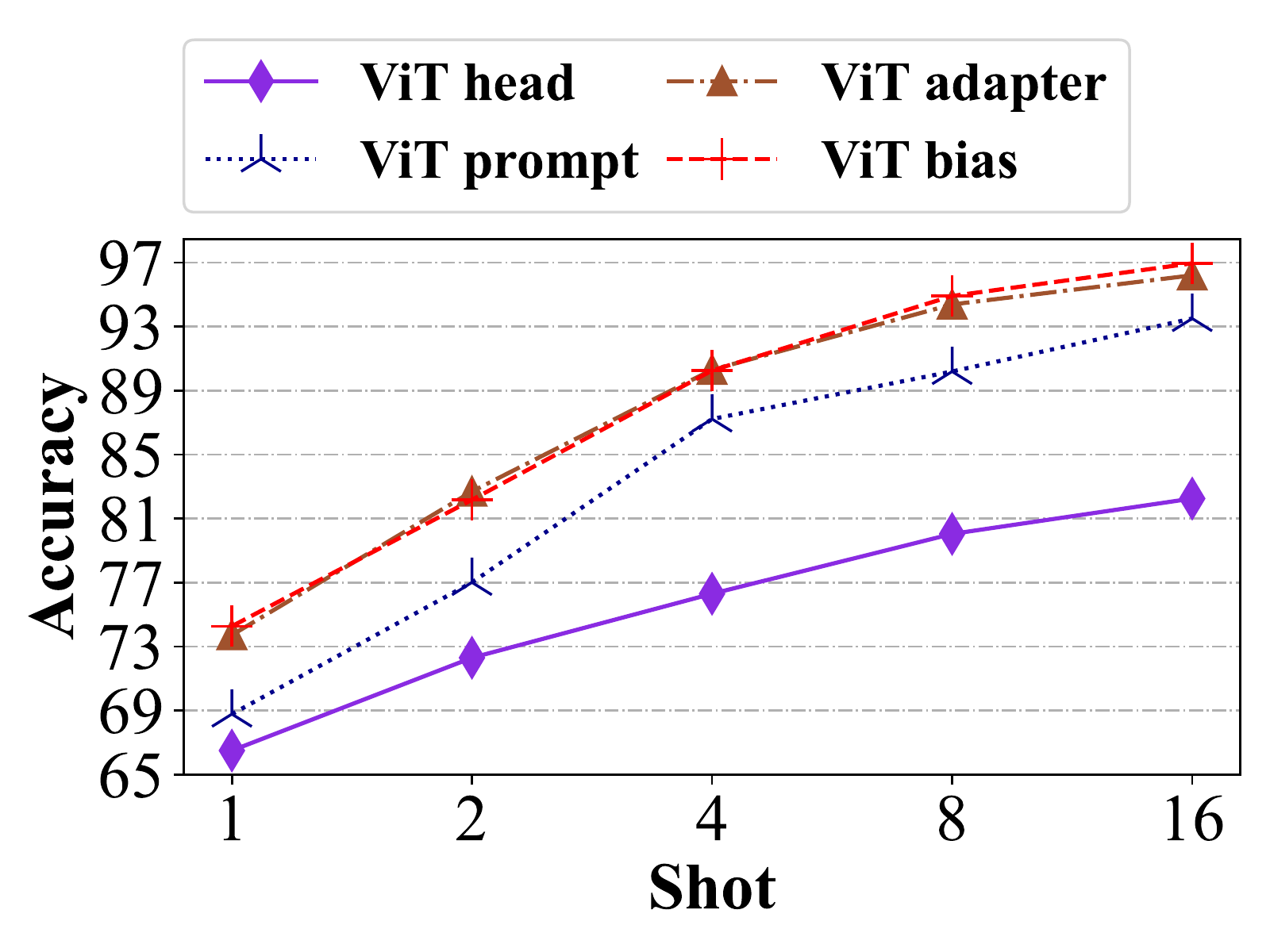}
    \caption{Accuracy comparison of ViT models under IID setting.}
    \label{fig:acc-b}
  \end{subfigure}
  \label{fig:short}
  \begin{subfigure}{0.48\linewidth}
    \includegraphics[width=1\textwidth]{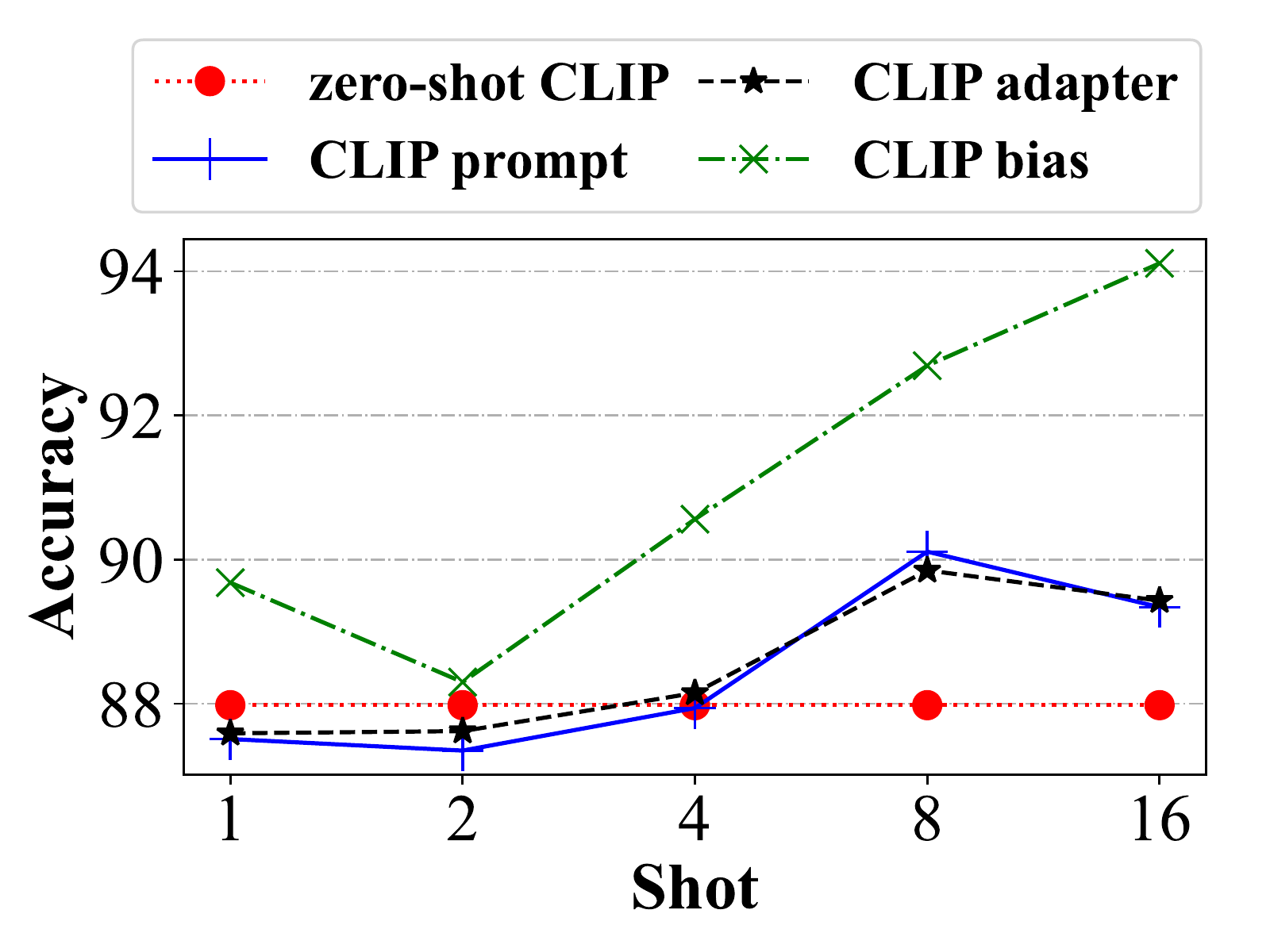}
    \caption{Accuracy comparison of CLIP models under non-IID setting.}
    \label{fig:acc-c}
  \end{subfigure}
  \begin{subfigure}{0.48\linewidth}
    \includegraphics[width=1\textwidth]{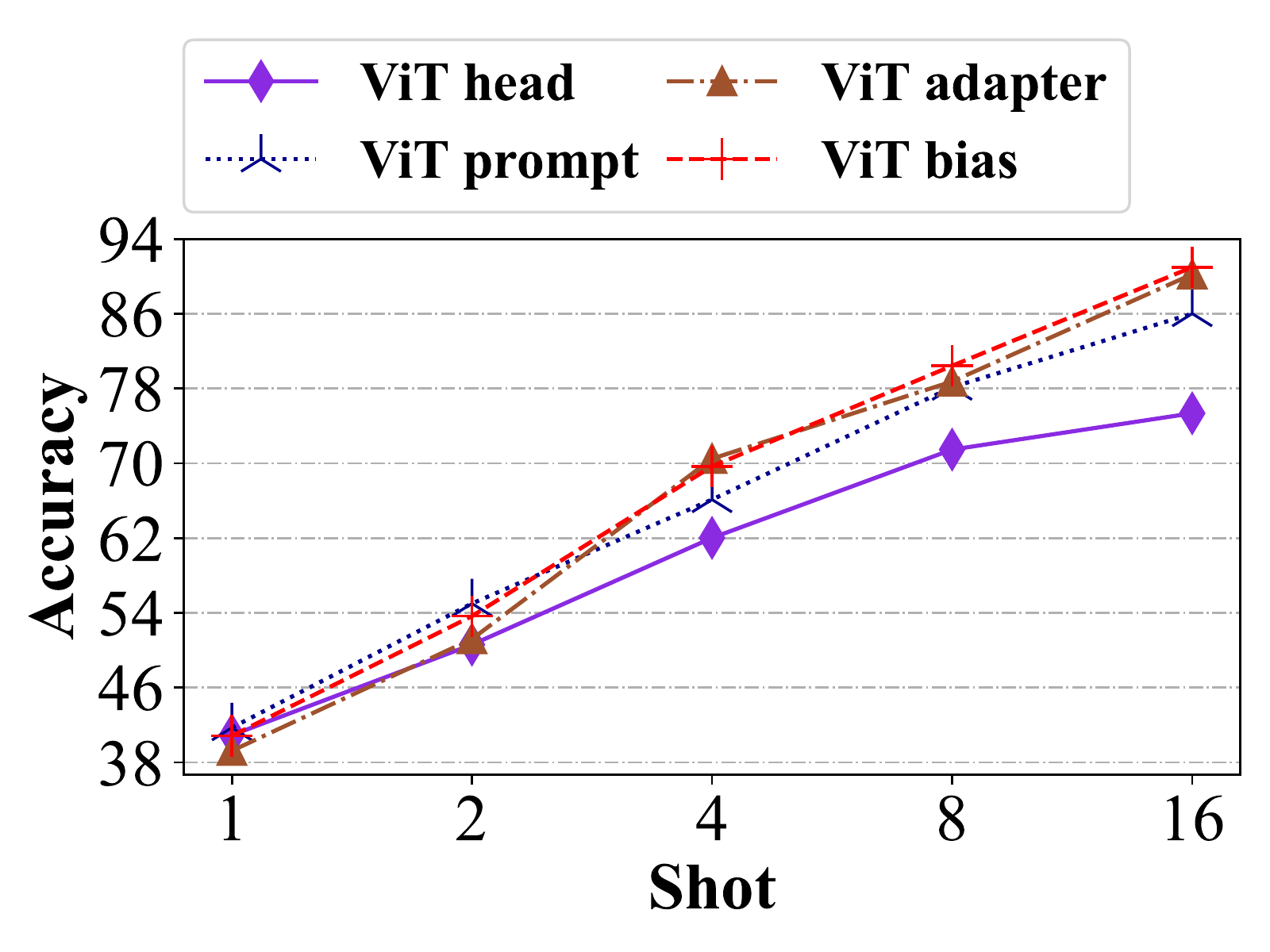}
    \caption{Accuracy comparison of ViT models under non-IID setting.}
    \label{fig:acc-d}
  \end{subfigure}
  \caption{Accuracy comparison under IID and non-IID settings.}
  \label{fig:acc}
\end{figure}

\begin{figure}[htp]
  \centering
   \includegraphics[width=0.8\linewidth]{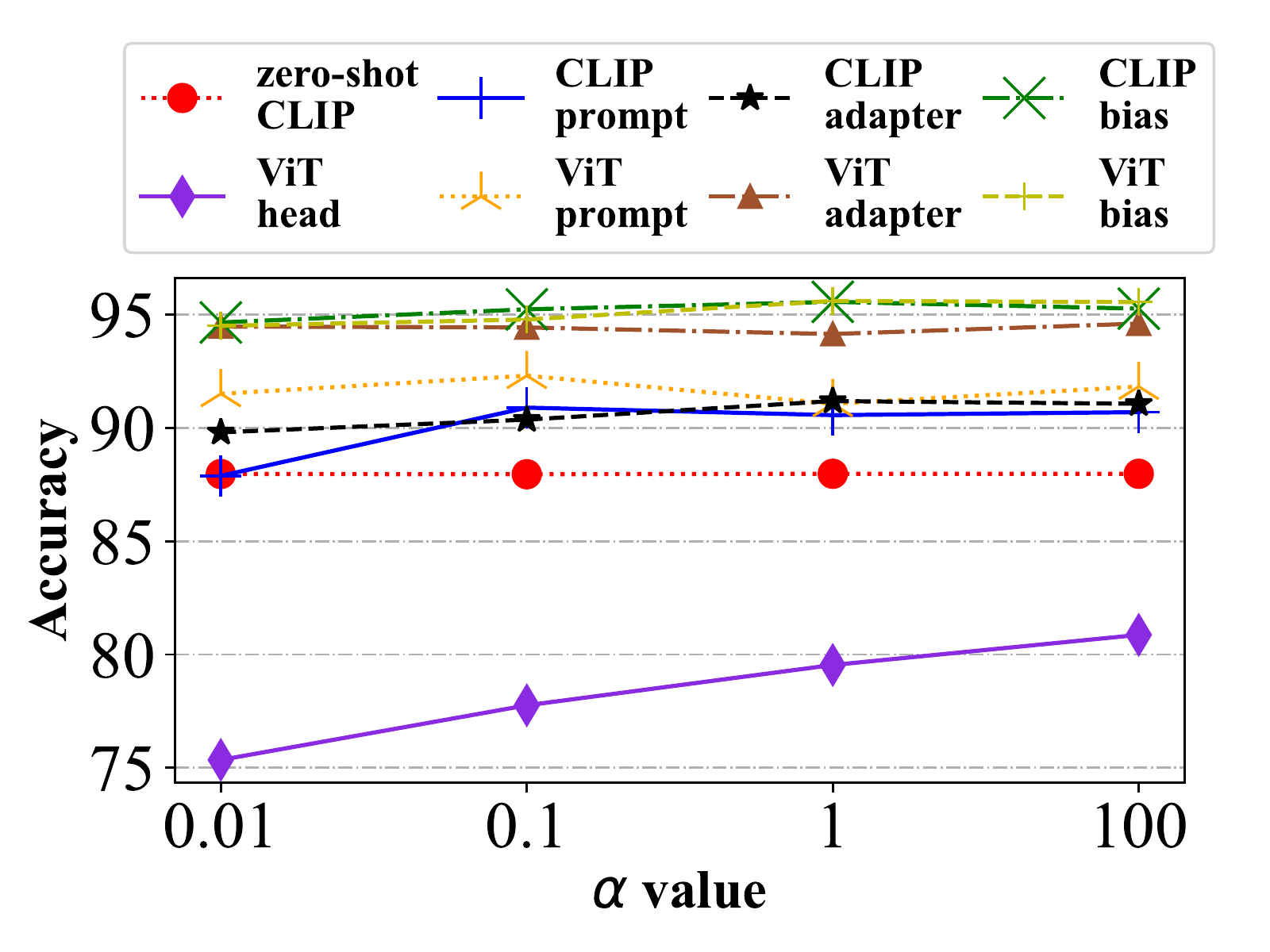}
   \caption{Accuracy comparison under different Dirichlet distributions.}
   \label{fig:dir_acc}
\end{figure}

The comparisons of accuracy under IID and non-IID settings are shown in Fig.~\ref{fig:acc}. In Fig.~\ref{fig:acc-a}, CLIP with bias tuning performs best and can reach 95.94\% accuracy in the 16-shot learning. CLIP prompt and CLIP adapter have similar performance and are better than zero-shot CLIP that has 87.98\% accuracy. In Fig.~\ref{fig:acc-b}, ViT adapter and ViT bias show similar trends and can reach about 97\% accuracy. ViT head is the worst because it doesn't use any tuning methods. CLIP bias, whose accuracy increases from 89.68\% to 94.11\%, shows a significant advantage in the non-IID setting in Fig.~\ref{fig:acc-c}. We notice that in the 1-shot and 2-shot learning, the performance of CLIP prompt and CLIP adapter is worse than that of the zero-shot CLIP. It indicates that these fine-tuned modules cannot learn enough information from local data and will cause over-fitting, which will damage the foundation model~\cite{prompt-aligned}. 
Data capacity has a noticeable impact on ViT methods, and there is a significant rising trend in Fig.~\ref{fig:acc-d}.
Besides, ViT bias in the 16-shot learning has the highest accuracy (i.e., 90.98\%).

Accuracy comparison under different Dirichlet distributions is shown in Fig.~\ref{fig:dir_acc}. The heterogeneity of data distribution has little impact on the performance of all fine-tuning methods. The accuracies of these methods are higher than 90\%, and the CLIP bias (95.27\% in 16 shots) and ViT bias (95.56\% in 16 shots) achieve the highest scores. All methods except ViT head are better than zero-shot CLIP.

One explanation for the performance of these tuning methods is that prompt learning is far from the output with a long propagation path that cannot process the high-level features, and the adapter is far from the input that cannot access the low-level features~\cite{liu2022late}. On the contrary, bias tuning trains the bias terms which are evenly distributed in the backbone and can deal with features at different levels. 
Besides, the diverse distribution of the pre-training dataset contributes to the success of CLIP~\cite{fang2022data}.

\textbf{Insight 1: } CLIP model with bias tuning outperforms other methods, and ViT model with head tuning is the worst. For different foundation models, CLIP has better performance than ViT and is more robust to various data distributions because it learns plentiful knowledge from the pre-trained dataset. For different fine-tuning methods, the accuracy comparisons are bias \textgreater adapter \textgreater prompt. All fine-tuning methods are robust to the heterogeneous data distributions. Fine-tuning methods of the CLIP model are better than zero-shot CLIP in most of the scenarios.

\subsection{Comparison with Local Training}

\begin{figure}[htp]
  \centering
  \begin{subfigure}{0.49\linewidth}
    \includegraphics[width=1\textwidth]{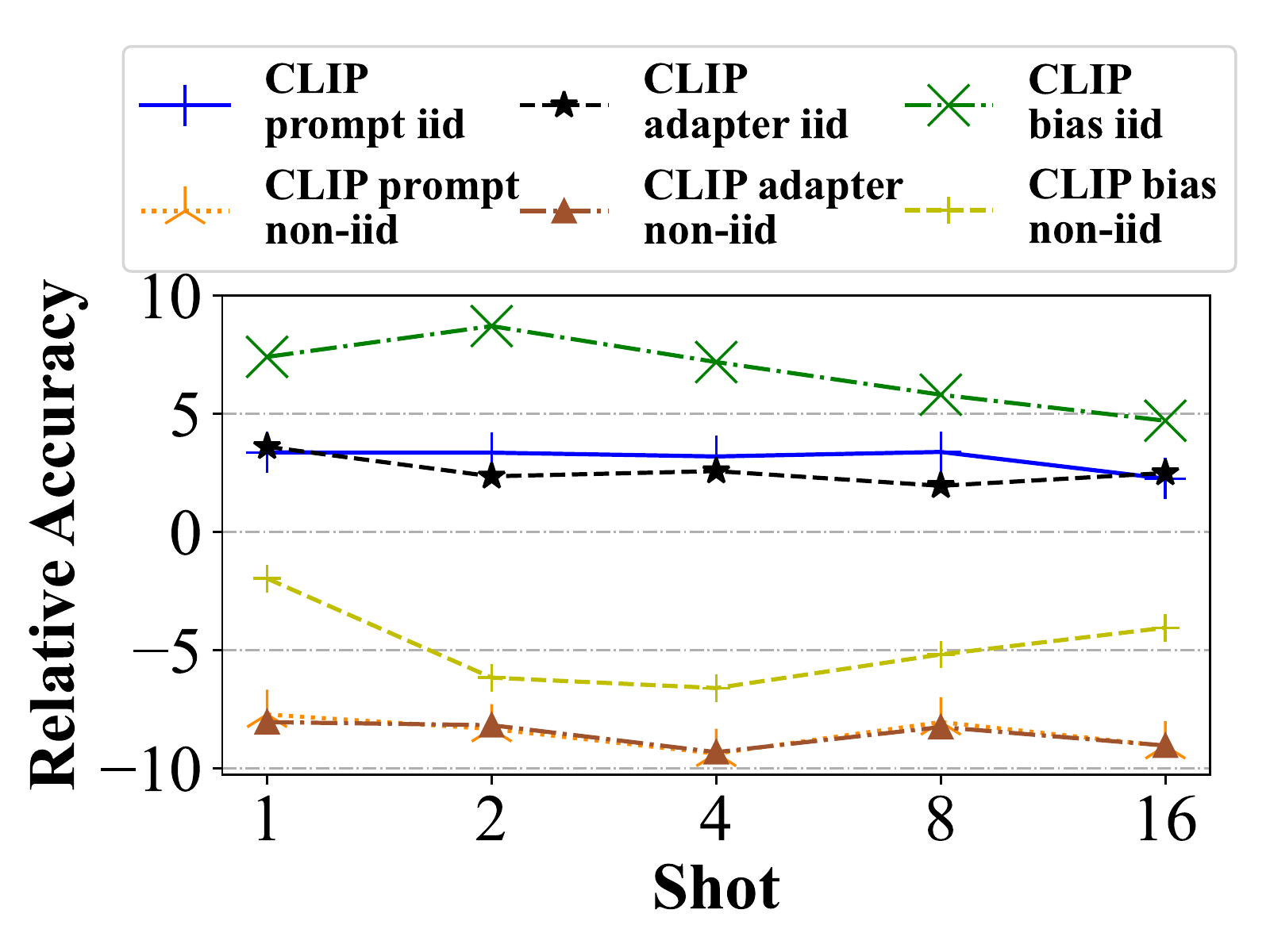}
    \caption{CLIP model.}
    \label{fig:local_clip}
  \end{subfigure}
  \hfill
  \begin{subfigure}{0.49\linewidth}
    \includegraphics[width=1\textwidth]{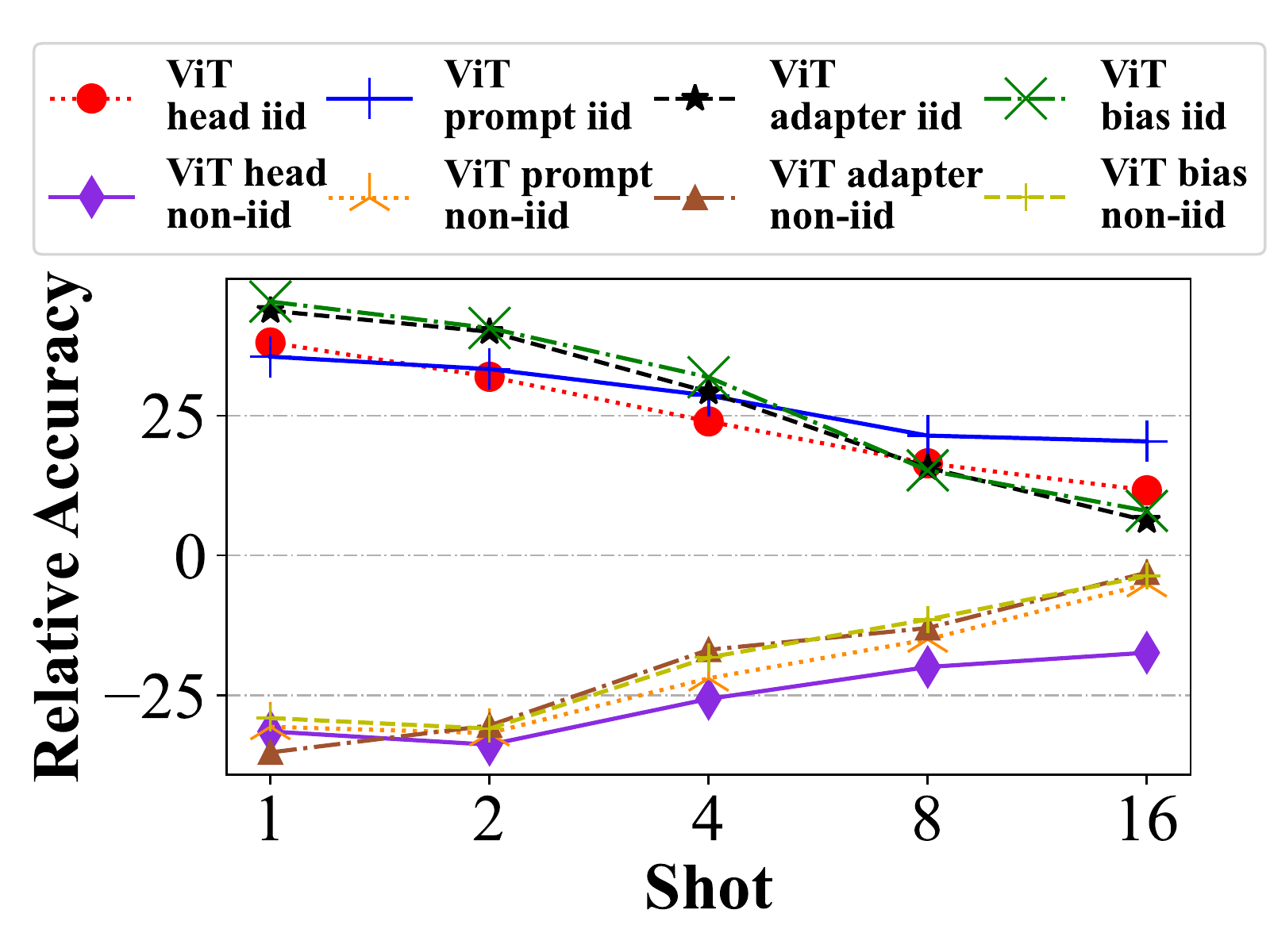}
    \caption{ViT model.}
    \label{fig:local_vit}
  \end{subfigure}
  \caption{Relative accuracy comparison of FL and local training for different shots.}
  \label{fig:local}
\end{figure}

\begin{figure}[htp]
  \centering
  \begin{subfigure}{0.49\linewidth}
    \includegraphics[width=1\textwidth]{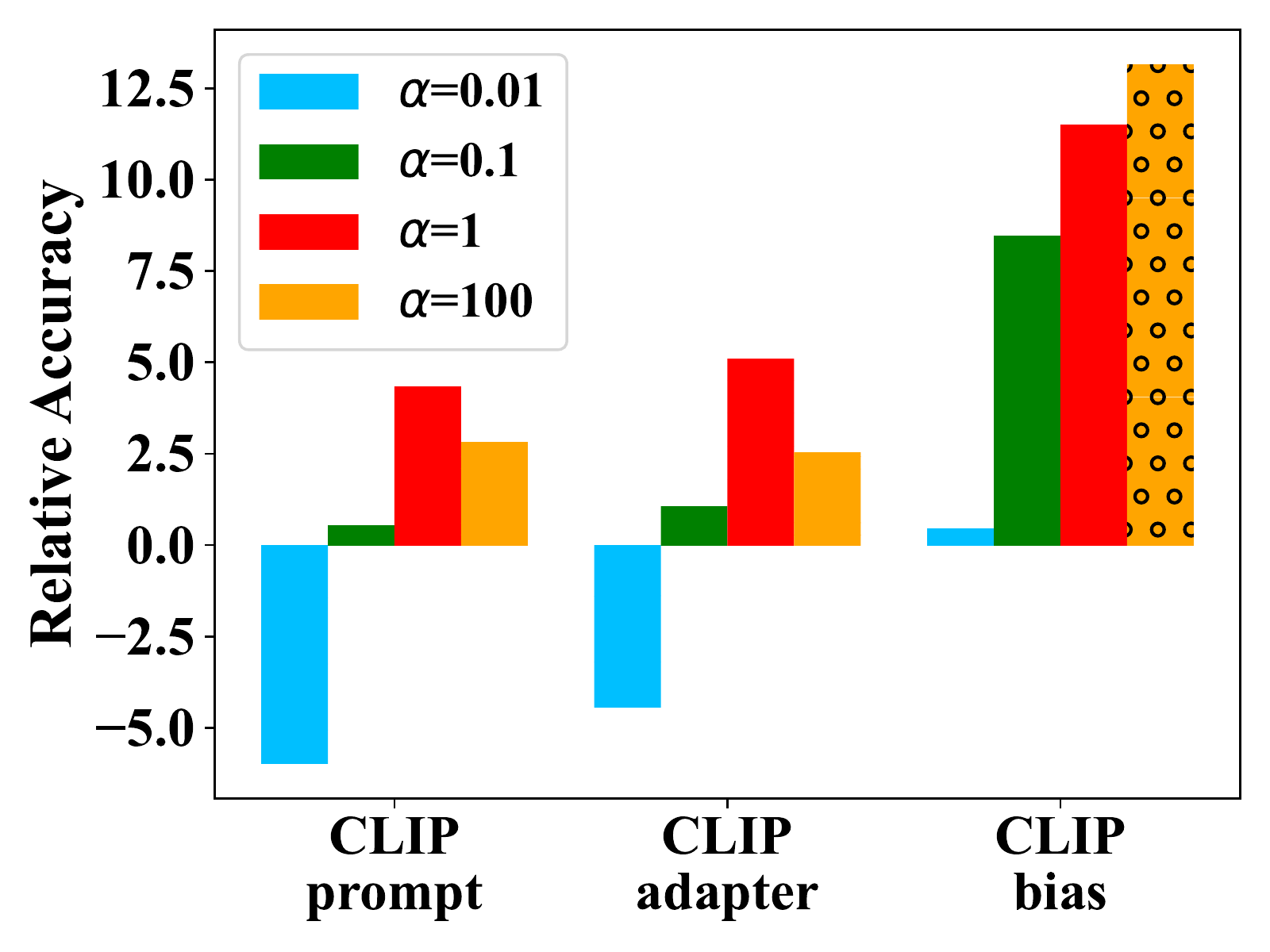}
    \caption{CLIP model.}
    \label{fig:local_clip_dir}
  \end{subfigure}
  \hfill
  \begin{subfigure}{0.49\linewidth}
    \includegraphics[width=1\textwidth]{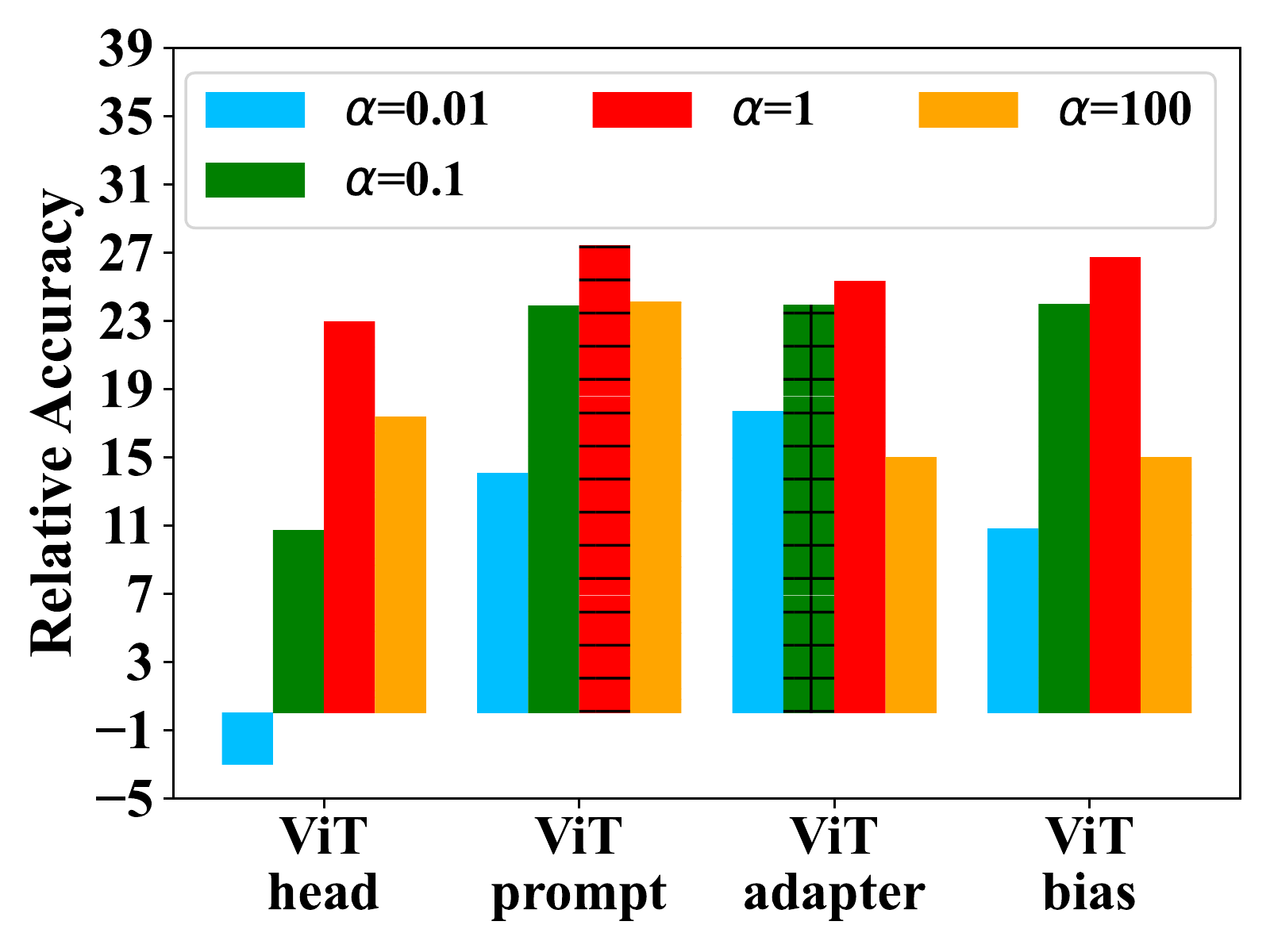}
    \caption{ViT model.}
    \label{fig:local_vit_dir}
  \end{subfigure}
  \caption{Relative accuracy comparison of FL and local training for different Dirichlet distributions.}
  \label{fig:local_dir}
\end{figure}

Previous works showed that fine-tuning methods with foundation models can achieve desirable performance in few-shot learning because the pre-trained model has been encoded into rich information and can achieve excellent transfer learning capabilities. In this section, we explore whether FL is still necessary in the age of foundation models. 

We define the relative accuracy as the accuracy of the FedAVG algorithm minus the accuracy of pure local training. From Fig.~\ref{fig:local}, it's observed that all fine-tuning methods on FedAVG can outperform pure local training in the IID setting. However, local training has a better performance in the non-IID setting. There are two main reasons for this phenomenon: 1) In the extreme non-IID setting, users may have completely different local distributions, and collaborative training has few benefits. 2) Foundation model with local fine-tuning behaves well even when data is scarce. In the non-IID setting, FedAVG has the worst performance in the 1-shot and 2-shot learning because data scarcity can further exacerbate distribution heterogeneity.

We further explore the relative accuracy with different Dirichlet distributions. In Fig.~\ref{fig:local_dir}, we can see that FedAVG has a better performance compared to local training in most of the scenarios. Local training can acquire desirable results with CLIP prompt, CLIP adapter, and ViT head because the tuned modules of these methods are relatively simple, and FL cannot release its advantages. FedAVG has more benefits for the ViT model than the CLIP model because FL can help ViT train the head module better.

\textbf{Insight 2:} FL is still necessary and helps learn a better global model. However, local training is a strong baseline in the extreme non-IID setting. ViT benefits more from FL because the CLIP model already has a satisfactory performance in the local training. The model aggregation in FL can help aggregate information from clients and alleviate the issue of over-fitting~\cite{wortsman2022robust,wortsman2022model}.

\subsection{Comparison with the CNN Model}

\begin{figure}[t]
  \centering
  \begin{subfigure}{0.49\linewidth}
    \includegraphics[width=1\textwidth]{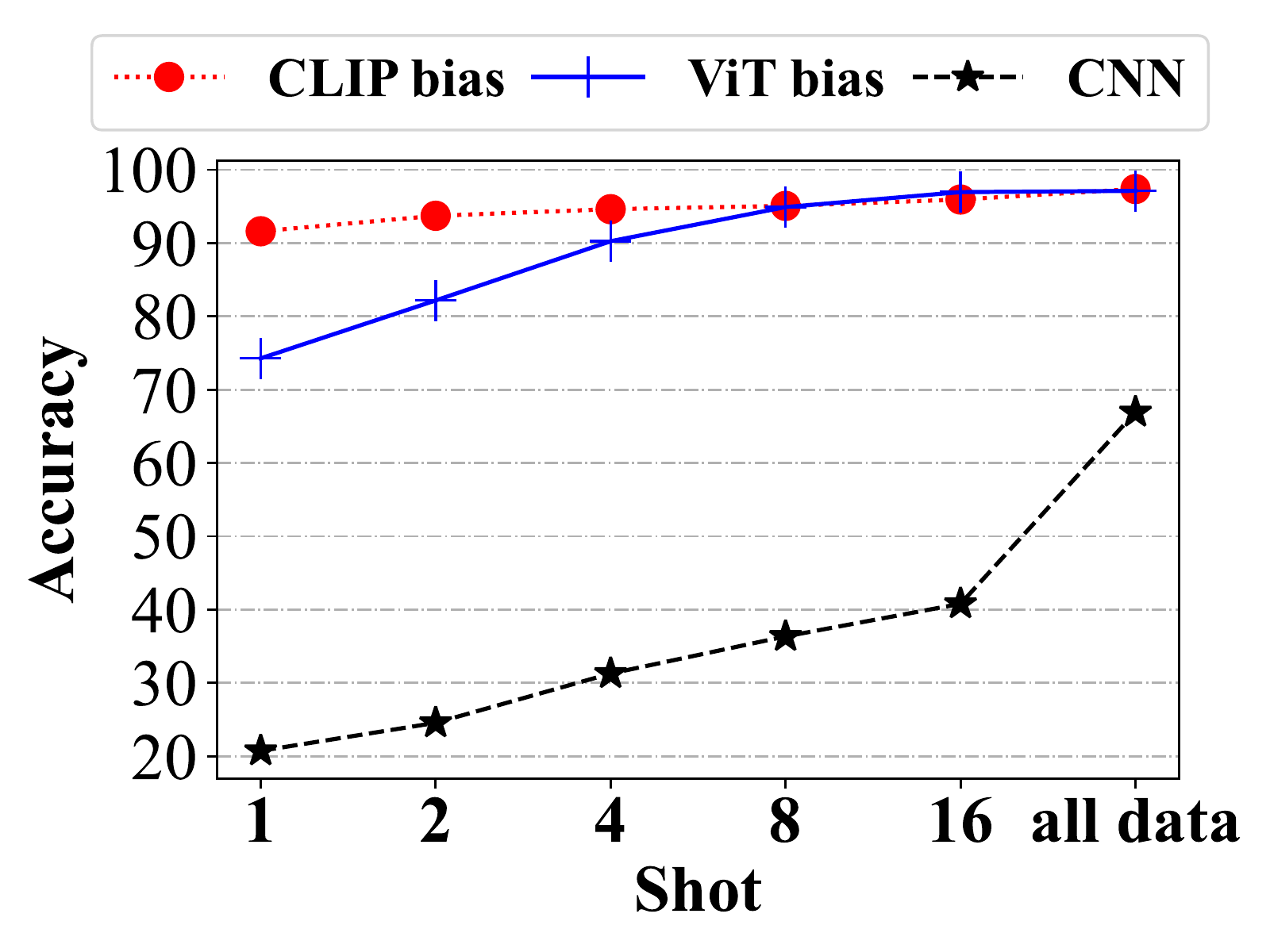}
    \caption{IID setting.}
    \label{fig:cnn_acc_iid}
  \end{subfigure}
  \hfill
  \begin{subfigure}{0.49\linewidth}
    \includegraphics[width=1\textwidth]{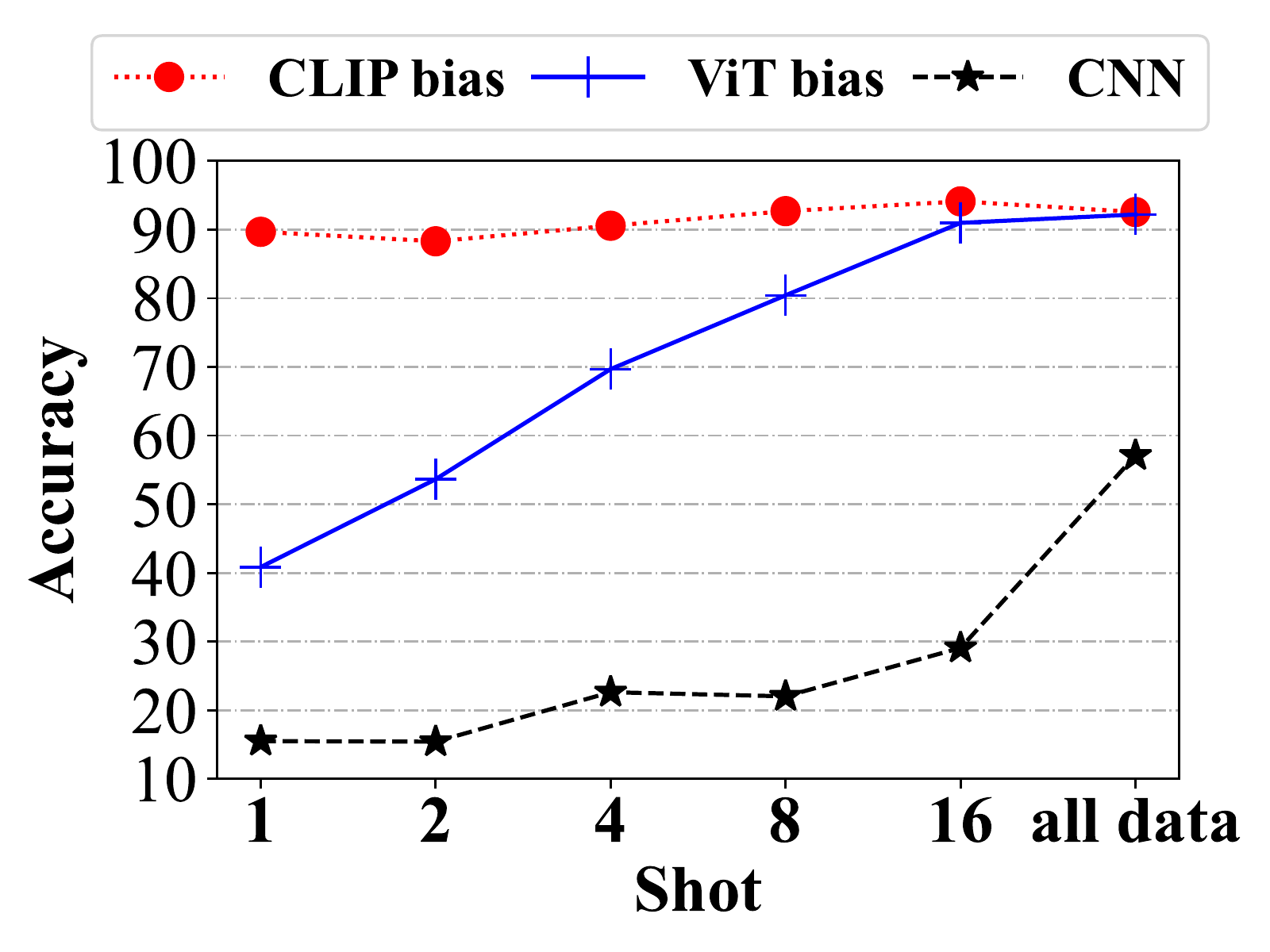}
    \caption{non-IID setting.}
    \label{fig:cnn_acc_noniid}
  \end{subfigure}
  \caption{Accuracy comparison of foundation models and the CNN model.}
  \label{fig:cnn_acc}
\end{figure}

\begin{figure}[t]
  \centering
   \includegraphics[width=0.55\linewidth]{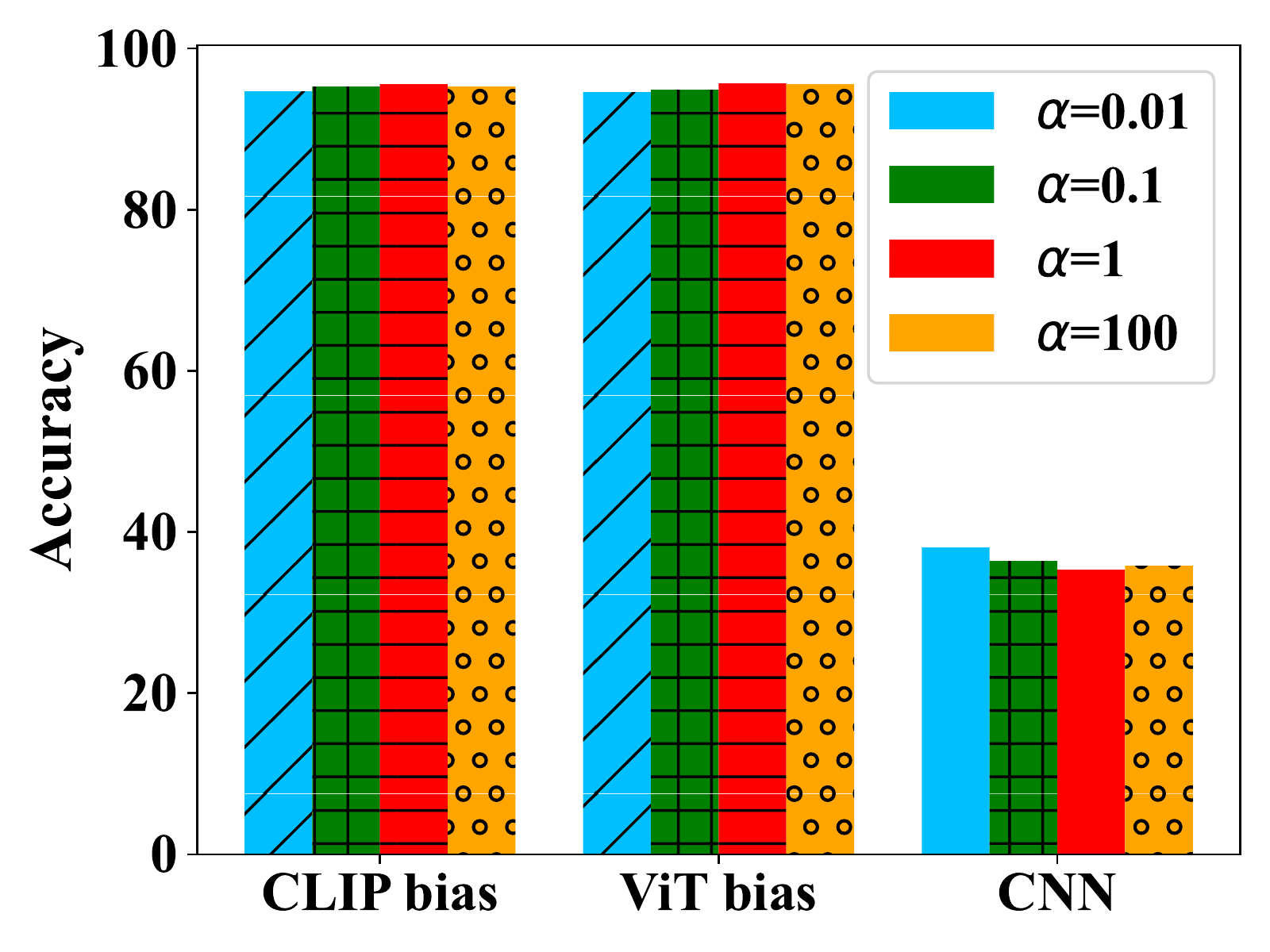}
   \caption{Accuracy comparison of foundation models and the CNN model under different Dirichlet distributions.}
   \label{fig:cnn_dir_acc}
\end{figure}

In this section, we evaluate the performance improvement of pre-trained models in FL compared to the CNN model used in previous works. We use the same CNN architecture as in FedFomo~\cite{zhang2020personalized} and pFedLA~\cite{ma2022layer}. The CNN model is trained from scratch, and the global round is set as 600 epochs. The results are shown in Fig.~\ref{fig:cnn_acc} and Fig.~\ref{fig:cnn_dir_acc}.

In Fig.~\ref{fig:cnn_acc_iid}, the accuracies of CLIP bias and ViT bias are two times higher than that of the CNN model. When trained on the full data, the accuracy of the CNN model can reach 66.93\% and is still much lower than that of the tuning methods. In the non-IID setting, FL with bias tuning has at least 35\% accuracy improvement compared to the vanilla methods. Data capacity has a large impact on the performance of the CNN model, and the accuracies are 20.73\% and 15.51\% in the 1-shot learning. From Fig.~\ref{fig:cnn_dir_acc}, we can see that FL with foundation models can improve at least 55\% accuracy compared to the CNN models under various Dirichlet distributions. 

\textbf{Insight 3:} FL with pre-trained models can significantly improve the accuracy compared to vanilla methods with the CNN model, which suffer from significant performance degradation when data is extremely scarce (e.g., 1-shot learning).

\subsection{Comparison under Various Settings}

In this section, we evaluate the performance of tuning methods with different foundation models and different client numbers.

\subsubsection{Influence of the Foundation Model}

We compare the performance of different variants of the CLIP model and the ViT model. 

\begin{figure}[htp]
  \centering
  \begin{subfigure}{0.49\linewidth}
    \includegraphics[width=1\textwidth]{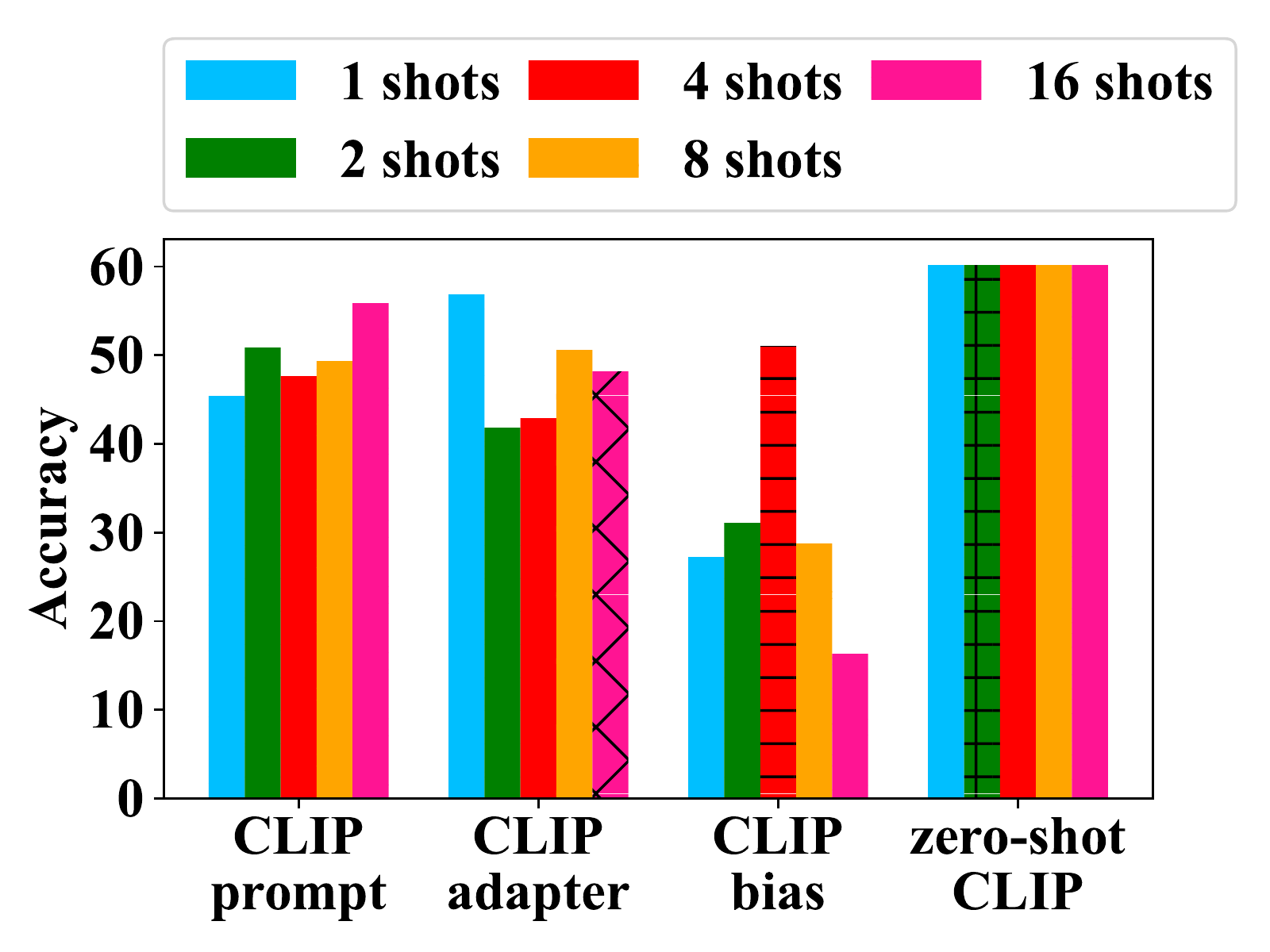}
    \caption{IID setting.}
    \label{fig:clip_res101_iid}
  \end{subfigure}
  \hfill
  \begin{subfigure}{0.49\linewidth}
    \includegraphics[width=1\textwidth]{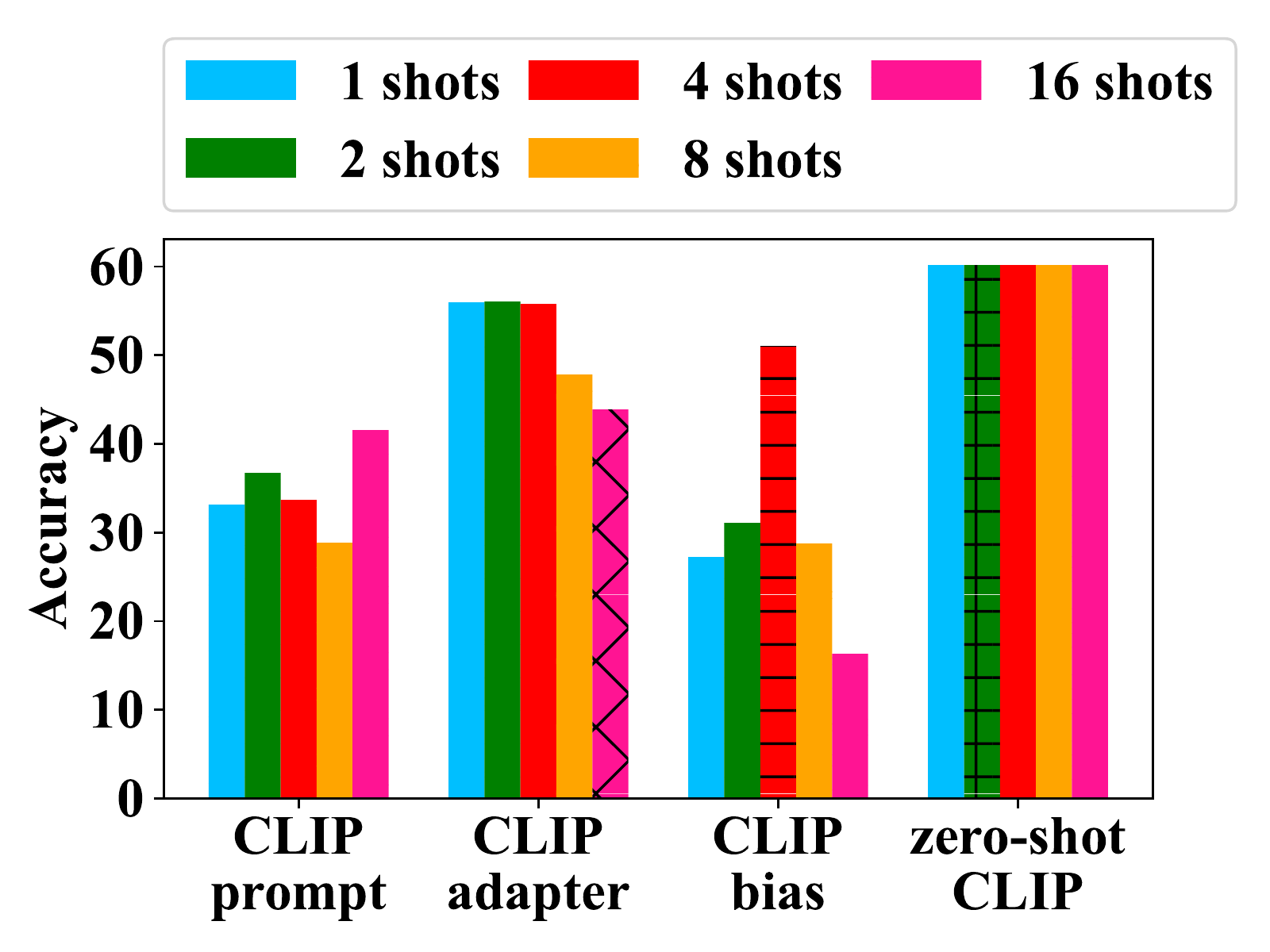}
    \caption{non-IID setting.}
    \label{fig:clip_res101_noniid}
  \end{subfigure}
  \caption{Accuracy comparison of the CLIP ResNet101 model.}
  \label{fig:clip_res101}
\end{figure}

\textbf{Variants of CLIP.} We replace the image encoder of CLIP with a CNN model (i.e., ResNet101) and measure the accuracy of all fine-tuning methods in FL. As shown in Fig.~\ref{fig:clip_res101}, all fine-tuning methods suffer from a considerable accuracy degradation and are worse than zero-shot CLIP. In this scenario, the best method is CLIP adapter with the accuracies of 56.84\% (IID) and 55.9\% (non-IID).

\begin{figure}[htp]
  \centering
  \begin{subfigure}{0.49\linewidth}
    \includegraphics[width=1\textwidth]{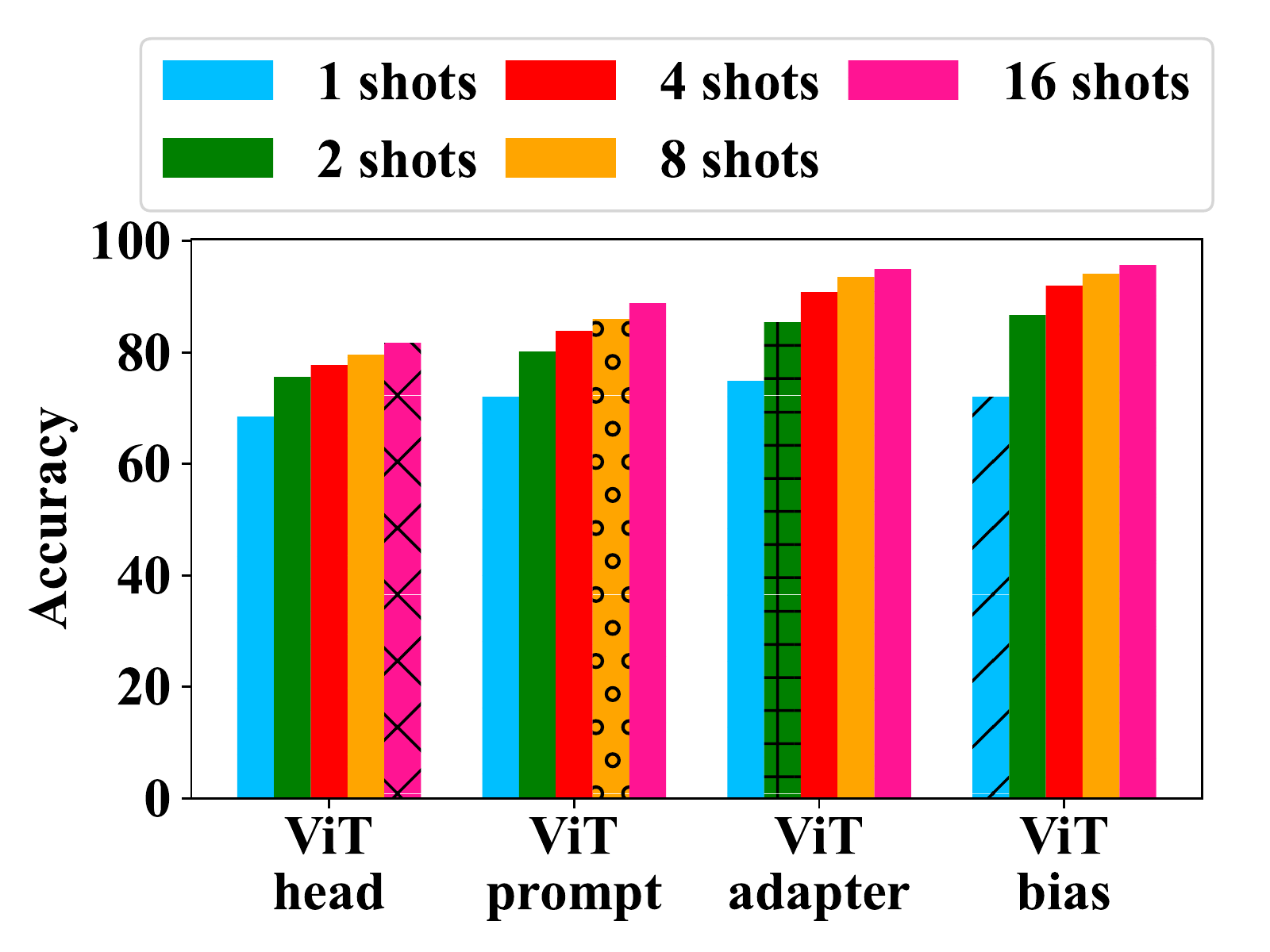}
    \caption{IID setting.}
    \label{fig:vit_small_iid}
  \end{subfigure}
  \hfill
  \begin{subfigure}{0.49\linewidth}
    \includegraphics[width=1\textwidth]{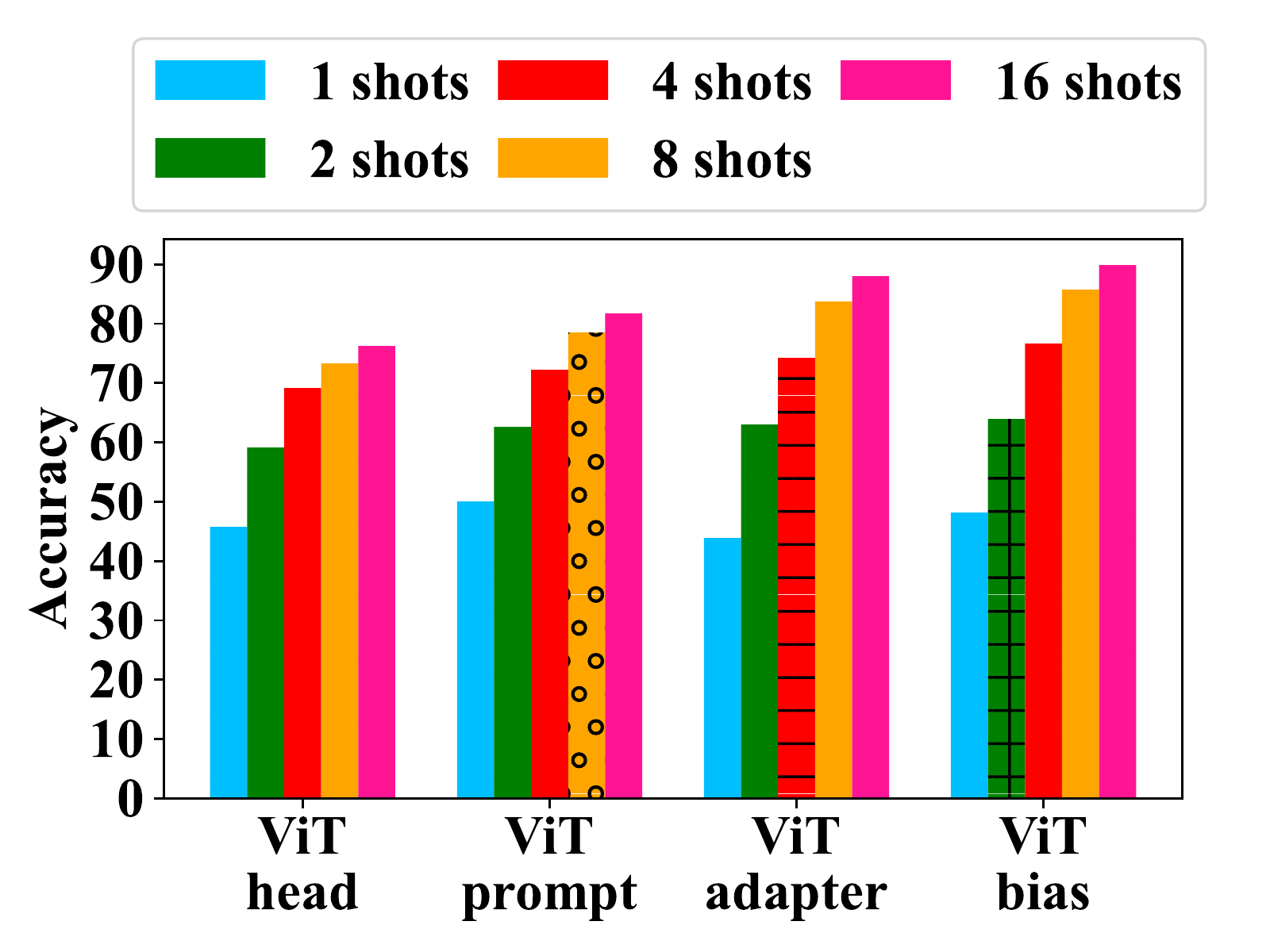}
    \caption{non-IID setting.}
    \label{fig:vit_small_noniid}
  \end{subfigure}
  \caption{Accuracy comparison of the ViT Small model.}
  \label{fig:vit_small}
\end{figure}

\textbf{Variants of ViT.} We replace the ViT Base model with the ViT Small model with 224 $\times$ 224 image size and 16 patches. The accuracy is shown in Fig.~\ref{fig:vit_small}. ViT bias still performs the best, and the average accuracies are 88.048\% (IID) and 89.9\% (non-IID). The performance gap of different tuning methods narrows in this scenario.

\textbf{Insight 4:} A strong visual encoder is essential for the vision-language model because it can learn more knowledge from the pre-training dataset and is robust to the domain shift. The Transformer architecture is more suitable to be the backbone of foundation models in FL~\cite{qu2022rethinking}. Model architecture has less influence on ViT. The bias tuning method relies on a strong pre-trained model.

\subsubsection{Influence of the Number of Clients}

\begin{figure}[htp]
  \centering
  \begin{subfigure}{0.48\linewidth}
    \includegraphics[width=1\textwidth]{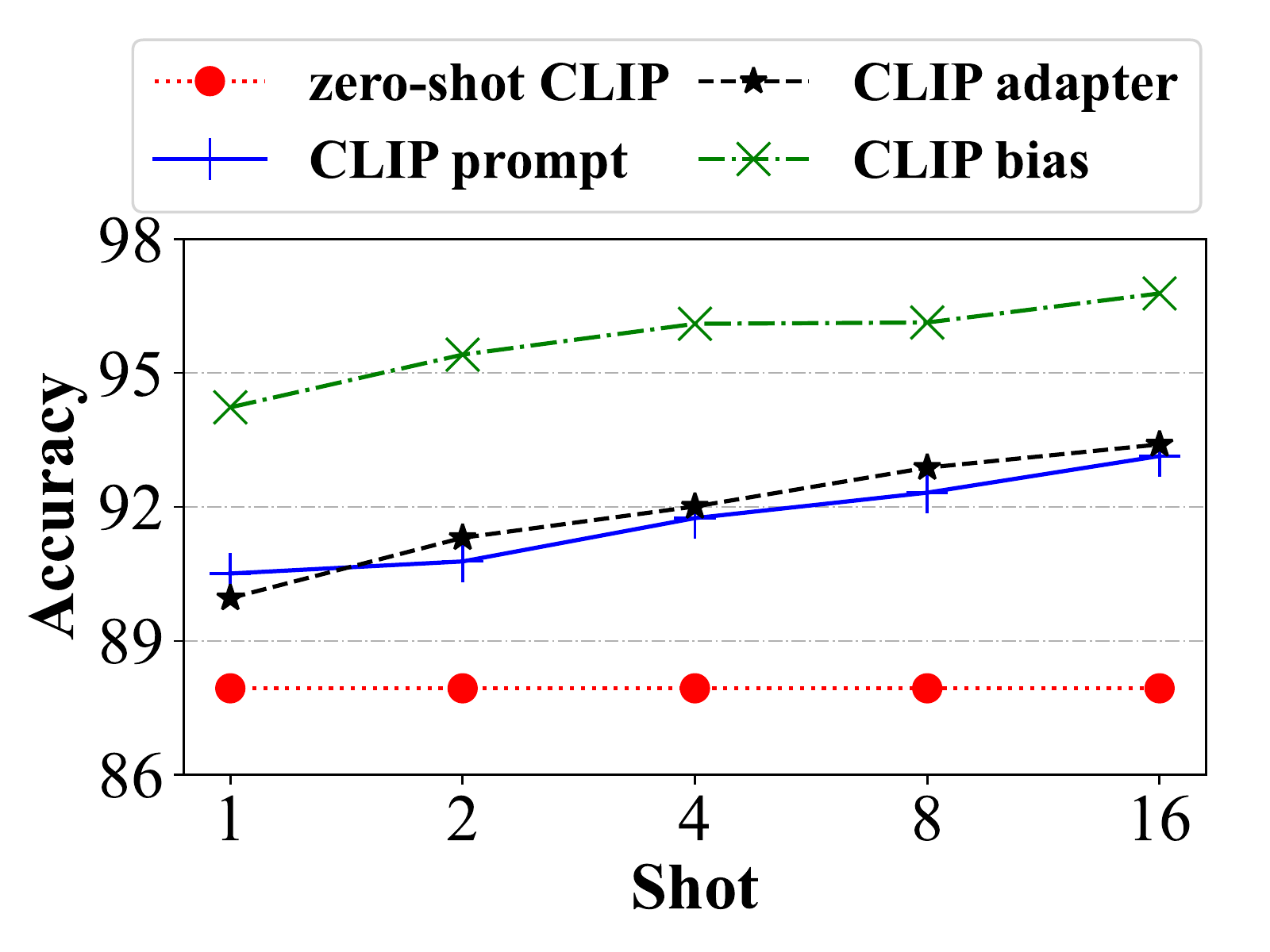}
    \caption{Accuracy comparison of CLIP models in the IID setting.}
    \label{fig:client50_acc-a}
  \end{subfigure}
  \begin{subfigure}{0.48\linewidth}
    \includegraphics[width=1\textwidth]{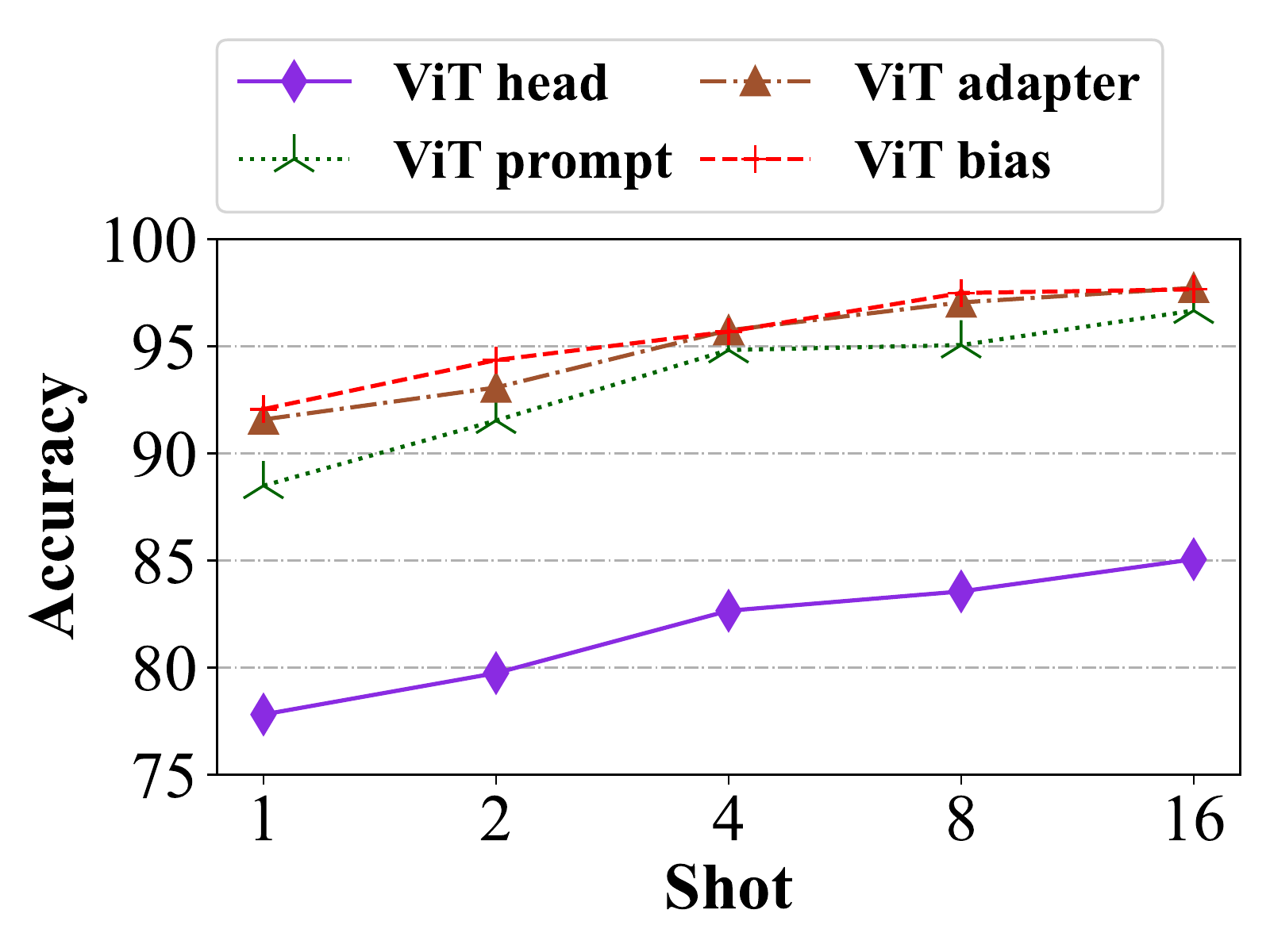}
    \caption{Accuracy comparison of ViT models in the IID setting.}
    \label{fig:client50_acc-b}
  \end{subfigure}
  \label{fig:short}
  \begin{subfigure}{0.48\linewidth}
    \includegraphics[width=1\textwidth]{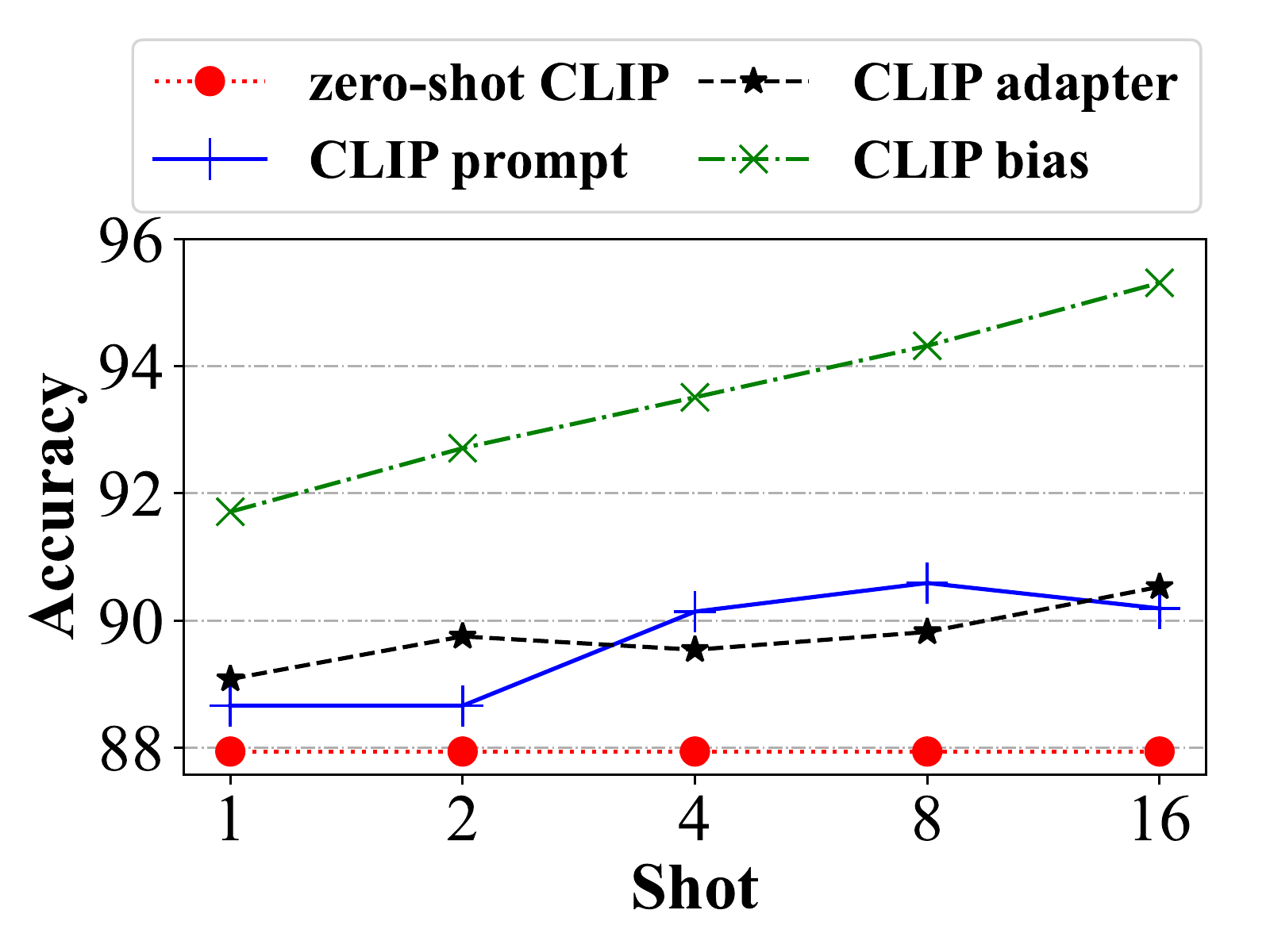}
    \caption{Accuracy comparison of CLIP models in the non-IID setting.}
    \label{fig:client50_acc-c}
  \end{subfigure}
  \begin{subfigure}{0.48\linewidth}
    \includegraphics[width=1\textwidth]{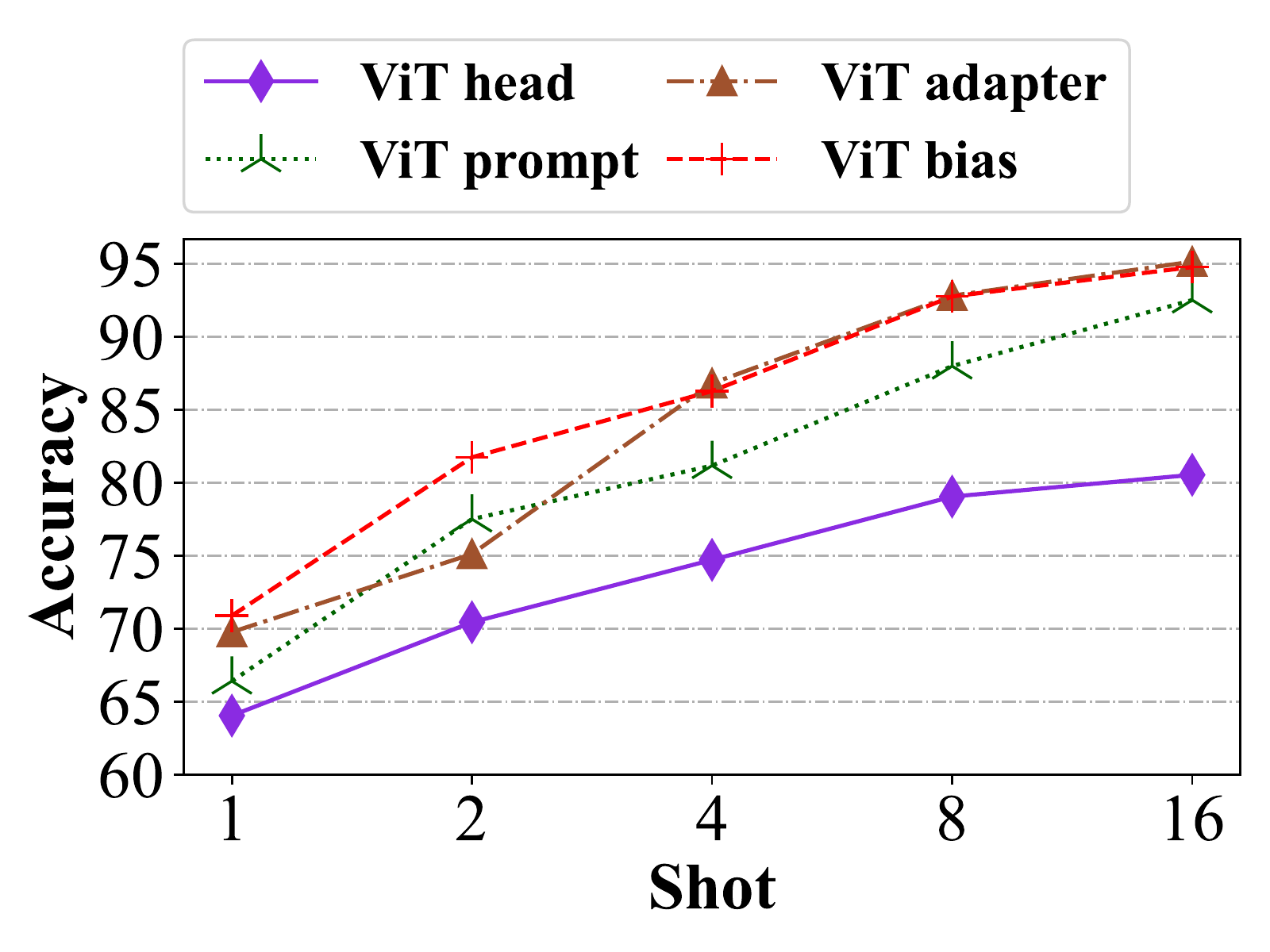}
    \caption{Accuracy comparison of ViT models in the non-IID setting.}
    \label{fig:client50_acc-d}
  \end{subfigure}
  \caption{Accuracy comparison in the IID and non-IID settings when the client number is 50.}
  \label{fig:client50_acc}
\end{figure}

In this section, we investigate the influence of client number. The client number is set as 50, and the sampling rate is 0.2. The global communication round is increased to 100. 

As shown in Fig.~\ref{fig:client50_acc}, CLIP bias and ViT bias still perform best. The average accuracies of CLIP bias are higher than 95\% in the IID setting and higher than 92\% in the non-IID setting.
The performance of algorithms in this scenario is consistent with previous measurements. 

\textbf{Insight 5:} When the number of clients is 50, CLIP bias and ViT bias still behave the best. Client number has little impact on the performance.

\subsection{Resource Cost}

\begin{table}[b]
  \centering
  \begin{tabular}{*{4}{c}}
      \toprule
      \multirow{2}*{Method} & \multicolumn{3}{c}{CLIP} \\
      \cmidrule(lr){2-4}
      & prompt & adapter & bias \\
      \midrule
      Size (KB) & 17.3 & 526 & 459.7 \\
      \bottomrule
    \end{tabular}
  \\[6pt]
  \begin{tabular}{*{5}{c}}
      \toprule
      \multirow{2}*{Method} & \multicolumn{4}{c}{ViT} \\
      \cmidrule(lr){2-5}
      & head & prompt & adapter & bias \\
      \midrule
      Size (KB) & 31.8 & 81.2 & 7166.5 & 466.8 \\
      \bottomrule
    \end{tabular}
  \caption{The sizes of tuning parameters.}
  \label{tab:finetunesize}
\end{table}

\begin{table*}
  \centering
    \begin{tabular}{*{8}{c}}
      \toprule
      \multirow{2}*{Method} & \multicolumn{3}{c}{CLIP} & \multicolumn{4}{c}{ViT}\\
      \cmidrule(lr){2-4}\cmidrule(lr){5-8}
      & prompt & adapter & bias & head & prompt & adapter & bias \\
      \midrule
      Round/IID  & 3 & 3 & 5 & 11 & 6 &2 &2 \\
      Size/IID (MB) & 1.038 & 31.56 & 45.97 & 6.996 & 9.744 & 429.99 & 28.008\\
      Round/non-IID & 5 &3 &4 &12 &13 &6 &8 \\
      Size/non-IID (MB) &1.73 &32.56 &36.776 &7.632 & 21.112 &859.98 & 74.688 \\
      \bottomrule
    \end{tabular}
    \caption{The total communication sizes in the 16-shot learning.}
  \label{tab:communication}
\end{table*}

\begin{table}[htp]
  \centering
  \begin{tabular}{*{5}{c}}
      \toprule
      \multirow{2}*{Method} & \multicolumn{2}{c}{CLIP} & \multicolumn{2}{c}{ViT} \\
      \cmidrule(lr){2-3}\cmidrule(lr){4-5}
      & ViT16 & ResNet101 & Small & Base \\
      \midrule
      Size (MB) & 335  & 279 & 115 & 392 \\
      \bottomrule
    \end{tabular}
  \caption{The sizes of foundation models.}
  \label{tab:pretrainmodelsize}
\end{table}

Resource cost is also an important factor of FL, and we measure the model sizes of all fine-tuning methods. As shown in Table.~\ref{tab:finetunesize}, all tuning methods are lightweight and can be deployed on local devices.
It's observed that the size of CLIP prompt is extremely small and is less than 0.1\% of the overall backbone, so prompt learning is quite efficient and effective. The size of CLIP adapter is 526KB and is slightly higher than that of the CLIP bias. The size of ViT adapter is the largest because it inserts some learnable modules into each transformer layer.

We further calculate the number of communication rounds needed to reach the convergence condition. The convergence condition is defined as that the global model achieves a pre-designated training accuracy of 99\% or the difference in the training accuracy between two communication rounds is smaller than 0.5\%. We define the communication round as $r$, the number of clients as $n$, and the model size as $s$. Then, the communication cost $c$ can be calculated by: $c=r \times n \times s \times 2$. The cost in the 16-shot learning is shown in Table.~\ref{tab:communication}. We can find that all tuning methods can converge within 15 global rounds, and the total communication size is acceptable under current network conditions. CLIP prompt has a minimal communication cost of 1.038MB. Clients need more global rounds for convergence in the non-IID setting.

We assume that the pre-trained models are pre-stored in the local devices. Table.~\ref{tab:pretrainmodelsize} shows the sizes of different foundation models. The sizes of the CLIP-ViT16 model and the ViT Base model are 335MB and 392MB, which don't consume too much storage. 

\textbf{Insight 6:} The tuning methods are lightweight and can be trained on local devices. Prompt learning is the most efficient approach with few accuracy losses. The foundation model can accelerate the convergence rate in FL and save communication resources. 


\section{In Conjunction with Existing Works}

\begin{figure}[htp]
  \centering
   \includegraphics[width=0.8\linewidth]{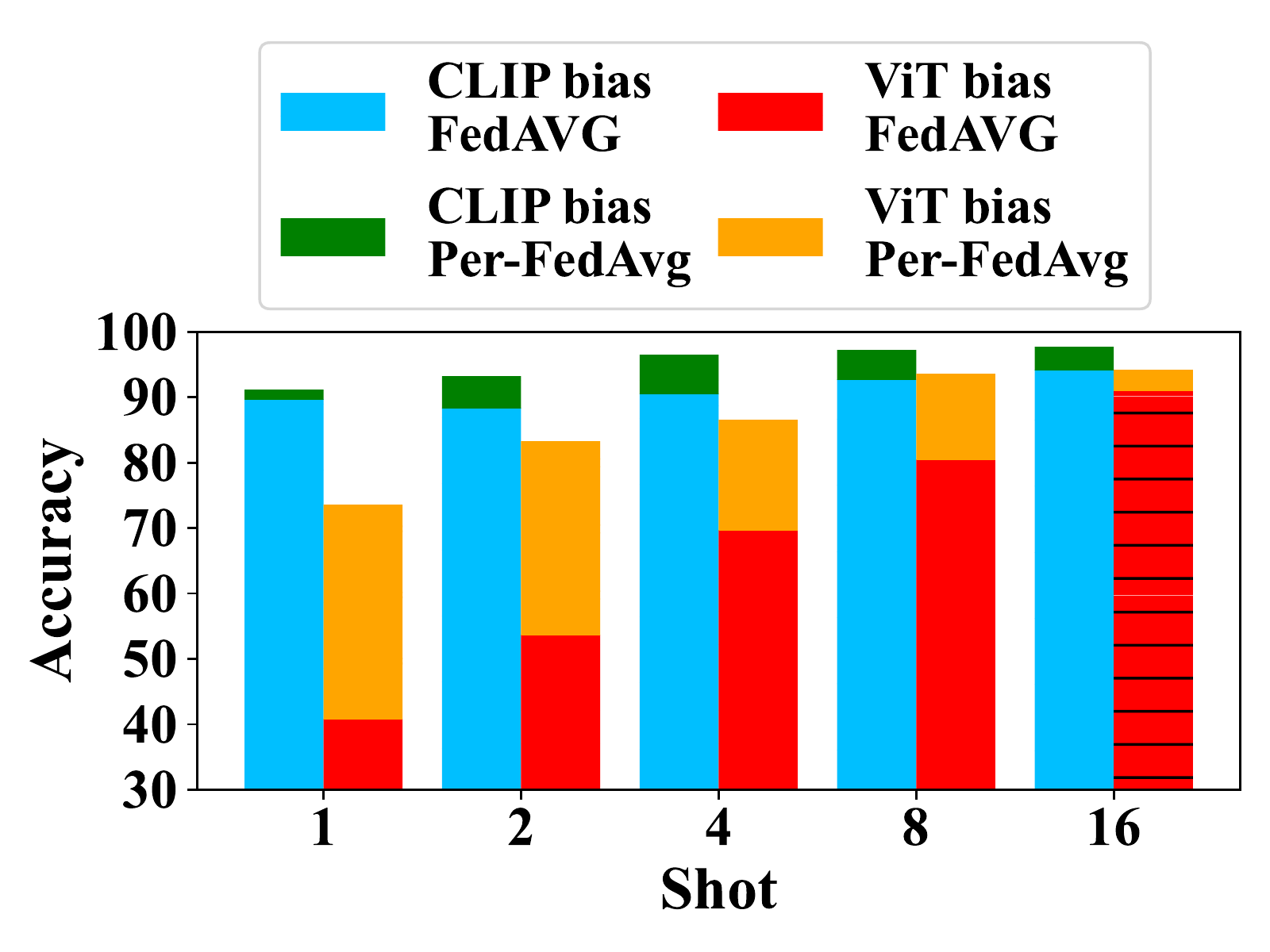}
   \caption{The accuracy of tuning methods with Per-FedAvg~\cite{fallah2020personalized} under non-IID setting.}
   \label{fig:perfedavg_acc}
\end{figure}

    Our efficient federated fine-tuning methods are compatible with existing FL research, and current approaches can be applied to this new framework. In this section, we modify the federated fine-tuning framework with a classical FL method, called Per-FedAvg~\cite{fallah2020personalized}, to fully utilize their advantages. Per-FedAvg is a personalized FL method that uses meta-learning to collaboratively train a global model which can better adapt to downstream tasks. We add 25 local rounds after training the global model. The accuracies in the non-IID setting are shown in Fig.~\ref{fig:perfedavg_acc}. 
    
    Per-FedAvg achieves better performance compared to FedAVG. CLIP bias in the 16-shot learning has the highest accuracy of 97.68\% and has 6.7\% accuracy improvement on FedAVG. Per-FedAvg is more effective on ViT bias and can improve 32.65\% accuracy in the 1-shot learning. Results reveal that personalized methods can cope with distribution shifts and improve the performance of federated fine-tuning.

\section{Conclusion}

In this paper, we propose a novel and unified framework of parameter-efficient fine-tuning with foundation models in FL. Then, we conduct an in-depth measurement to unveil the performance of these methods. Our experimental results show that foundation models can help FL train a better global model and converge faster. Bias tuning has the best performance, and FL is still necessary for collaboratively fine-tuning. These tuning methods are quite lightweight and can reduce resource costs.
We hope this paper can encourage more explorations of foundation models and FL.

{\small
\bibliographystyle{ieee_fullname}
\bibliography{egbib}
}
\clearpage

\begin{appendices}

\section{Problem Formulation}

In this section, we provide formal problem formulation of federated fine-tuning with the pre-trained model. We use $c$ to represent a client and assume there are $n$ users in the FL system. Each user can only access his/her private dataset $D_c:= \{x_i,y_i\}$, where $x_i$ and $y_i$ are the $i$-th input (i.e., image) and its corresponding label. The client's model $h$ is divided into two parts: the frozen pre-trained backbone with parameter $\theta$ and the lightweight trainable module with parameter $w$. Therefore, the output of a sample $x_i$ can be represented as $p(x_i)=h(x_i;w,\theta)$. 

Clients train their local models with the cross-entropy loss $L_{CE}$ and the average loss value of client $c$ can be calculated by:
\begin{equation}
  L_c(w) = \frac{1}{|D_c|} \sum_{i=1}^{|D_c|} L_{CE}(w;p(x_i),y_i).
  \label{eq:loss_client}
\end{equation}

Therefore, the overall objective of the FL system can be defined as:
\begin{equation}
  \mathop{\min}_{w} L(w) = \sum_{c=1}^{n} \frac{|D_c|}{|D|} L_c(w).
  \label{eq:loss_system}
\end{equation}

\section{The Pseudo-code of Federated Fine-tuning}

In this section, we provide the pseudo-code of federated fine-tuning with the pre-trained model in Algorithm.~\ref{alg}. At each global round, a fraction of users are randomly selected as active workers, and the global model is distributed to these users. Each worker parallelly fine-tunes the pre-trained model with local data with a few epochs. Then, they extract the gradients of the fine-tuned modules and send them back to the server, which aggregates the updates with weighted averages.

\begin{algorithm}
    \renewcommand{\algorithmicrequire}{\textbf{Input:}}
	\renewcommand{\algorithmicensure}{\textbf{Output:}}
    \caption{The procedure of federated fine-tuning with the pre-train model with FedAVG.} 
	\label{alg} 
    \begin{algorithmic}[1]
    \Require $n$ clients indexed by $c$; global epoch $T$; local epoch $E$; pre-trained model $\theta$; learning rate $\eta$;
    \Ensure a fine-tuned global model $w$;
    \Function {Server execute:}{}
        \State Initialize $w_0$;
        \For{each global round $t=1 \sim T$}
        \State Select active workers $S_t$;
        \State Distribute the global model $w_t$;
        \For{$c \in S_t$ \textbf{in parallel}}
        \State $\Delta w_{t+1}^{c} \gets$ CLIENTUPDATE($c,w_t$); 
        \EndFor
        \State $w_{t+1} \gets w_{t} + \sum_{c=1}^{n}\frac{|D_c|}{|D|} \Delta w_{t+1}^{c}$;
        \EndFor
        \State \Return $w_{T+1}$;
    \EndFunction
    \Function {ClientUpdate:}{$c,w$}
        \State Load the pre-trained model $\theta$;
        \State Load the fine-tuned module $w$;
        \For{each local epoch $e=1 \sim E$}
        \For{a batch of data}
        \State $w \gets w - \eta \bigtriangledown L_c(w)$;
        \EndFor
        \EndFor
        \State Extract $\Delta w$ and send $\Delta w$ to server;
    \EndFunction
    \end{algorithmic}
\end{algorithm}

\section{Visualization of Data Distributions of Clients}
\label{sec:intro}

\begin{figure}[htp]
  \centering
  \begin{subfigure}{0.48\linewidth}
    \includegraphics[width=1\textwidth]{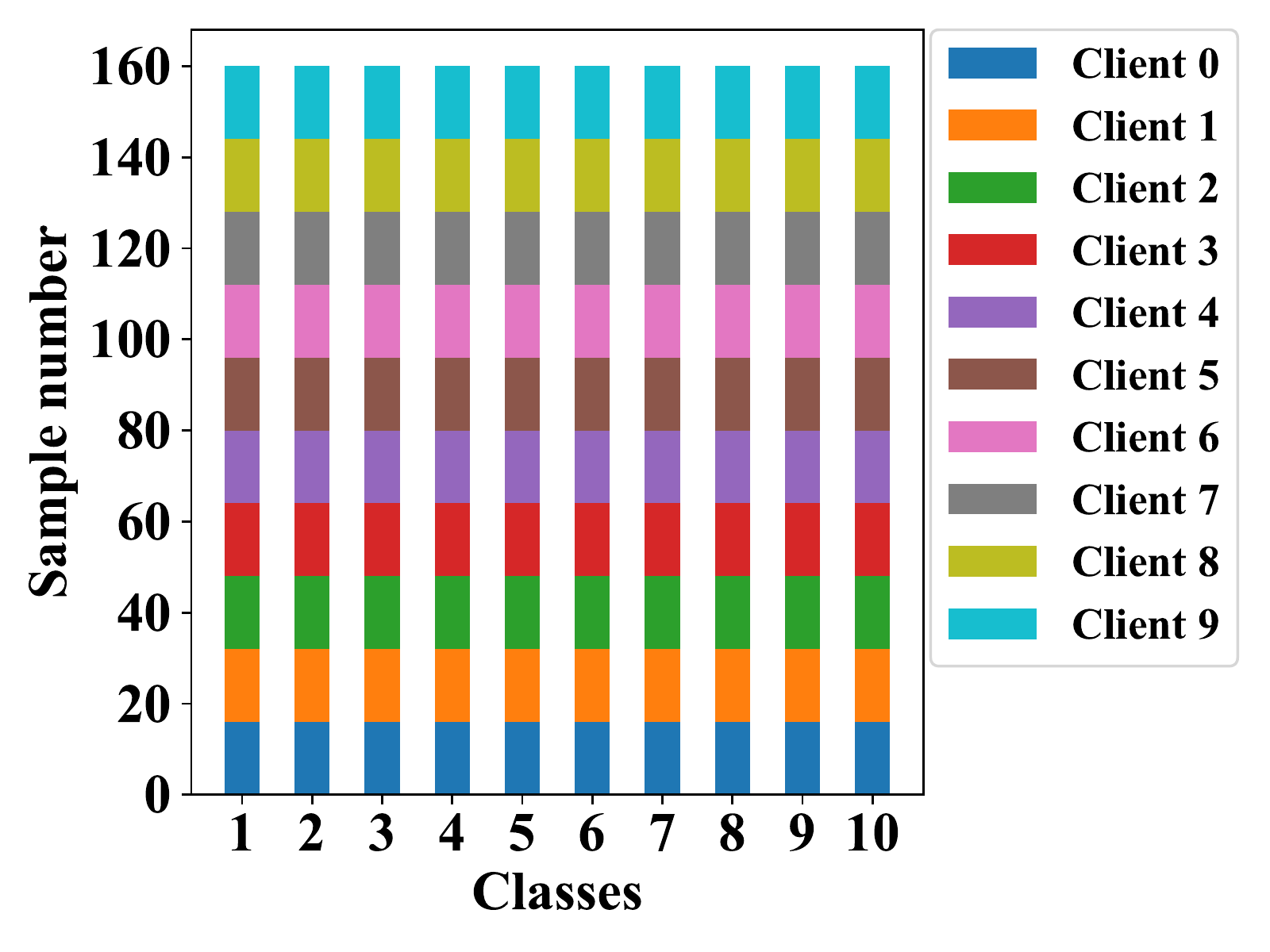}
    \caption{IID setting.}
    \label{fig:dis_iid1}
  \end{subfigure}
  \hfill
  \begin{subfigure}{0.48\linewidth}
    \includegraphics[width=1\textwidth]{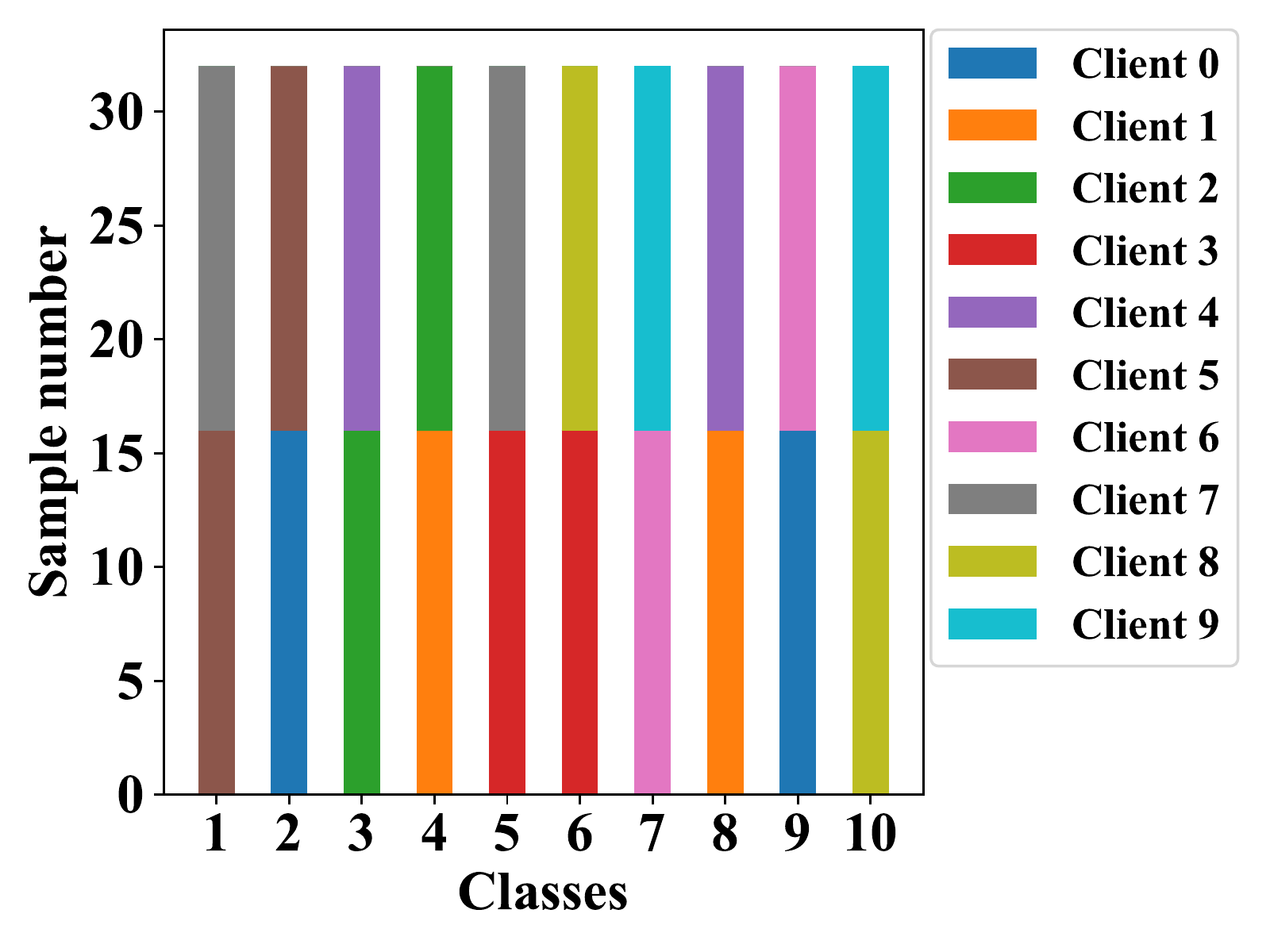}
    \caption{non-IID setting.}
    \label{fig:dis_iid0}
  \end{subfigure}
  \caption{Data distributions of clients in the 16-shot learning.}
  \label{fig:dis_vis}
\end{figure}

\begin{figure}[htp]
  \centering
  \begin{subfigure}{0.48\linewidth}
    \includegraphics[width=1\textwidth]{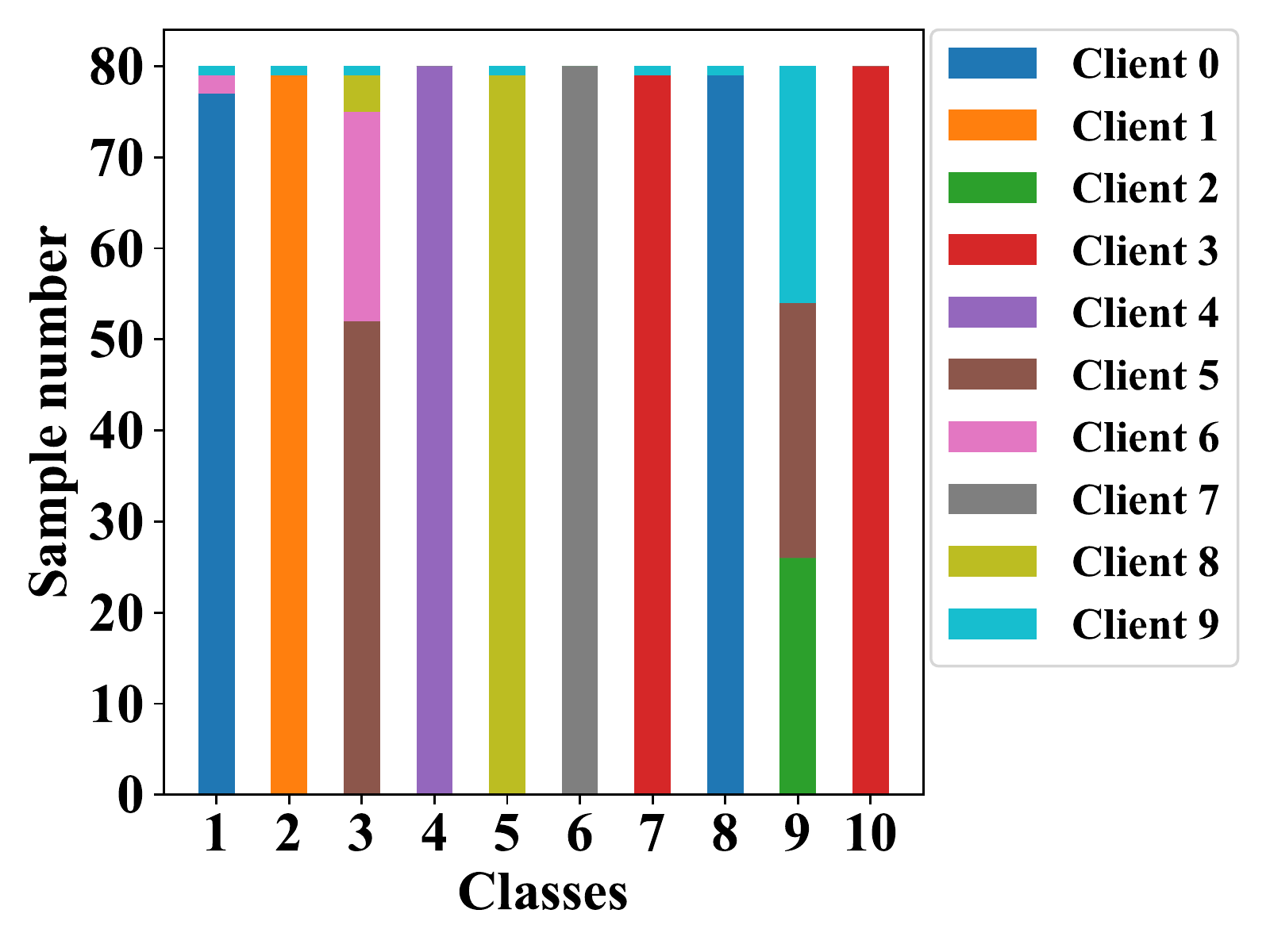}
    \caption{$\alpha=0.01$.}
    \label{fig:dis_dir001}
  \end{subfigure}
  \hfill
  \begin{subfigure}{0.48\linewidth}
    \includegraphics[width=1\textwidth]{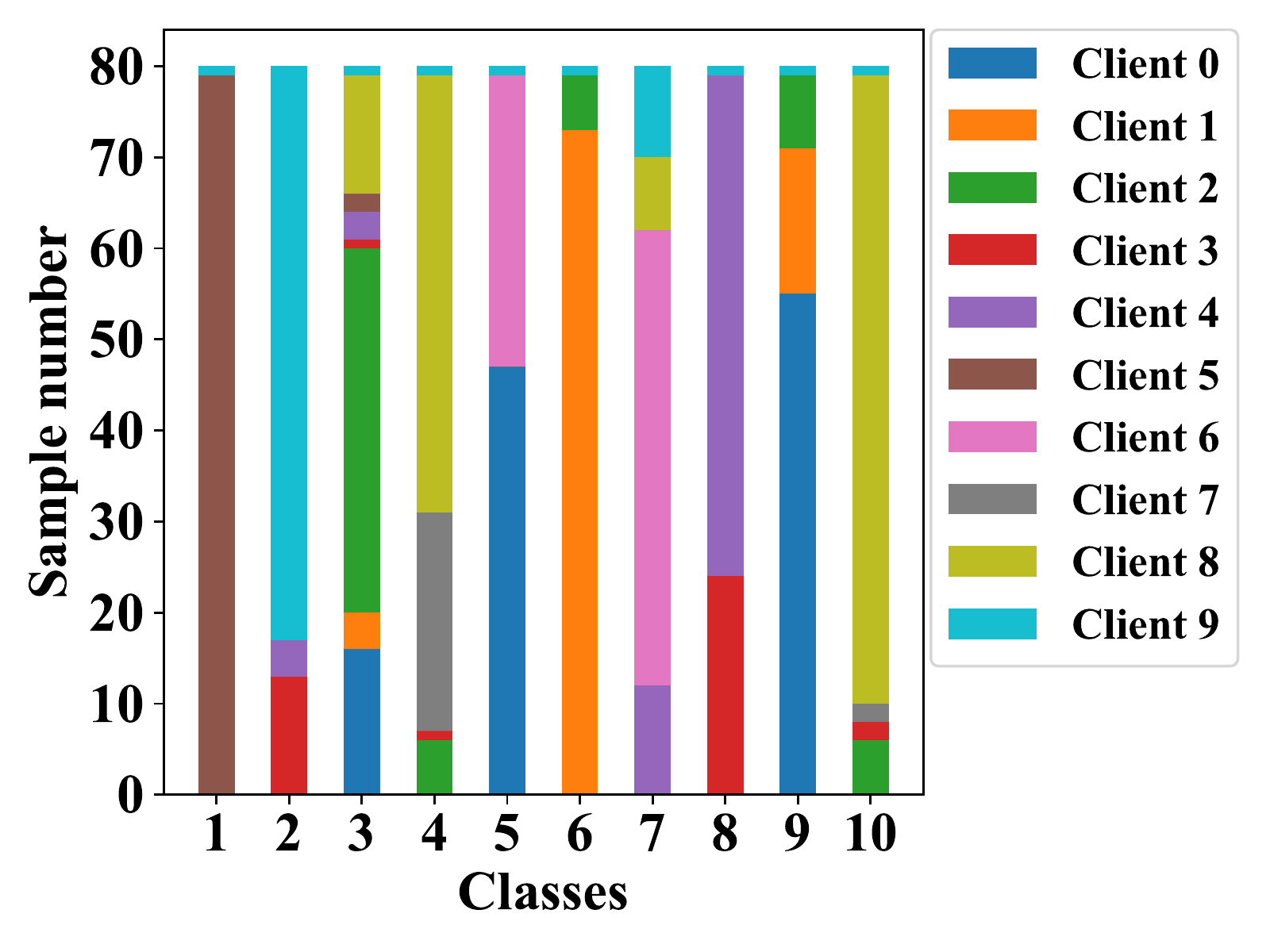}
    \caption{$\alpha=0.1$.}
    \label{fig:dis_dir01}
  \end{subfigure}
  \hfill
  \begin{subfigure}{0.48\linewidth}
    \includegraphics[width=1\textwidth]{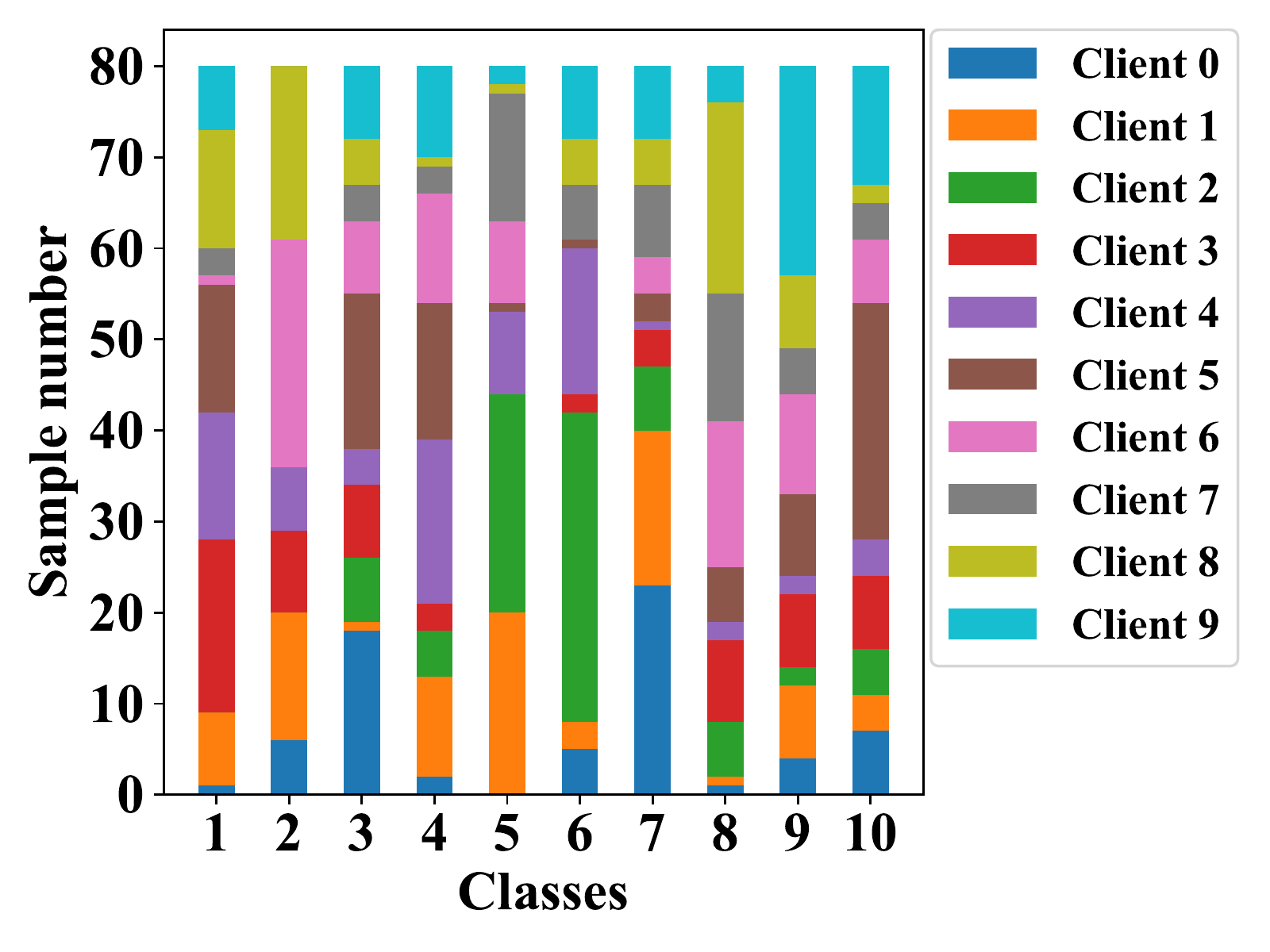}
    \caption{$\alpha=1$.}
    \label{fig:dis_dir1}
  \end{subfigure}
  \hfill
  \begin{subfigure}{0.48\linewidth}
    \includegraphics[width=1\textwidth]{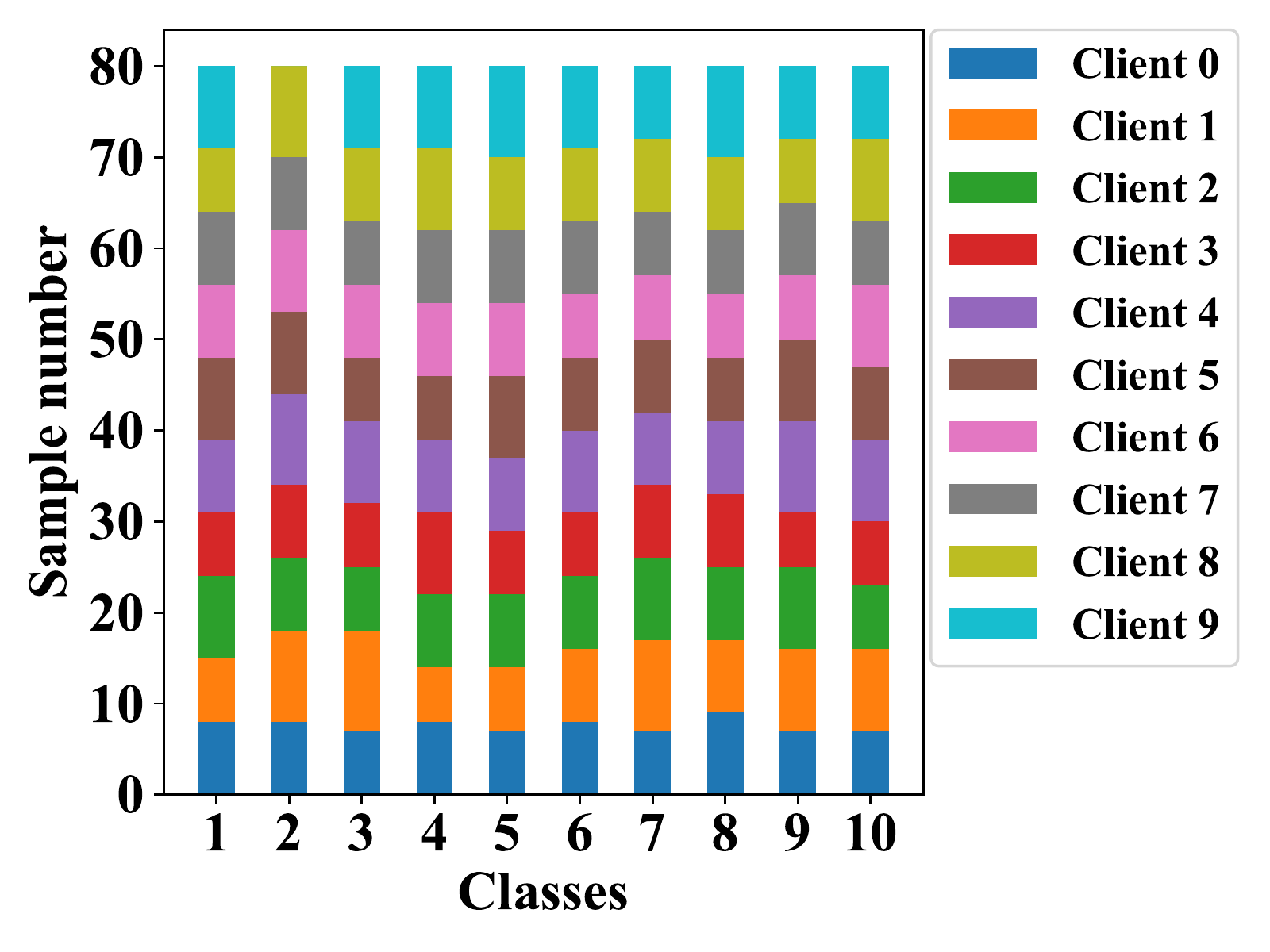}
    \caption{$\alpha=100$.}
    \label{fig:dis_dir100}
  \end{subfigure}
  \caption{Data distributions of clients under different Dirichlet distributions.}
  \label{fig:dis_dir}
\end{figure}

In this section, we visualize the data distributions of clients to understand the distribution heterogeneity. In Fig.~\ref{fig:dis_vis}, users have the same number of samples per class in the IID setting. In contrast, a user owns two different classes in the non-IID setting. The FL algorithm needs to group users who have similar class distributions. In Fig.~\ref{fig:dis_dir}, data distribution changes with the value of $\alpha$. When $\alpha=0.01$, data distributions are highly heterogeneous, and the classes of clients are quite different. When $\alpha=100$, data distribution is similar to that of the IID setting.

\section{Comparison Results}

\begin{figure}[htp]
  \centering
  \begin{subfigure}{0.48\linewidth}
    \includegraphics[width=1\textwidth]{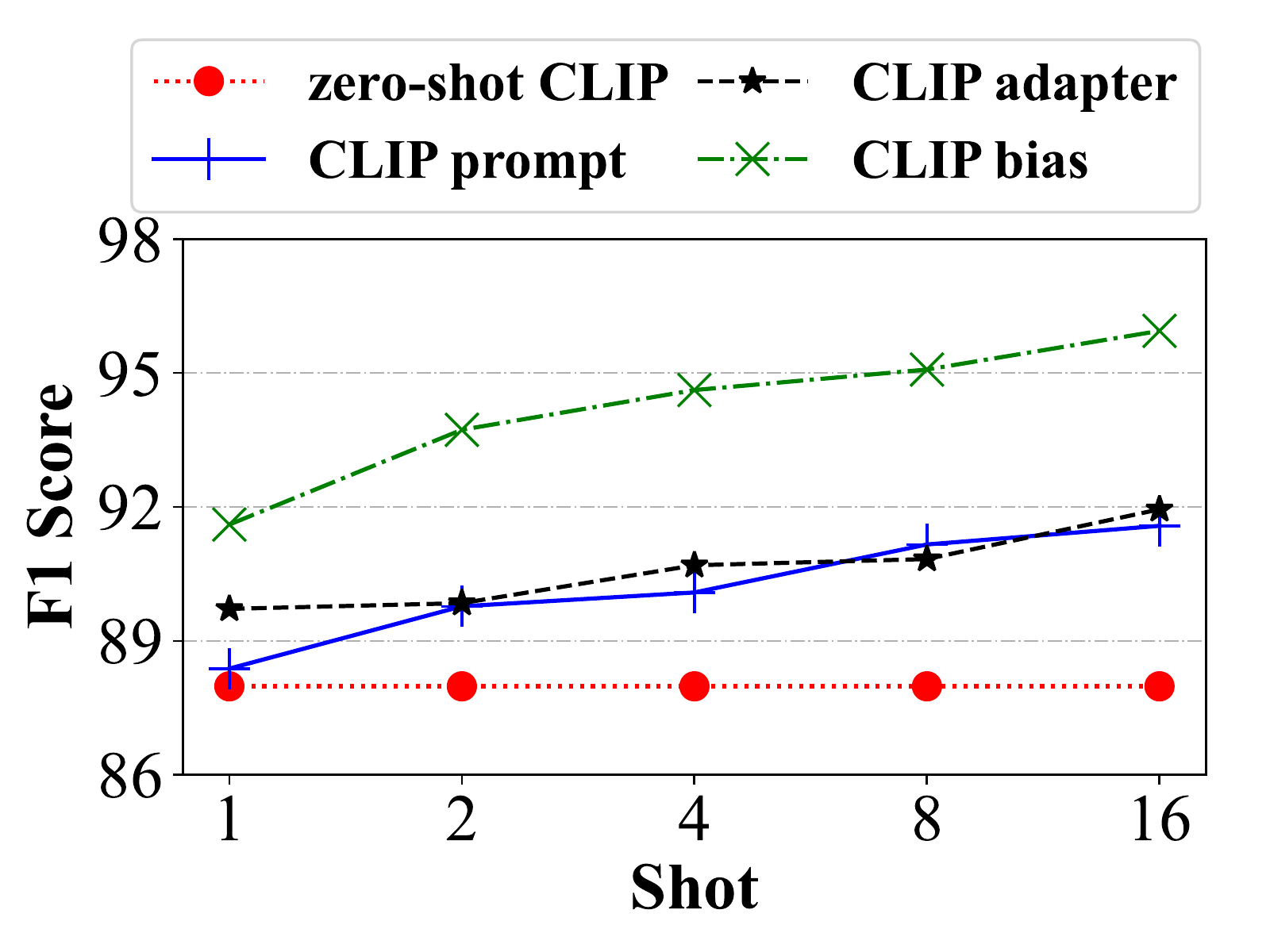}
    \caption{F1 score comparison of CLIP models under IID setting.}
    \label{fig:f1-a}
  \end{subfigure}
  \begin{subfigure}{0.48\linewidth}
    \includegraphics[width=1\textwidth]{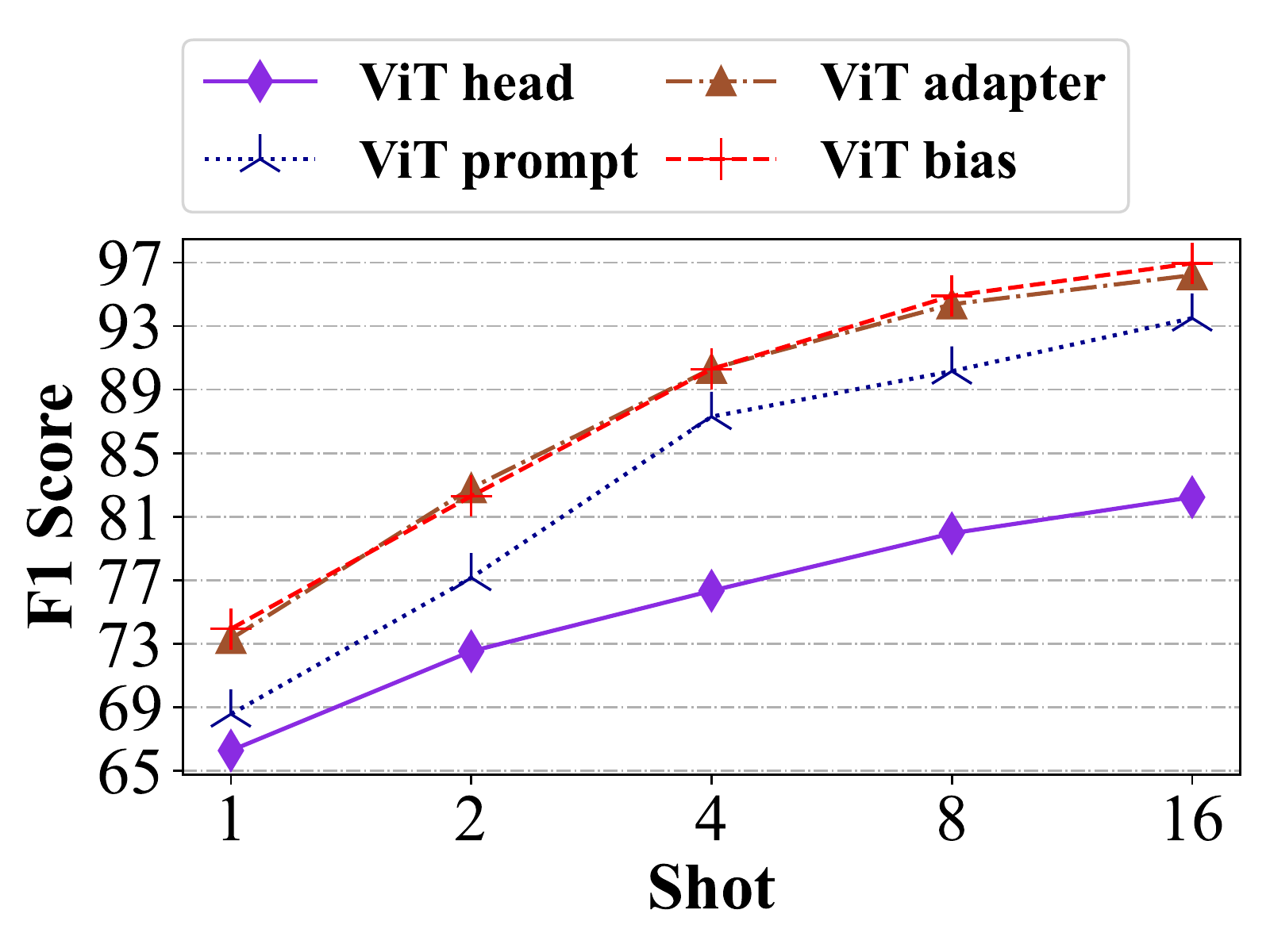}
    \caption{F1 score comparison of ViT models under IID setting.}
    \label{fig:f1-b}
  \end{subfigure}
  \label{fig:short}
  \begin{subfigure}{0.48\linewidth}
    \includegraphics[width=1\textwidth]{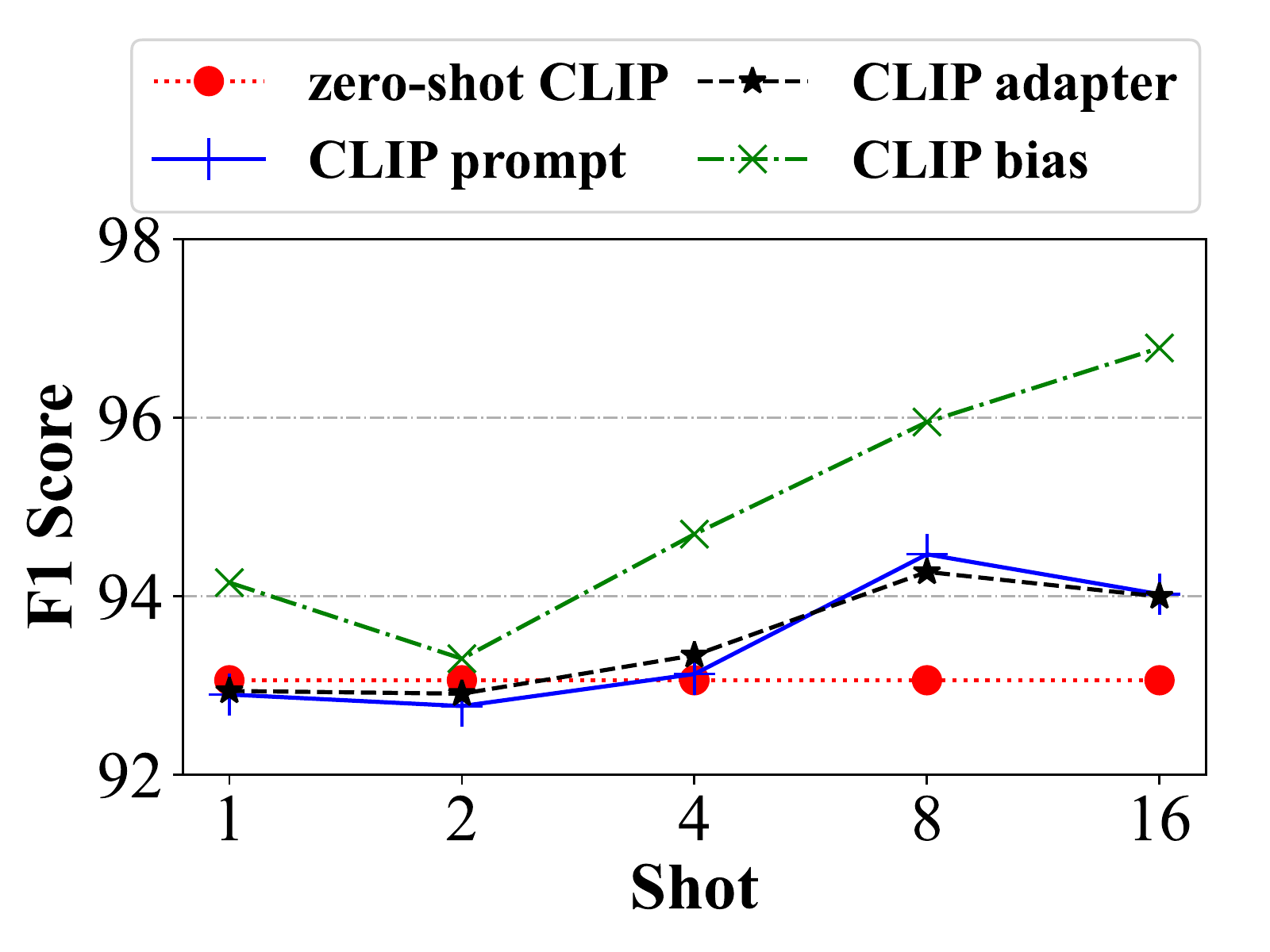}
    \caption{F1 score comparison of CLIP models under non-IID setting.}
    \label{fig:f1-c}
  \end{subfigure}
  \begin{subfigure}{0.48\linewidth}
    \includegraphics[width=1\textwidth]{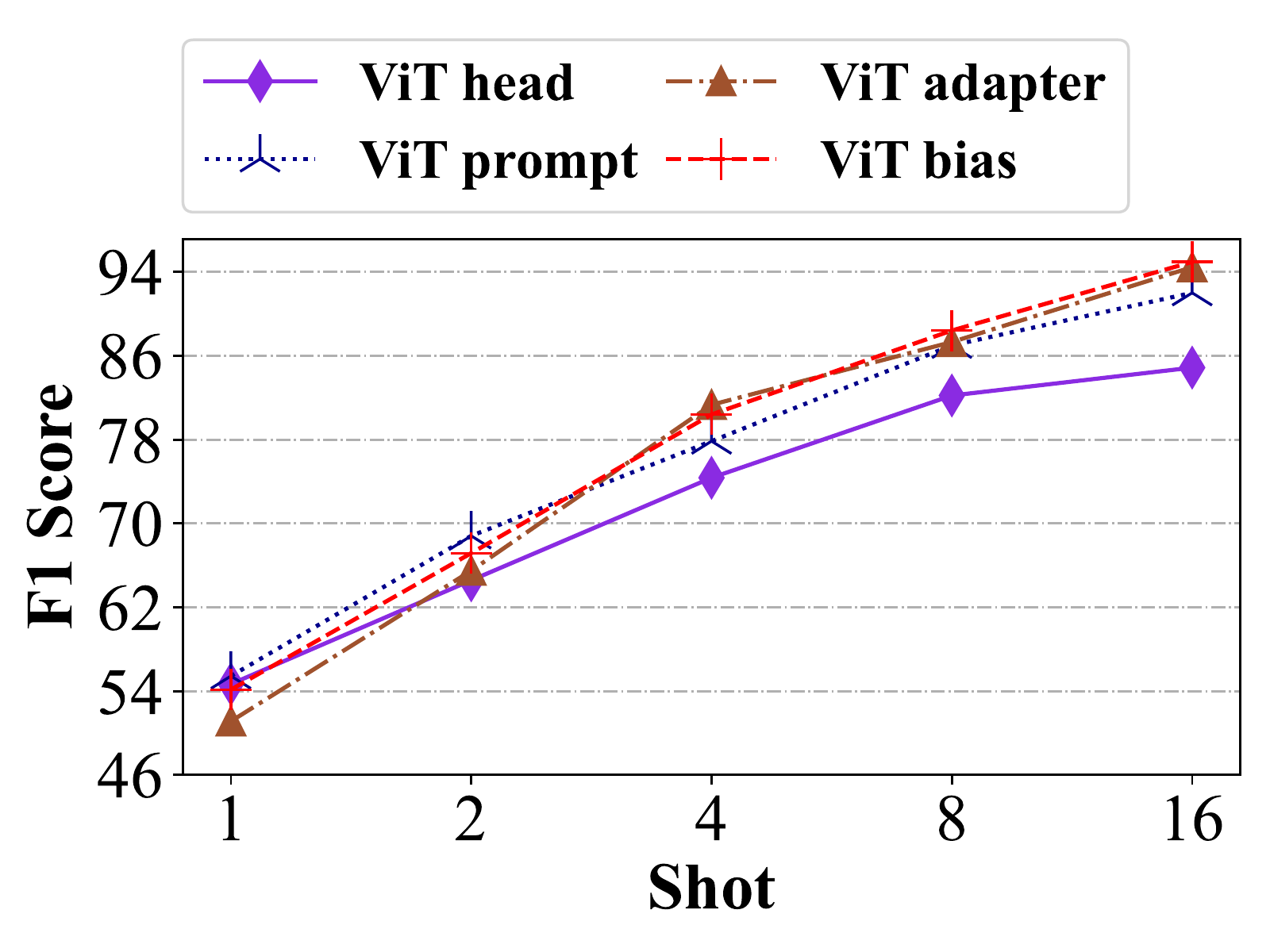}
    \caption{F1 score comparison of ViT models under non-IID setting.}
    \label{fig:f1-d}
  \end{subfigure}
  \caption{F1 score comparison under IID and non-IID settings on CIFAR-10 dataset.}
  \label{fig:f1}
\end{figure}

\begin{figure}[htp]
  \centering
   \includegraphics[width=1\linewidth]{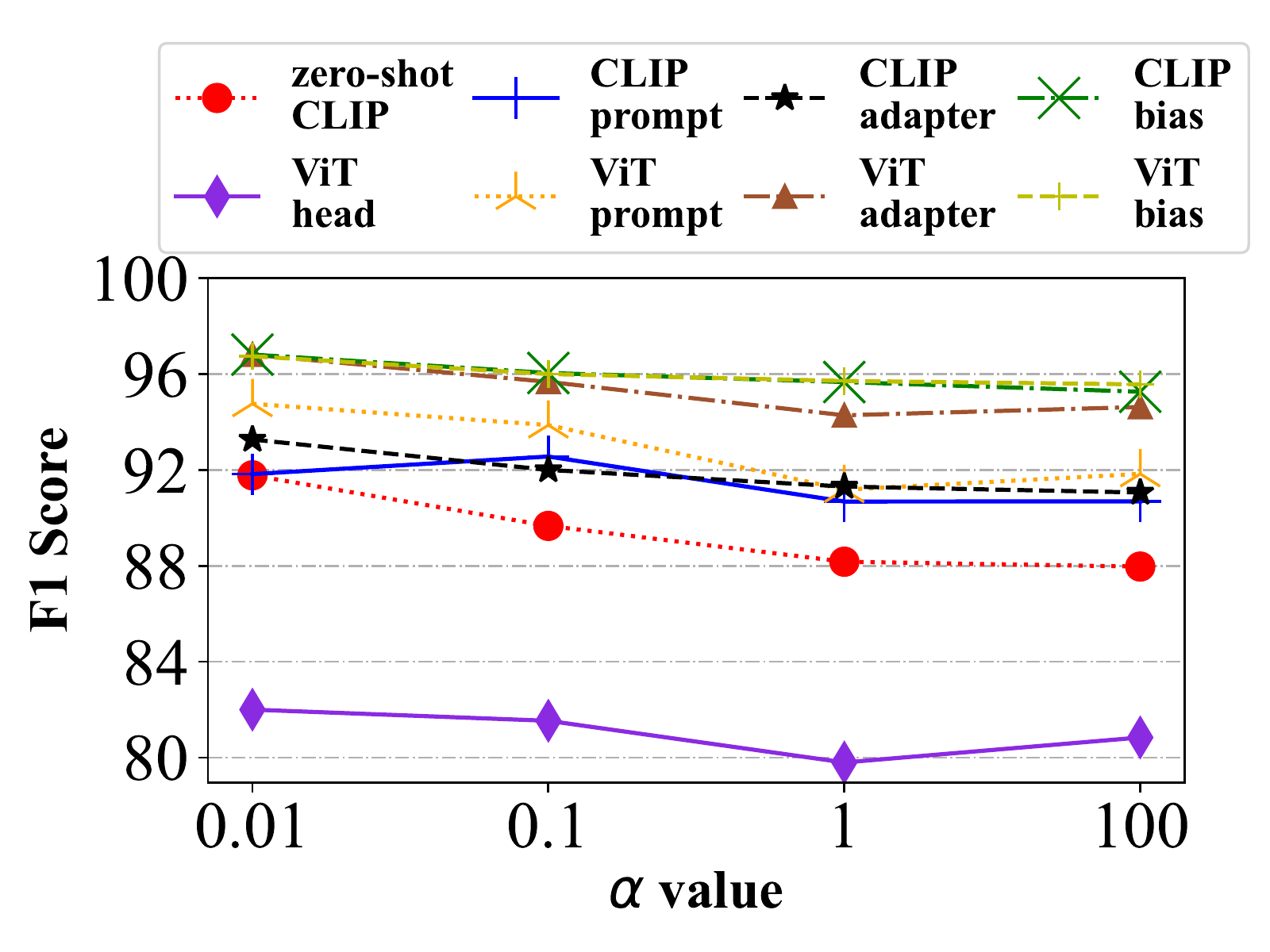}
   \caption{F1 score comparison under different Dirichlet distributions on CIFAR-10 dataset.}
   \label{fig:dir_f1}
\end{figure}

\begin{figure}[htp]
  \centering
  \begin{subfigure}{0.48\linewidth}
    \includegraphics[width=1\textwidth]{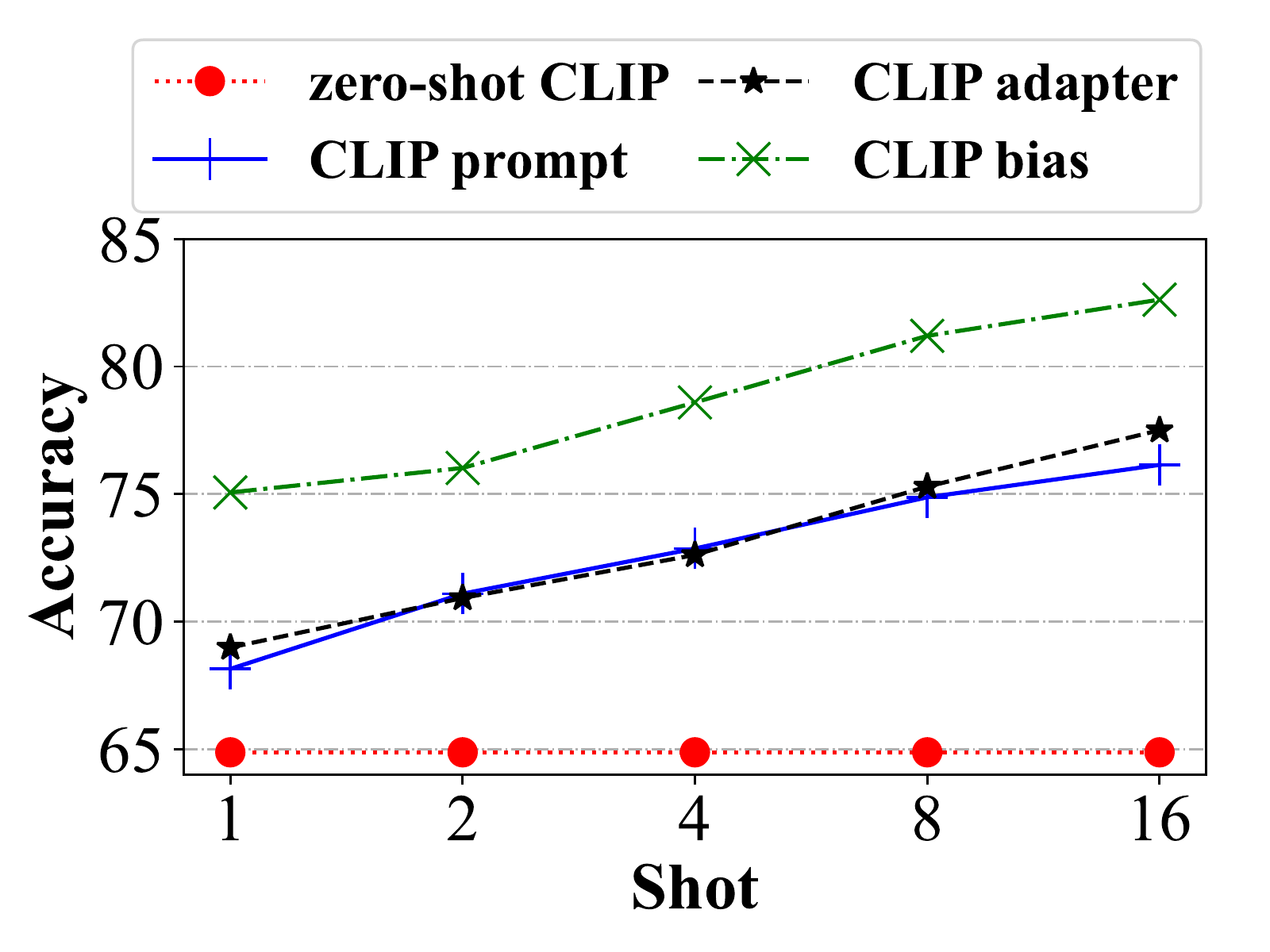}
    \caption{Accuracy comparison of CLIP models under IID setting on CIFAR-100 dataset.}
    \label{fig:cifar100_acc-a}
  \end{subfigure}
  \begin{subfigure}{0.48\linewidth}
    \includegraphics[width=1\textwidth]{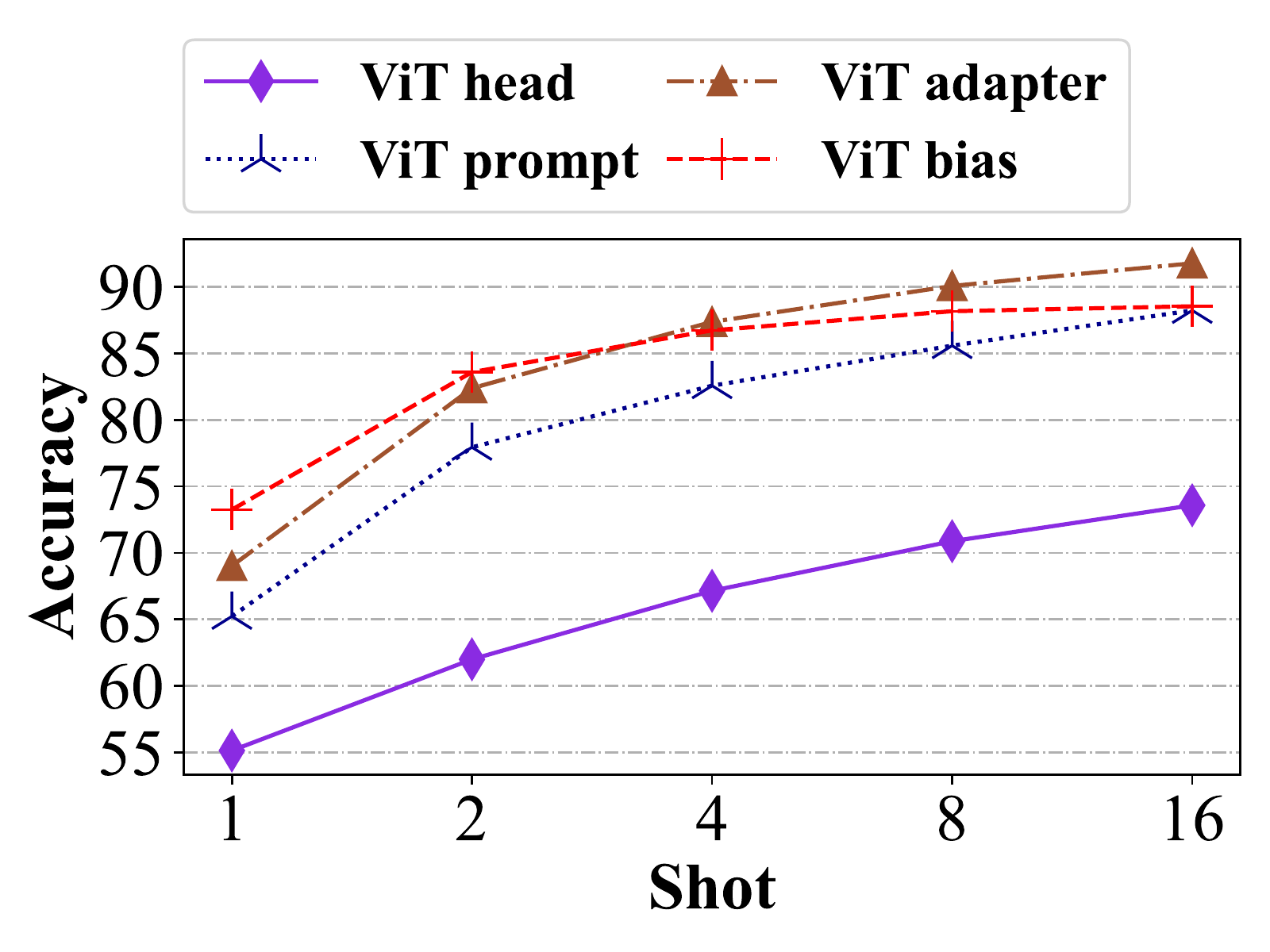}
    \caption{Accuracy comparison of ViT models under IID setting on CIFAR-100 dataset.}
    \label{fig:cifar100_acc-b}
  \end{subfigure}
  \label{fig:short}
  \begin{subfigure}{0.48\linewidth}
    \includegraphics[width=1\textwidth]{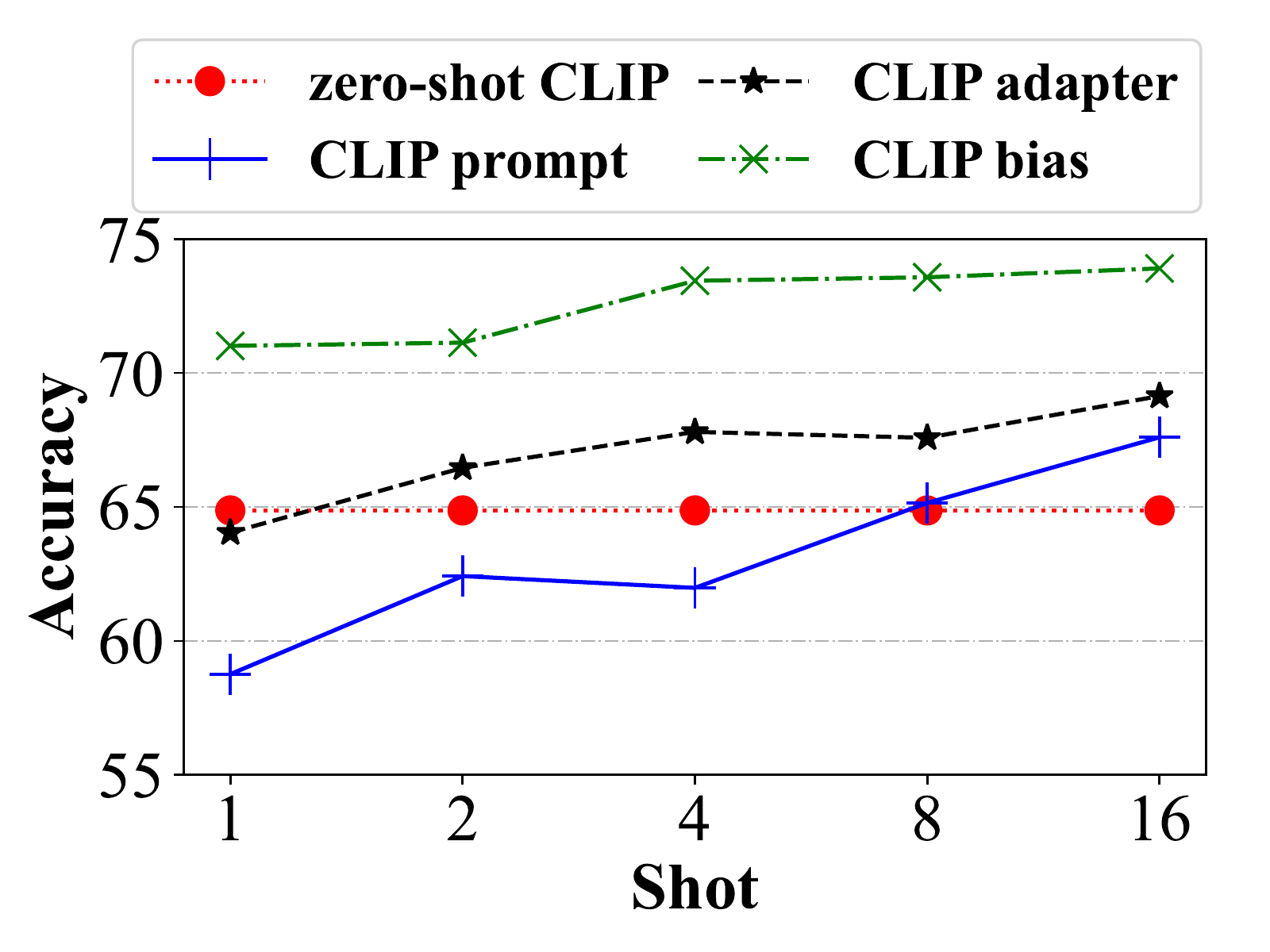}
    \caption{Accuracy comparison of CLIP models under non-IID setting on CIFAR-100 dataset.}
    \label{fig:cifar100_acc-c}
  \end{subfigure}
  \begin{subfigure}{0.48\linewidth}
    \includegraphics[width=1\textwidth]{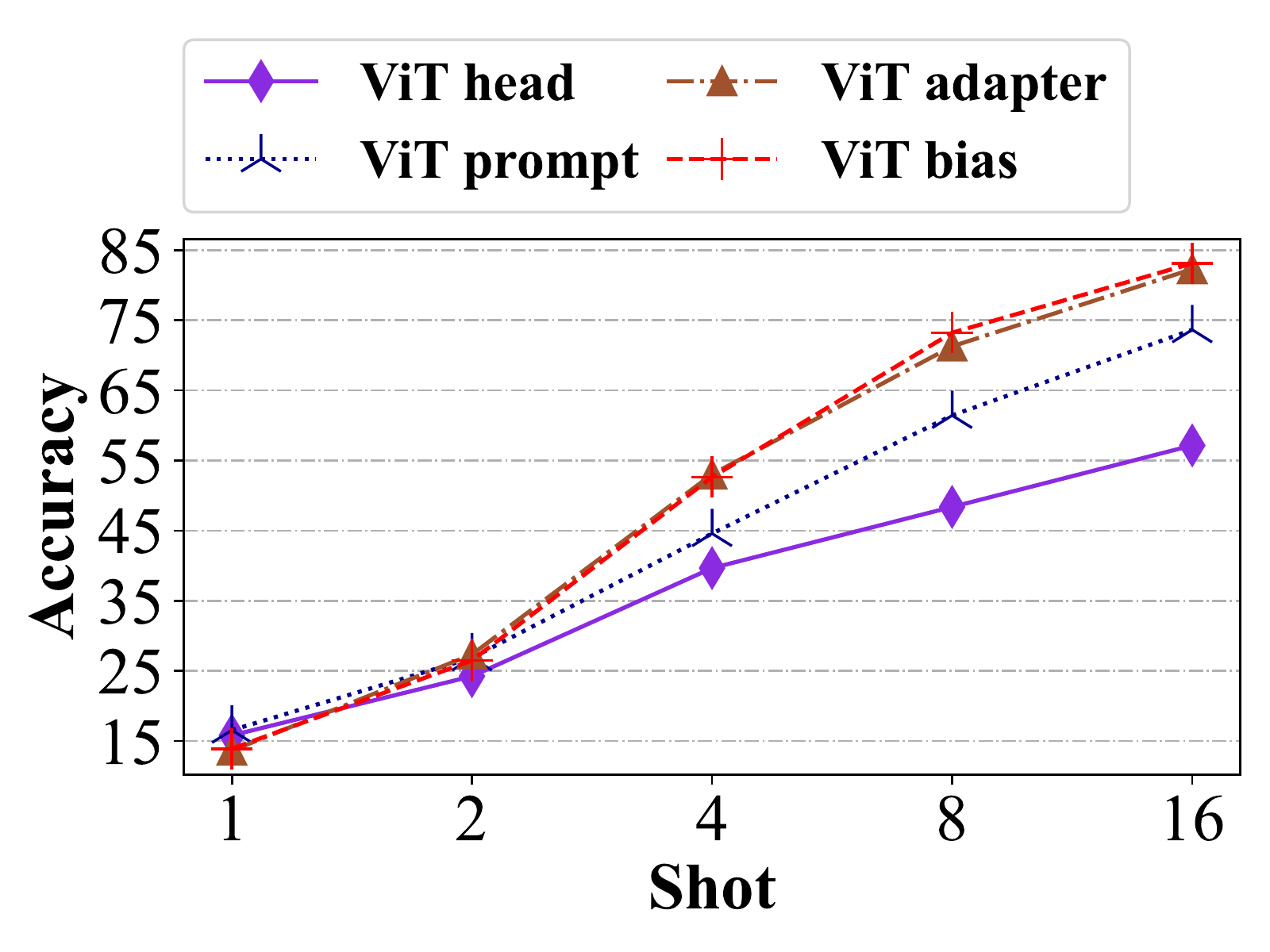}
    \caption{Accuracy comparison of ViT models under non-IID setting on CIFAR-100 dataset.}
    \label{fig:cifar100_acc-d}
  \end{subfigure}
  \caption{Accuracy comparison under IID and non-IID settings on CIFAR-100 dataset.}
  \label{fig:cifar100_acc}
\end{figure}

We also use the F1 score as an evaluation indicator.
Fig.~\ref{fig:f1} shows the F1 scores of different fine-tuning methods. For CLIP model, the F1 score of bias tuning is above 90\% in the IID setting and can reach 95.94\% in the 16-shot learning. The F1 scores of all tuning methods are higher than that of the zero-shot CLIP. For ViT methods, bias tuning and adapter tuning behave similarly, and their results increase from 73\% to 95\%. In Fig.~\ref{fig:f1-c}, CLIP bias is still better than other methods, and the score can reach 96.78\% in the 16-shot learning. It's observed that if the data is highly scarce (e.g., 1-shot learning), the performance of adapter learning and prompt learning is worse than that of the zero-shot learning because the model may overfit the local data and the clients' distributions are highly heterogeneous. In Fig.~\ref{fig:f1-d}, three tuning methods have similar performance. The data capacity has a large influence on the tuning methods of ViT.

The F1 scores of all tuning methods under various Dirichlet distributions are shown in Fig.~\ref{fig:dir_f1}. CLIP bias and ViT bias have the highest F1 scores, and the average values are higher than 95\%. The results of the F1 score are consistent with that of the accuracy.

\begin{figure}[t]
  \centering
   \includegraphics[width=0.8\linewidth]{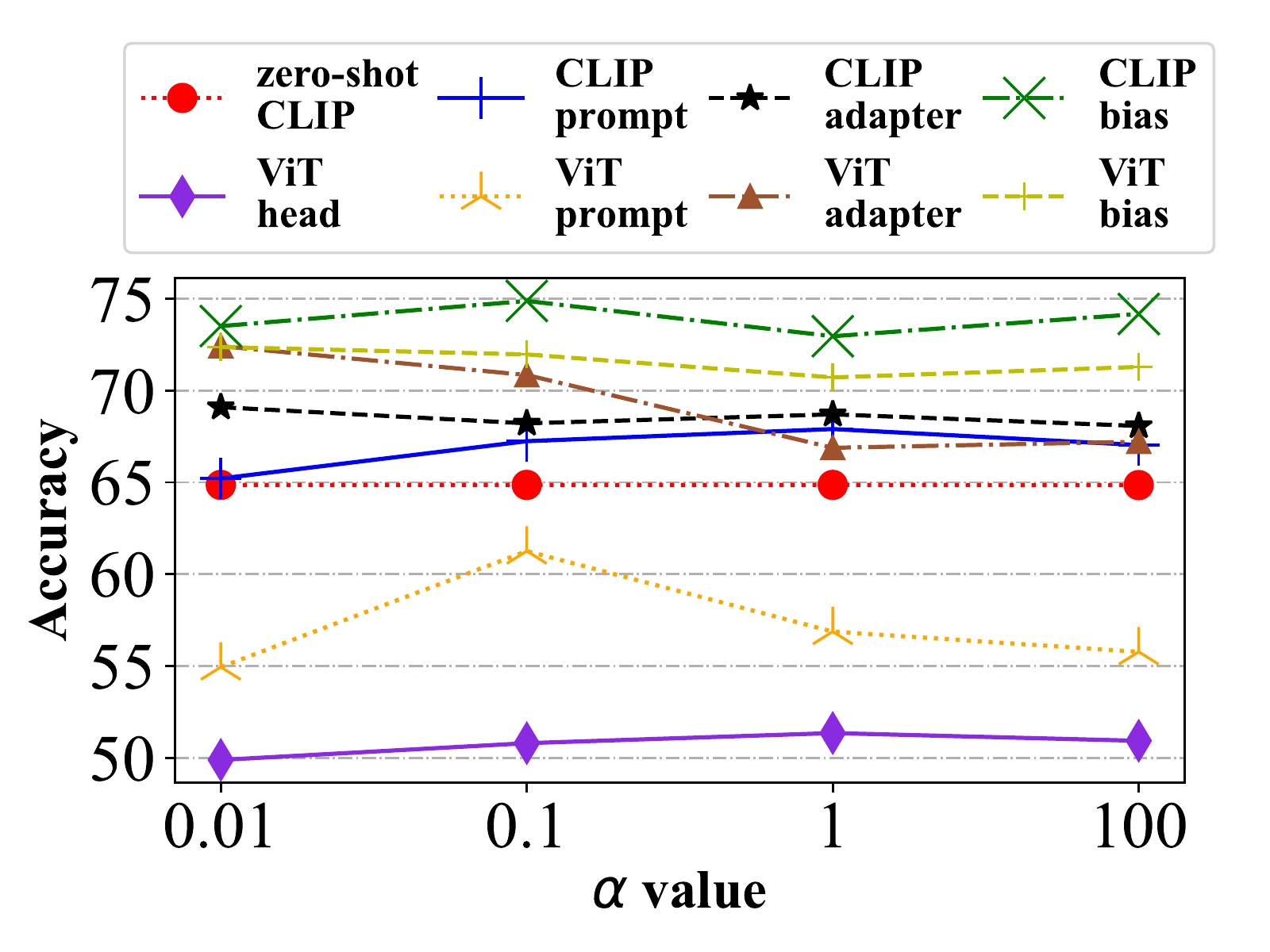}
   \caption{Accuracy comparison under different Dirichlet distributions on CIFAR-100 dataset.}
   \label{fig:cifar100_dir_acc}
\end{figure}

We evaluate the accuracy of different tuning methods on CIFAR-100 dataset. In Fig.~\ref{fig:cifar100_acc-a}, CLIP bias achieves the best performance, and its accuracy increases from 75.06\% to 82.62\%. CLIP prompt and CLIP adapter have similar performance, which are higher than zero-shot CLIP with 64.87\% accuracy. In Fig.~\ref{fig:cifar100_acc-b}, there is a noticeable increasing trend for ViT methods. ViT adapter has the highest value in the 16-shot learning, which is greater than 90\%. The accuracy of ViT prompt changes from 65.23\% to 88.23\%, which is about 15\% higher than that of the ViT head. In the non-IID setting, the accuracy of CLIP bias decreases to between 70\% and 75\%. The values of CLIP adapter and CLIP bias are 69.14\% and 67.6\% in the 16-shot learning. We can see that CLIP prompt is worse than zero-shot CLIP in the 1-shot, 2-shot, and 3-shot learning because of the issues of non-IID and over-fitting. Data distribution has a significant impact on the ViT methods as shown in Fig.~\ref{fig:cifar100_acc-d}. The values of ViT bias and ViT adapter increase from 15\% to 85\%. ViT head behaves the worst with 57.15\% accuracy in the 16-shot learning. It's observed that the performance on CIFAR-100 dataset is worse than that on CIFAR-10 dataset, and the foundation models are sensitive to the number of classes.

Accuracy comparison under different Dirichlet distributions is illustrated in Fig.~\ref{fig:cifar100_dir_acc}. CLIP bias achieves the best performance with about 75\% accuracy. Most of the methods have an accuracy between 65\% to 75\%, and the CLIP methods are better than the ViT methods. All approaches are robust to data heterogeneity of clients.

\section{Comparison with Local Training}

\begin{figure}[htp]
  \centering
  \begin{subfigure}{0.49\linewidth}
    \includegraphics[width=1\textwidth]{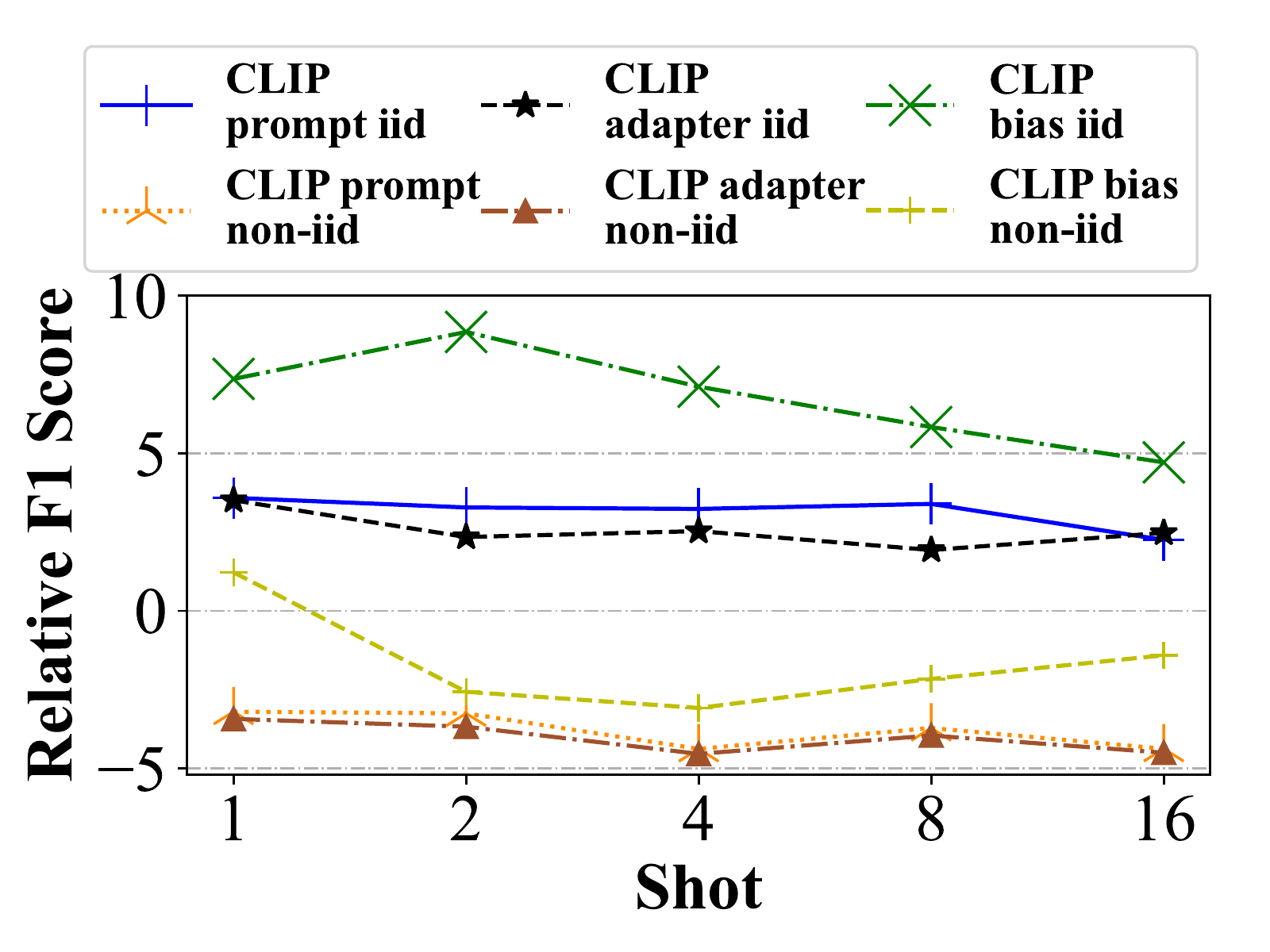}
    \caption{CLIP model.}
    \label{fig:local_clip_f1}
  \end{subfigure}
  \hfill
  \begin{subfigure}{0.49\linewidth}
    \includegraphics[width=1\textwidth]{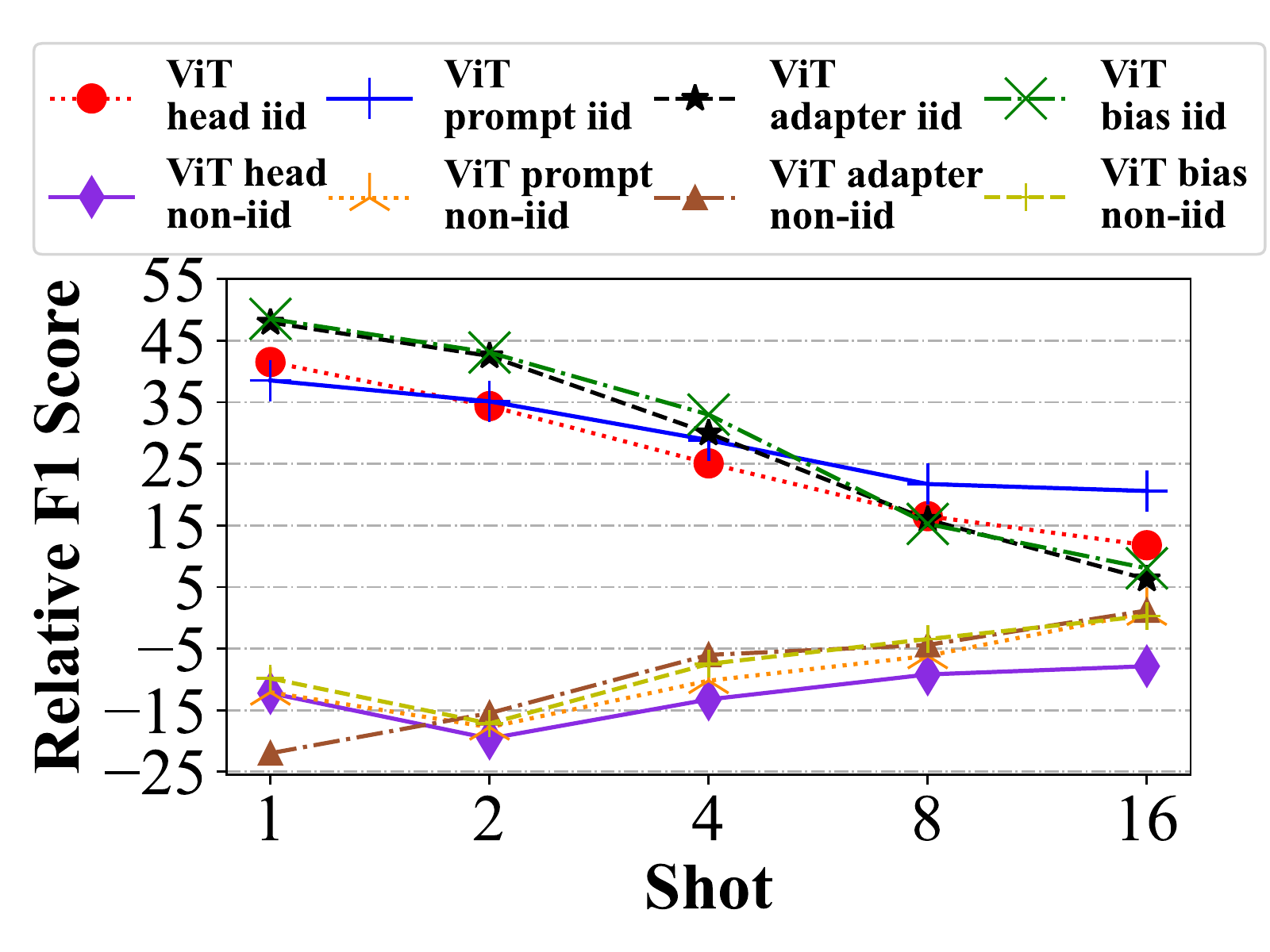}
    \caption{ViT model.}
    \label{fig:local_vit_f1}
  \end{subfigure}
  \caption{Relative F1 score comparison of FL and local training for different shots on CIFAR-10 dataset.}
  \label{fig:local_f1}
\end{figure}

\begin{figure}[htp]
  \centering
  \begin{subfigure}{0.49\linewidth}
    \includegraphics[width=1\textwidth]{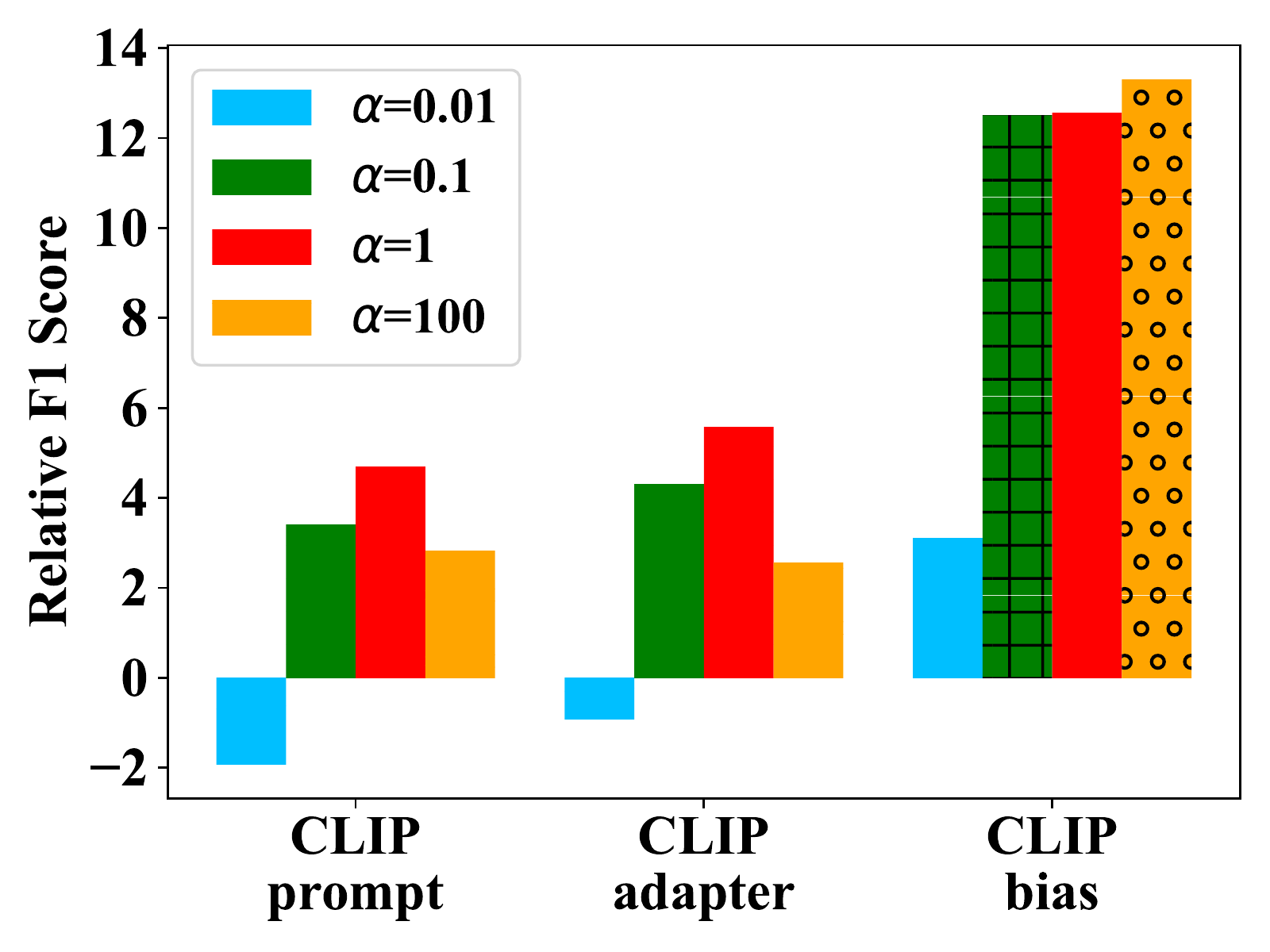}
    \caption{CLIP model.}
    \label{fig:local_clip_dir_f1}
  \end{subfigure}
  \hfill
  \begin{subfigure}{0.49\linewidth}
    \includegraphics[width=1\textwidth]{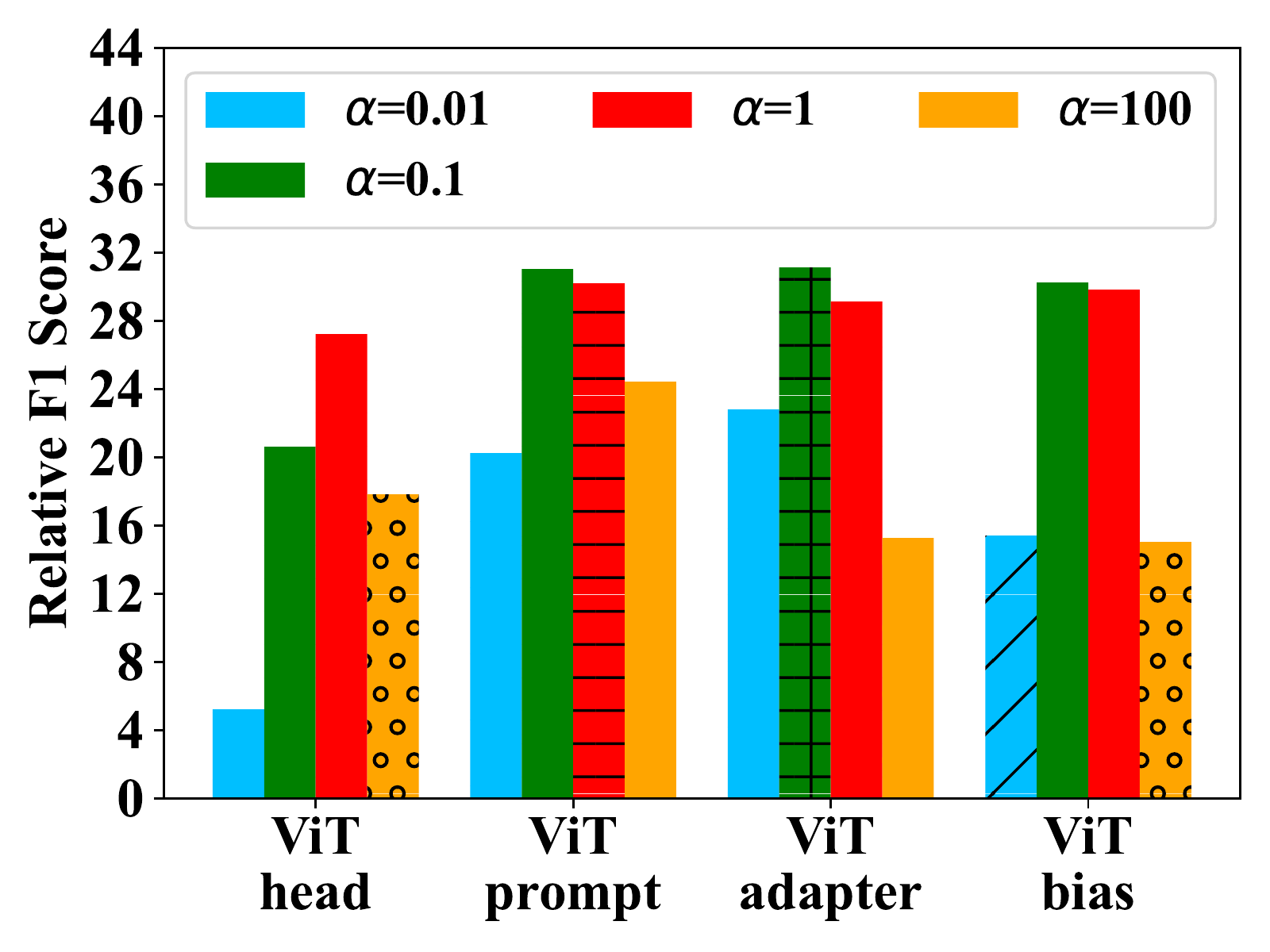}
    \caption{ViT model.}
    \label{fig:local_vit_dir_f1}
  \end{subfigure}
  \caption{Relative F1 score comparison of FL and local training for different Dirichlet distributions on CIFAR-10 dataset.}
  \label{fig:local_dir_f1}
\end{figure}

In this section, we show the F1 scores of tuning methods compared with local training in Fig.~\ref{fig:local_f1} and Fig.~\ref{fig:local_dir_f1}. We define the relative F1 score as the F1 score of FL minus the F1 score of local training. When data distribution is IID, the relative F1 scores can reach 8.84\% for CLIP and 48.53\% for ViT. However, when data distribution is non-IID, the results are worse than that of local training because there is a distribution drift among users. For ViT methods, the relative score decreases with shot numbers because the performance of local training improves with the data capacity. Besides, the F1 value shows a rising trend in Fig.~\ref{fig:local_vit_dir_f1}. It's because the increasing volumes of data can alleviate the issues of non-IID and improve the performance of FL.

\section{Comparison with the CNN Model}

\begin{figure}[htp]
  \centering
  \begin{subfigure}{0.49\linewidth}
    \includegraphics[width=1\textwidth]{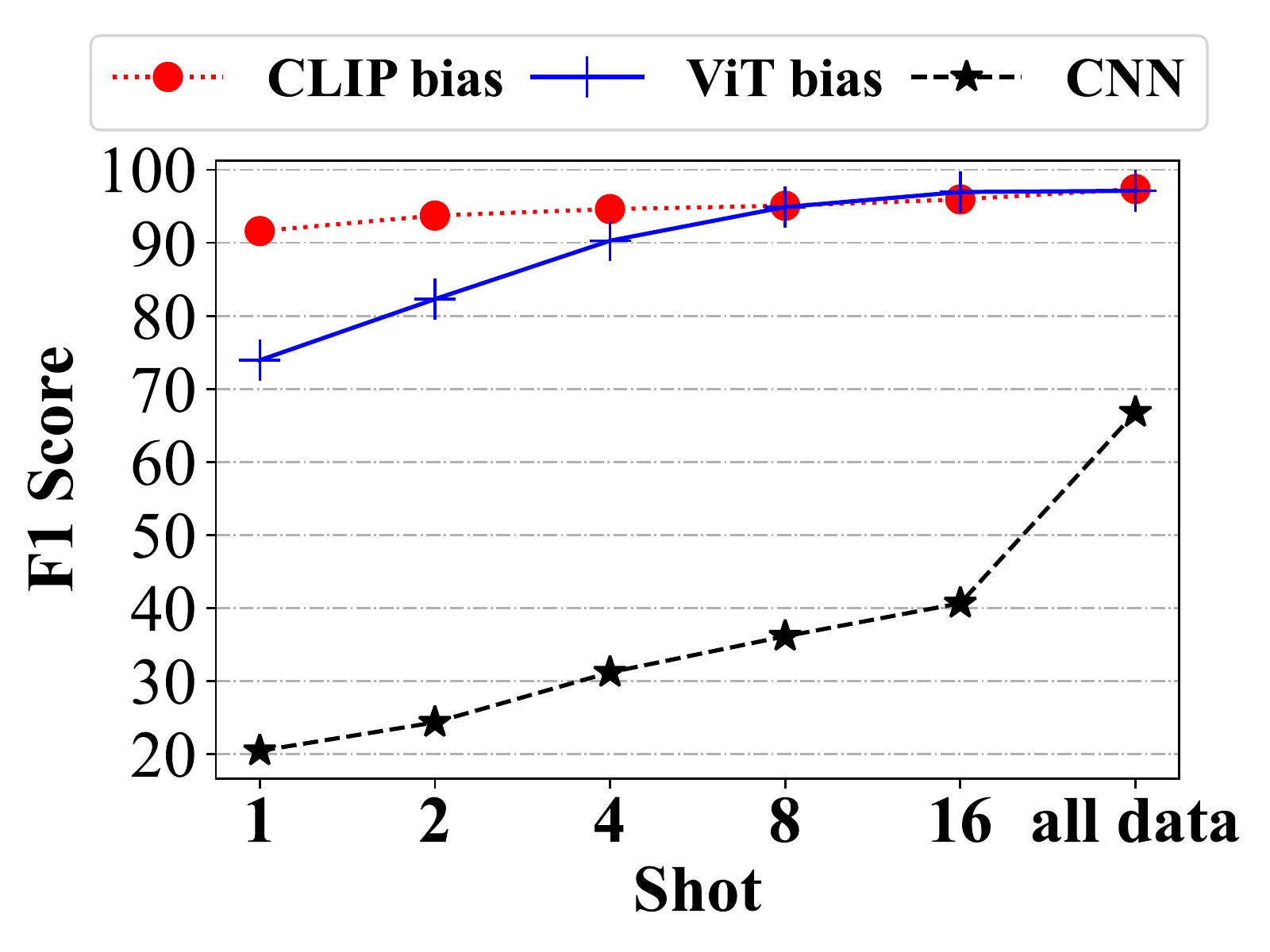}
    \caption{IID setting.}
    \label{fig:cnn_f1_iid}
  \end{subfigure}
  \hfill
  \begin{subfigure}{0.49\linewidth}
    \includegraphics[width=1\textwidth]{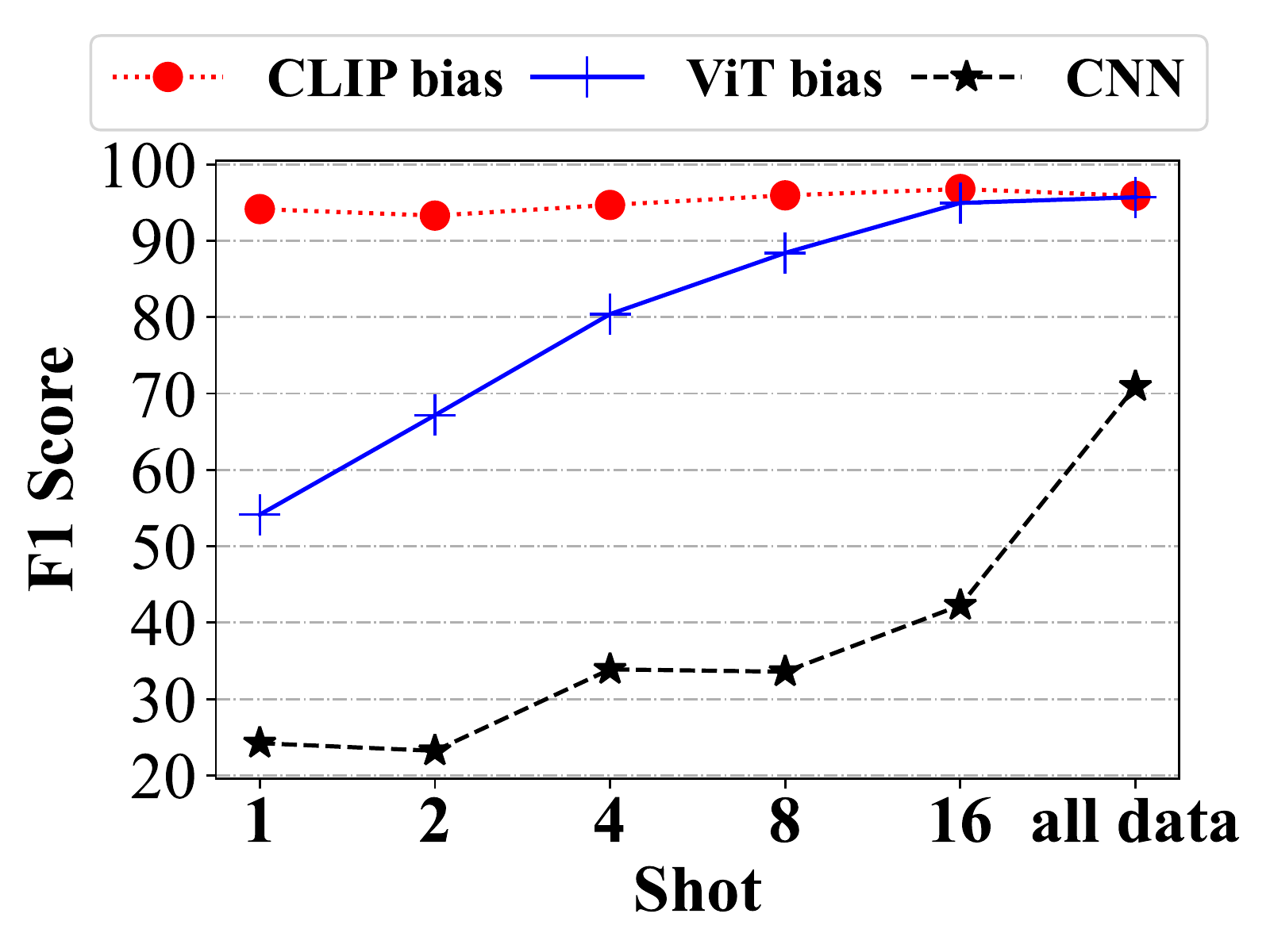}
    \caption{non-IID setting.}
    \label{fig:cnn_f1_noniid}
  \end{subfigure}
  \caption{F1 score comparison of foundation models and the CNN model on CIFAR-10 dataset.}
  \label{fig:cnn_f1}
\end{figure}

\begin{figure}[htp]
  \centering
   \includegraphics[width=0.6\linewidth]{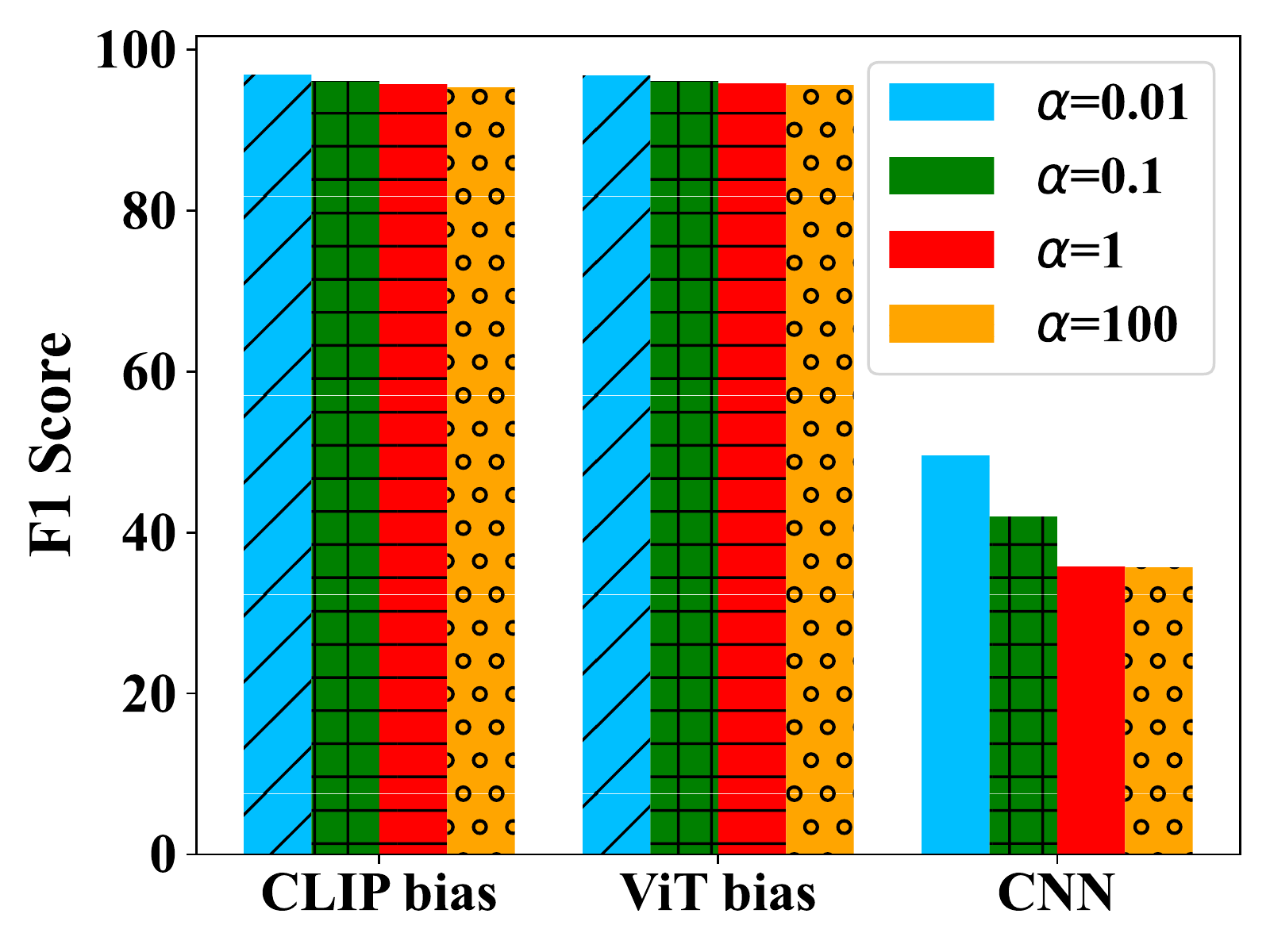}
   \caption{F1 score comparison of foundation models and the CNN model under different Dirichlet distributions on CIFAR-10 dataset.}
   \label{fig:cnn_dir_f1}
\end{figure}

In this section, we report the F1 scores of tuning methods compared with the traditional CNN model. FL with the pre-trained models can significantly improve the performance than the CNN model. In Fig.~\ref{fig:cnn_f1}, the F1 score shows an increasing trend because more data can be used for training. Little data don't contain too much information and will result in over-fitting. CLIP bias outperform the CNN model by 30.56\% and 24.97\% F1 scores when trained with overall data. Fig.~\ref{fig:cnn_dir_f1} shows the F1 scores under various Dirichlet distributions, and FL with fine-tuning methods can improve at least 47\% F1 scores.

\section{Comparison under Various Settings}

\subsection{Influence of the Foundation Model}

\begin{figure}[htp]
  \centering
  \begin{subfigure}{0.49\linewidth}
    \includegraphics[width=1\textwidth]{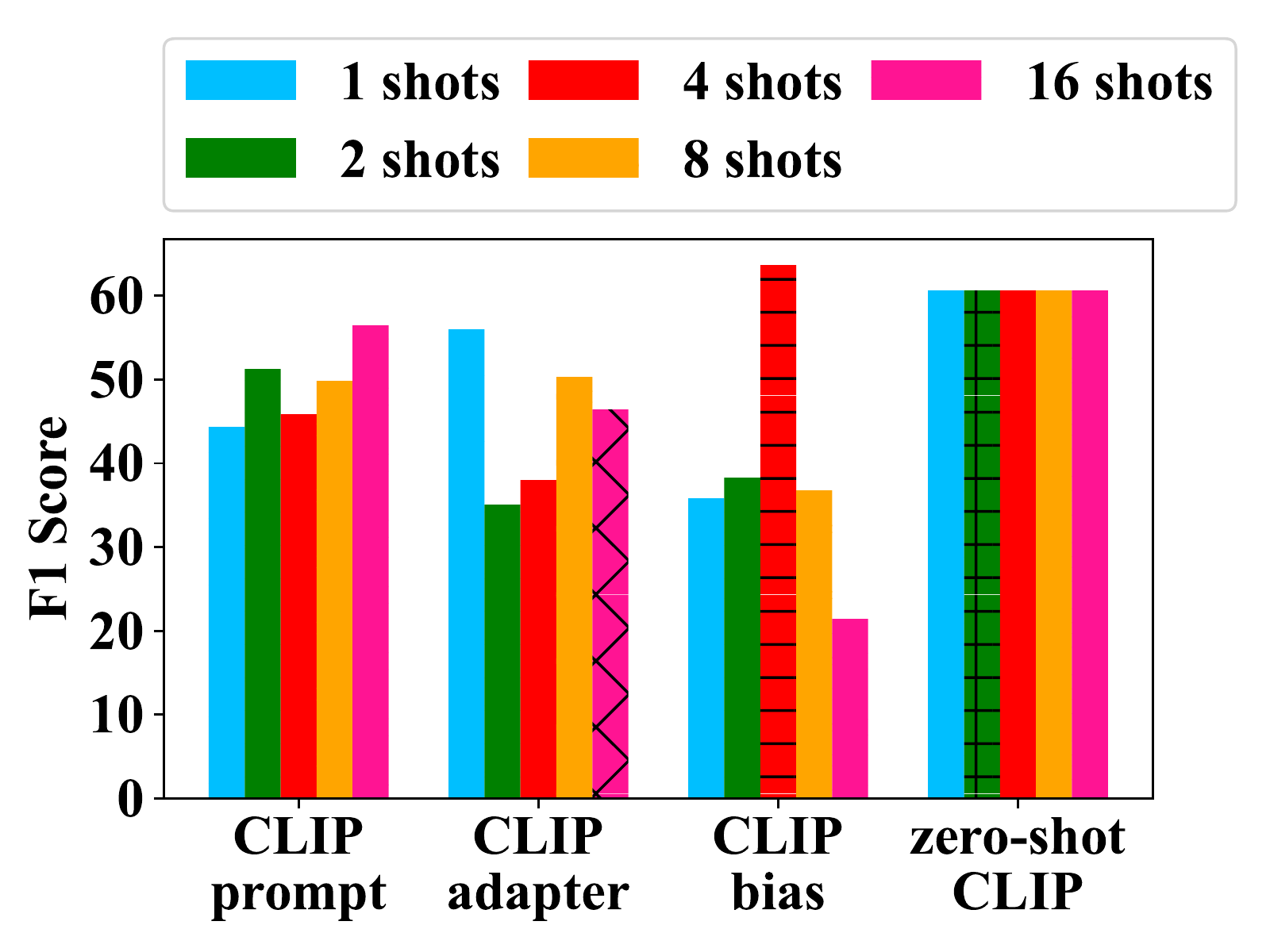}
    \caption{IID setting.}
    \label{fig:clip_res101_iid_f1}
  \end{subfigure}
  \hfill
  \begin{subfigure}{0.49\linewidth}
    \includegraphics[width=1\textwidth]{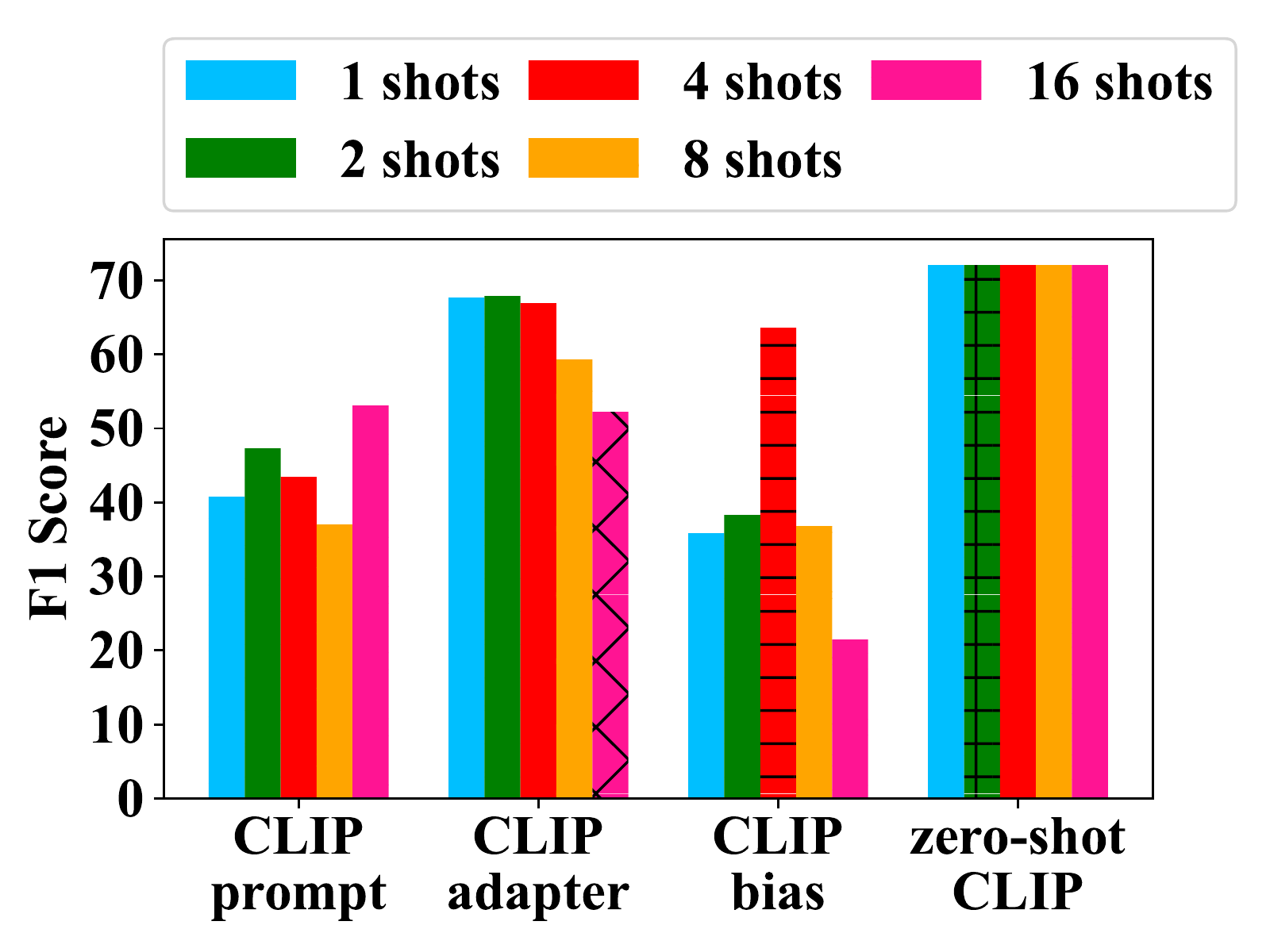}
    \caption{non-IID setting.}
    \label{fig:clip_res101_noniid_f1}
  \end{subfigure}
  \caption{F1 Score comparison of the CLIP ResNet101 model on CIFAR-10 dataset.}
  \label{fig:clip_res101_f1}
\end{figure}

We calculate the F1 scores of CLIP model with ResNet101 in Fig.~\ref{fig:clip_res101_f1}. It's observed that the results of fine-tuning methods are worse than that of the zero-shot CLIP. CLIP bias is most affected by the change of model architecture. Therefore, a strong Transformer-based pre-trained model is necessary for FL. 

\begin{figure}[htp]
  \centering
  \begin{subfigure}{0.49\linewidth}
    \includegraphics[width=1\textwidth]{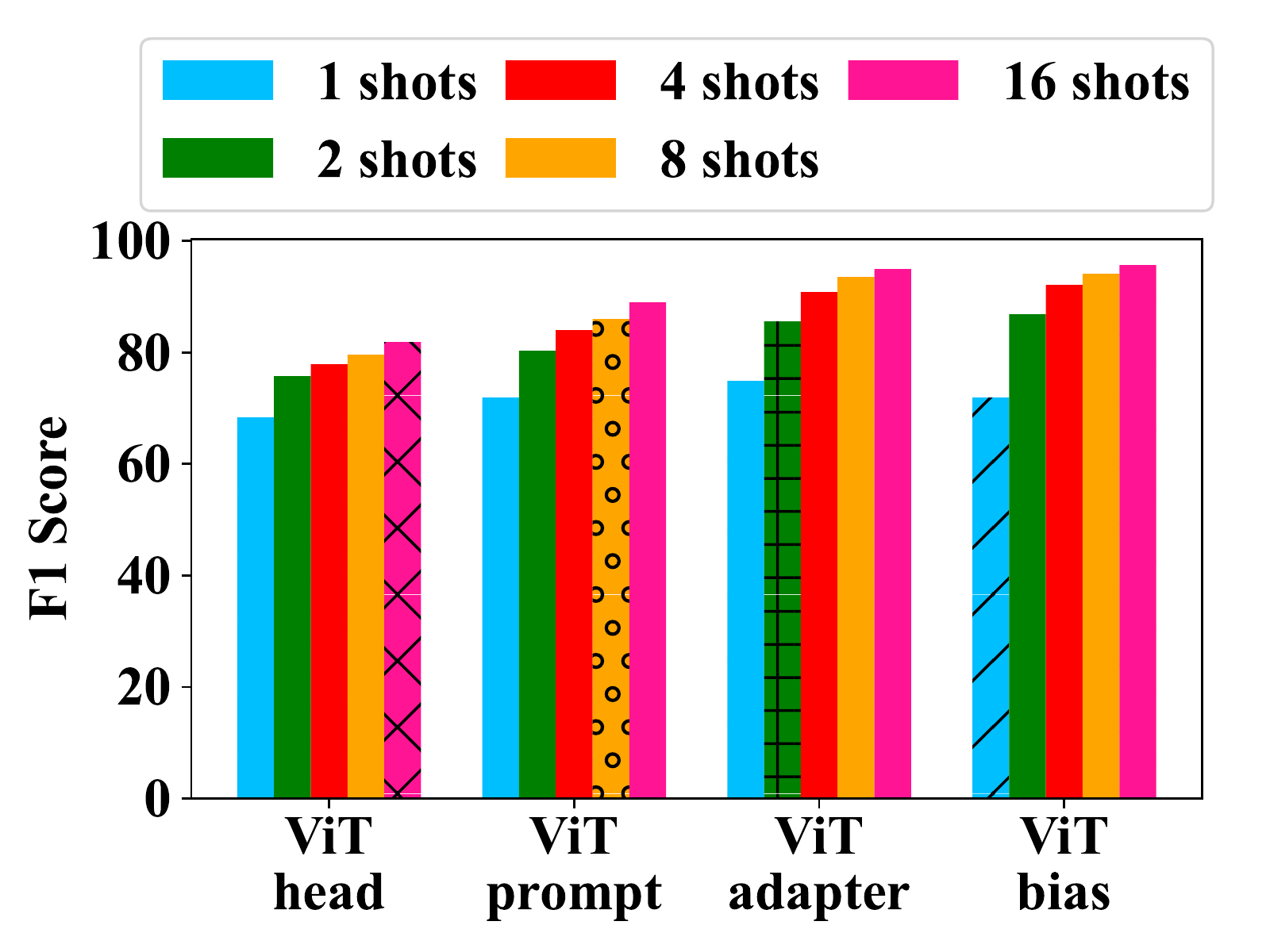}
    \caption{IID setting.}
    \label{fig:vit_small_iid_f1}
  \end{subfigure}
  \hfill
  \begin{subfigure}{0.49\linewidth}
    \includegraphics[width=1\textwidth]{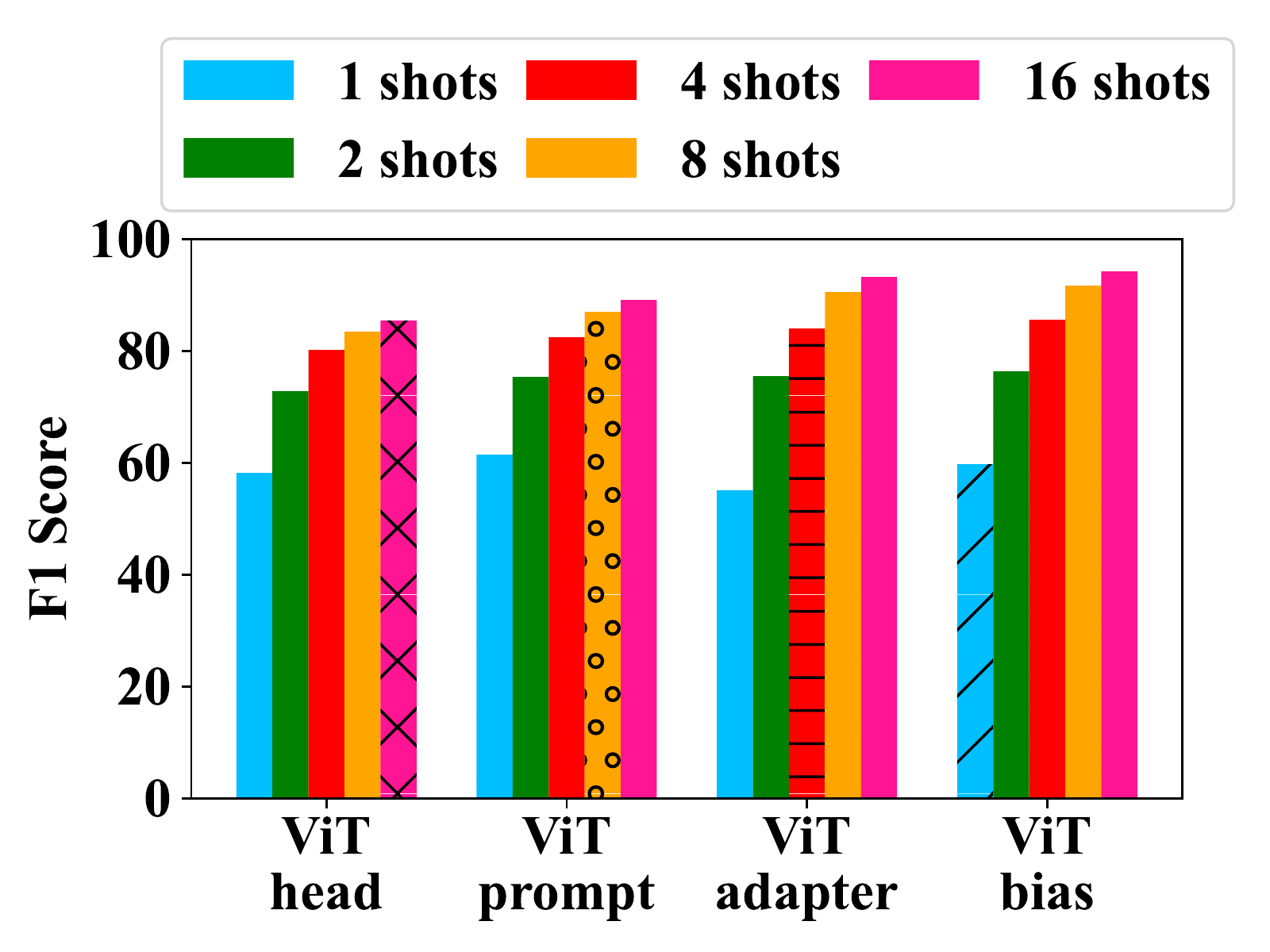}
    \caption{non-IID setting.}
    \label{fig:vit_small_noniid_f1}
  \end{subfigure}
  \caption{F1 score comparison of the ViT Small model on CIFAR-10 dataset.}
  \label{fig:vit_small_f1}
\end{figure}

We replace the ViT Base model with the ViT Small model and report the F1 scores in Fig.~\ref{fig:vit_small_f1}. We can see that the performance has a slight decrease. ViT bias behaves the best, and the values are still higher than 90\% in the 16-shot setting.

\subsection{Influence of the Number of Clients}

\begin{figure}[htp]
  \centering
  \begin{subfigure}{0.48\linewidth}
    \includegraphics[width=1\textwidth]{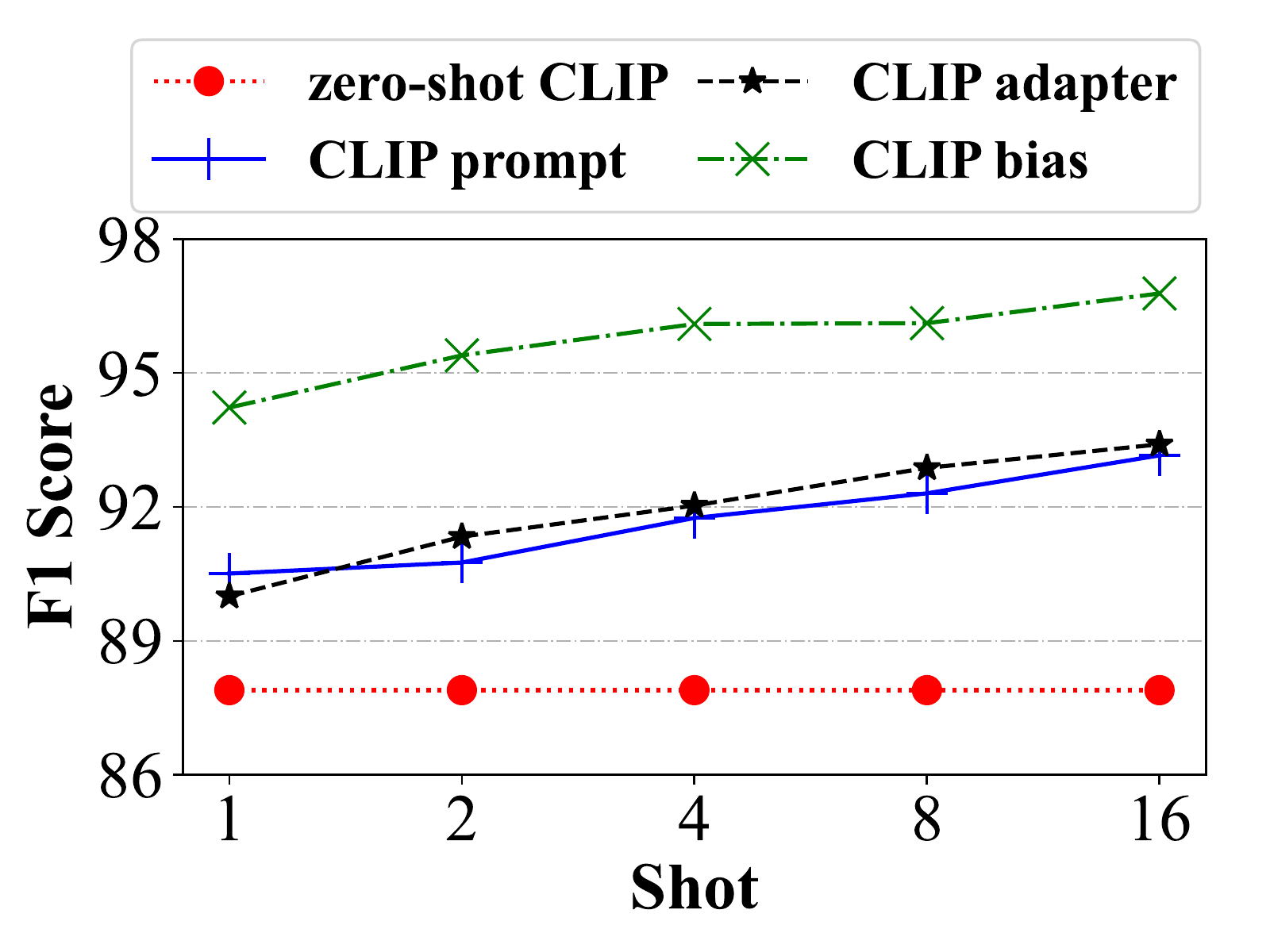}
    \caption{F1 score comparison of CLIP models under IID setting.}
    \label{fig:client50_f1-a}
  \end{subfigure}
  \begin{subfigure}{0.48\linewidth}
    \includegraphics[width=1\textwidth]{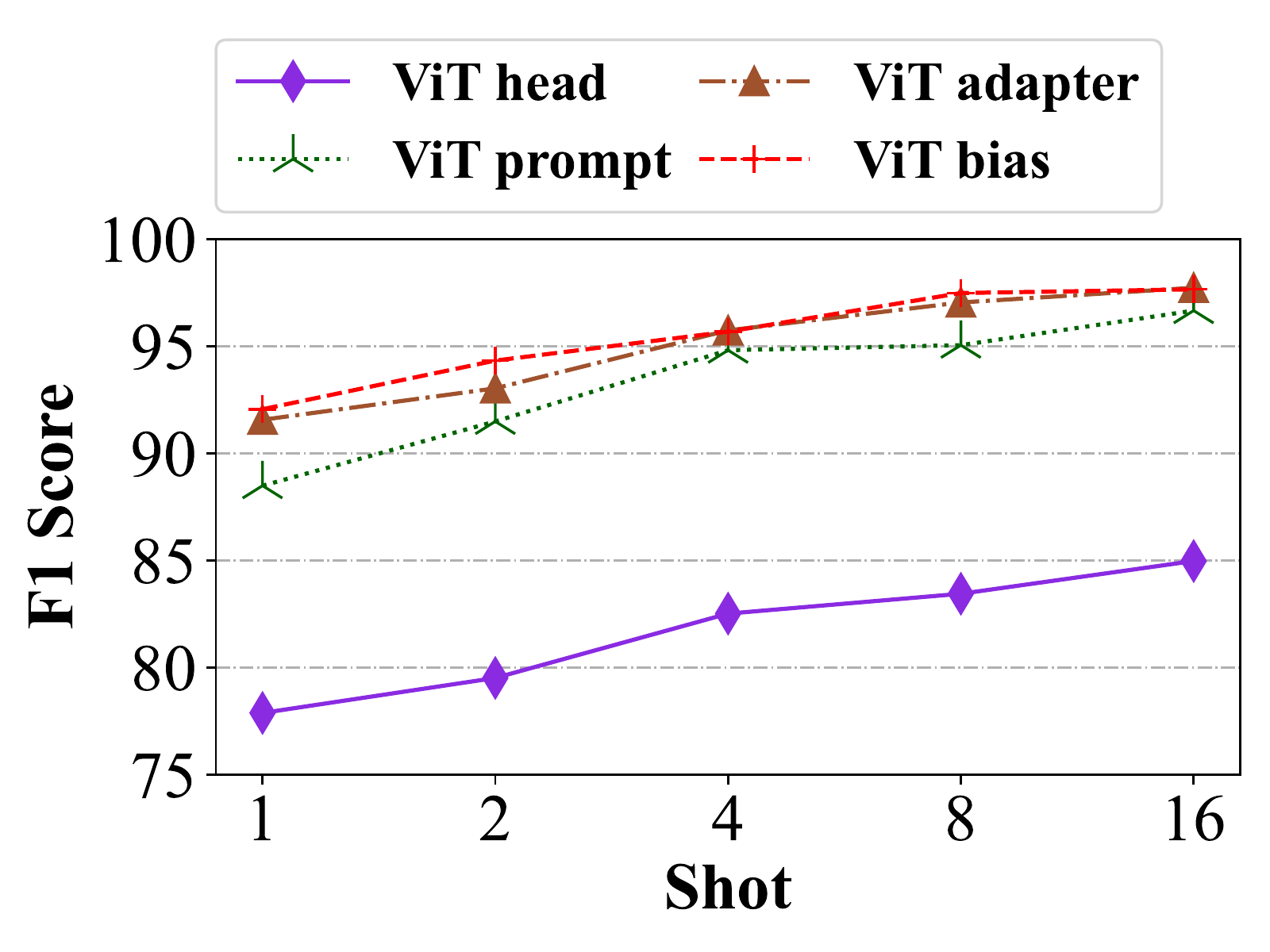}
    \caption{F1 score comparison of ViT models under IID setting.}
    \label{fig:client50_f1-b}
  \end{subfigure}
  \label{fig:short}
  \begin{subfigure}{0.48\linewidth}
    \includegraphics[width=1\textwidth]{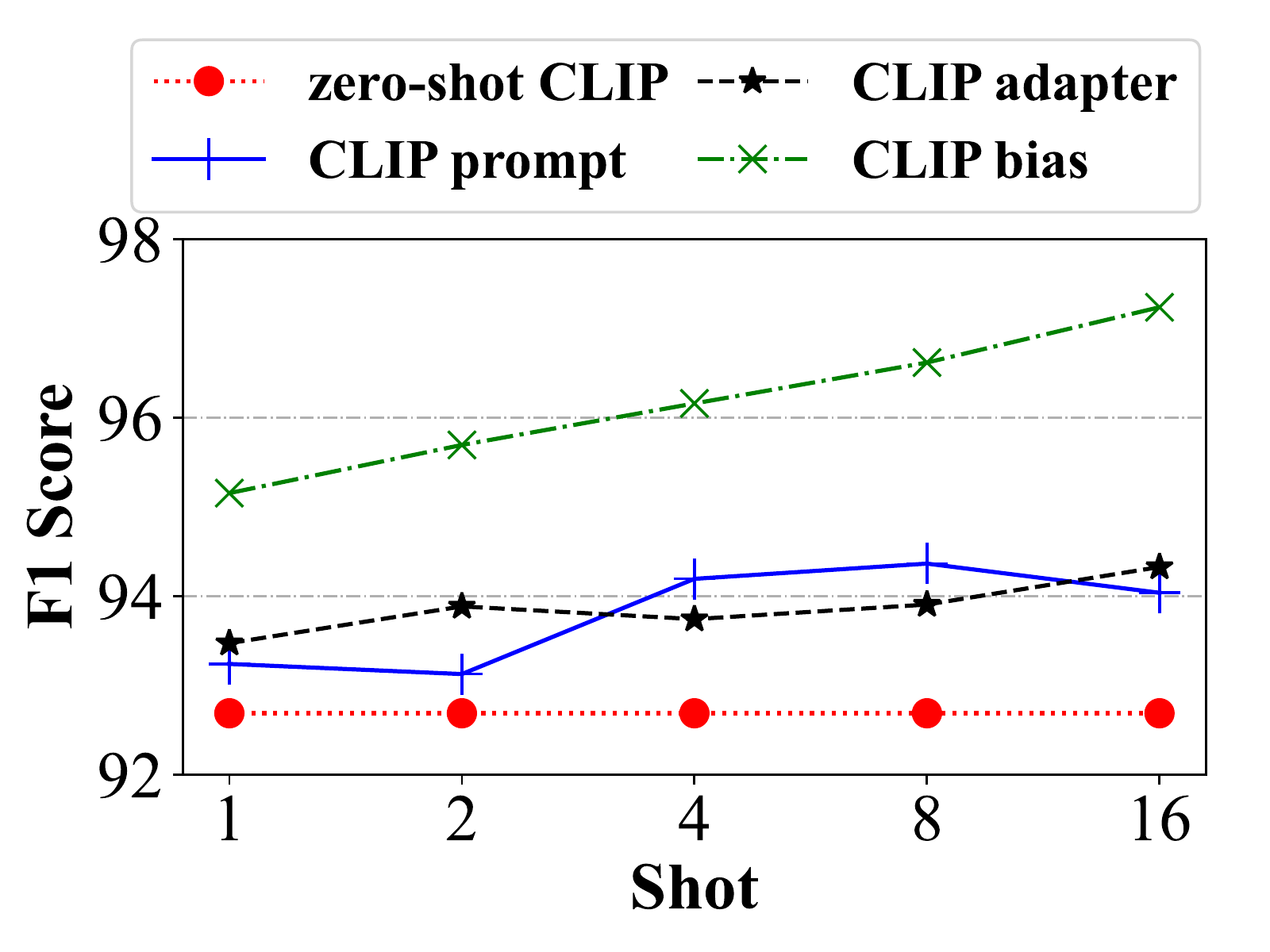}
    \caption{F1 score comparison of CLIP models under non-IID setting.}
    \label{fig:client50_f1-c}
  \end{subfigure}
  \begin{subfigure}{0.48\linewidth}
    \includegraphics[width=1\textwidth]{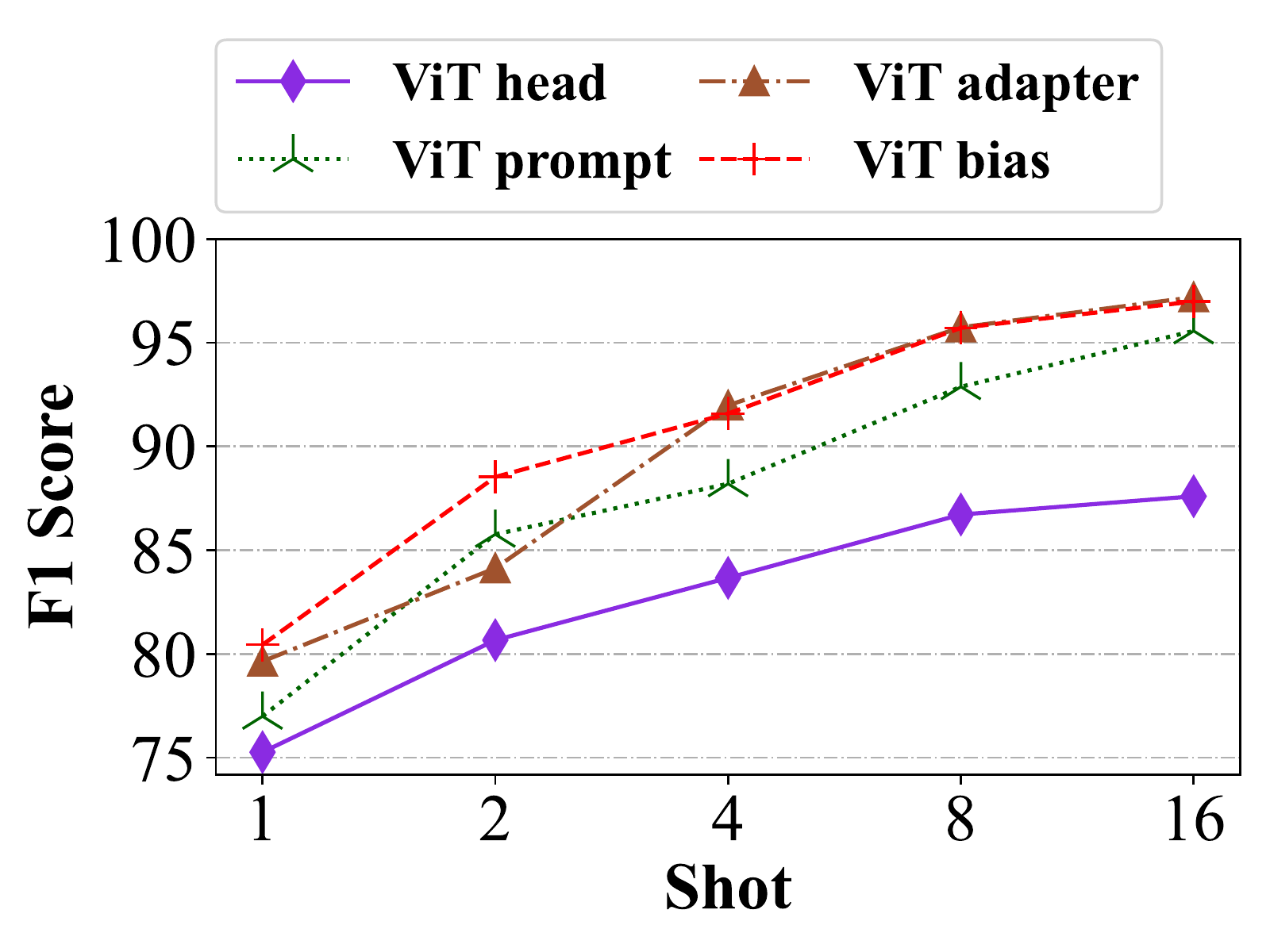}
    \caption{F1 score comparison of ViT models under non-IID setting.}
    \label{fig:client50_f1-d}
  \end{subfigure}
  \caption{F1 score comparison under IID and non-IID settings when the client number is 50 on CIFAR-10 dataset.}
  \label{fig:client50_f1}
\end{figure}

In this section, we investigate the influence of client number on F1 score, and the results are shown in Fig.~\ref{fig:client50_f1}.  When the client number is set as 50, CLIP bias and ViT bias still have the highest F1 values. 

\section{Resource Cost}

The global rounds and communication costs to meet the convergence condition are shown from Table.~\ref{tab:communication_1shot} to Table.~\ref{tab:communication_8shot}. All algorithms can converge within 15 global rounds and have small communication costs because all clients execute FL based on a powerful pre-trained model. 
We find that prompt learning is particularly lightweight and has little accuracy loss. Therefore, we can use prompt learning in FL when the network condition is poor.

\begin{table*}[htp]
  \centering
    \begin{tabular}{*{8}{c}}
      \toprule
      \multirow{2}*{Method} & \multicolumn{3}{c}{CLIP} & \multicolumn{4}{c}{ViT}\\
      \cmidrule(lr){2-4}\cmidrule(lr){5-8}
      & prompt & adapter & bias & head & prompt & adapter & bias \\
      \midrule
      Round/IID  & 4 & 3 & 3 & 7 & 7 &4 &5 \\
      Size/IID (MB) & 1.384 & 31.56 & 27.582 & 4.452 & 11.368 & 573.32 & 46.68 \\
      Round/non-IID & 3 & 4 & 1 & 4& 4 &4 &4 \\
      Size/non-IID (MB) &1.038 & 42.08 & 9.194 & 2.544& 6.496 &573.32 & 37.344 \\
      \bottomrule
    \end{tabular}
    \caption{The total communication size in the 1-shot learning on CIFAR-10 dataset.}
  \label{tab:communication_1shot}
\end{table*}

\begin{table*}
  \centering
    \begin{tabular}{*{8}{c}}
      \toprule
      \multirow{2}*{Method} & \multicolumn{3}{c}{CLIP} & \multicolumn{4}{c}{ViT}\\
      \cmidrule(lr){2-4}\cmidrule(lr){5-8}
      & prompt & adapter & bias & head & prompt & adapter & bias \\
      \midrule
      Round/IID  & 2 & 2 & 3 & 7 & 4 &4 &6 \\
      Size/IID (MB) & 0.692 & 21.04 & 27.582 & 4.452 & 6.496 & 573.32 & 56.016 \\
      Round/non-IID & 3 & 3 & 3 & 4& 8 &5 &4 \\
      Size/non-IID (MB) &1.038 & 31.56 & 27.582 & 2.544& 12.992 &716.65 & 37.344 \\
      \bottomrule
    \end{tabular}
    \caption{The total communication size in the 2-shot learning on CIFAR-10 dataset.}
  \label{tab:communication_2shot}
\end{table*}

\begin{table*}
  \centering
    \begin{tabular}{*{8}{c}}
      \toprule
      \multirow{2}*{Method} & \multicolumn{3}{c}{CLIP} & \multicolumn{4}{c}{ViT}\\
      \cmidrule(lr){2-4}\cmidrule(lr){5-8}
      & prompt & adapter & bias & head & prompt & adapter & bias \\
      \midrule
      Round/IID  & 2 & 2 & 4 & 7 & 7 &3 &4 \\
      Size/IID (MB) & 0.692 & 21.04 & 36.776 & 4.452 & 11.368 & 429.99 & 37.344 \\
      Round/non-IID & 3 & 3 & 2 & 10&11 &9 &8 \\
      Size/non-IID (MB) &1.038 & 31.56 & 18.388 & 6.36& 17.864 &1289.97 & 74.688 \\
      \bottomrule
    \end{tabular}
    \caption{The total communication size in the 4-shot learning on CIFAR-10 dataset.}
  \label{tab:communication_4shot}
\end{table*}

\begin{table*}
  \centering
    \begin{tabular}{*{8}{c}}
      \toprule
      \multirow{2}*{Method} & \multicolumn{3}{c}{CLIP} & \multicolumn{4}{c}{ViT}\\
      \cmidrule(lr){2-4}\cmidrule(lr){5-8}
      & prompt & adapter & bias & head & prompt & adapter & bias \\
      \midrule
      Round/IID  & 2 & 2 & 2 & 10 & 7 &3 &3 \\
      Size/IID (MB) & 0.692 & 21.04 & 18.388 & 6.36 & 11.368 & 429.99 & 28.008 \\
      Round/non-IID & 6 & 4 & 3 & 14&13 &9 &9 \\
      Size/non-IID (MB) &2.076 & 42.08 & 27.582 & 8.904 & 21.112 &1289.97 & 84.024 \\
      \bottomrule
    \end{tabular}
    \caption{The total communication size in the 8-shot learning on CIFAR-10 dataset.}
  \label{tab:communication_8shot}
\end{table*}

\end{appendices}
\end{document}